\documentclass[twoside]{article}

\usepackage[preprint]{aistats2026}
\usepackage[most]{tcolorbox}
\usepackage{empheq}

\usepackage[utf8]{inputenc} 
\usepackage[T1]{fontenc}    
\usepackage{booktabs}       
\usepackage{amsfonts}       
\usepackage{nicefrac}       
\usepackage{microtype}      
\usepackage[dvipsnames]{xcolor} 
\usepackage[mathscr]{eucal}
\usepackage{dsfont}
\usepackage{graphicx}
\usepackage{amsmath}
\usepackage{bm}
\usepackage{amsfonts}
\usepackage{tikz}
\usepackage{natbib}
\usepackage{textcomp}

\usepackage{algorithm,caption}
\usepackage{algorithmic}

\usepackage[utf8]{inputenc} 
\usepackage[T1]{fontenc}    
\usepackage{url}            
\usepackage{booktabs}       
\usepackage{amsfonts,amsmath}       
\usepackage{nicefrac}       
\usepackage{microtype}      
\usepackage[dvipsnames]{xcolor}         
\usepackage{amsmath,mleftright}
\allowdisplaybreaks
\usepackage{amsthm}

\usepackage{amsfonts,amsmath,amssymb,amsthm}       
\usepackage{bm}
\usepackage{graphicx}
\usepackage{subfigure}

\usepackage{dsfont}
\usepackage{xspace}
\usepackage{bigints}
\usepackage{amssymb,amsopn,float,bbm,bm,enumerate,color,multirow,gensymb}

\usepackage{epsfig,subfigure,graphicx}
\usepackage{comment}
\usepackage{afterpage}
\usepackage{thmtools,thm-restate}

\usepackage[suppress]{color-edits}

\addauthor{ab}{teal}
\addauthor{rd}{violet}
\addauthor{ql}{blue}
\addauthor{ms}{Green}

\newcommand{\beq}{\begin{equation}}
\newcommand{\eeq}{\end{equation}}

\newcommand\norm[1]{\left\lVert#1\right\rVert}
\renewcommand\vec[1]{\operatorname{vec}#1}

\newcommand\E{\mathbb{E}}
\renewcommand\P{\mathbb{P}}

\newcommand\R{\mathbb{R}}

\newcommand\1{\mathbbm{1}}

\newcommand{\g}{\mathbf{g}}

\newcommand{\w}{\mathbf{w}}
\newcommand{\x}{\mathbf{x}}

\newcommand{\cC}{{\cal C}}
\newcommand{\cH}{{\cal H}}

\newcommand{\cL}{{\cal L}}
\newcommand{\cM}{{\cal M}}
\newcommand{\cN}{{\cal N}}

\newcommand{\cT}{{\cal T}}

\newcommand{\cX}{{\cal X}}

\newcommand{\cF}{{\cal F}}

\newcommand{\cA}{{\cal A}}
\newcommand{\cB}{{\cal B}}

\newcommand{\cG}{{\cal G}}

\newcommand{\bxi}{\boldsymbol{\xi}}

\DeclareMathOperator{\argmax}{argmax}
\DeclareMathOperator{\argmin}{argmin}

\definecolor{darkgreen}{rgb}{0.05, 0.5, 0.2}
\newtheorem{innerthm}{Theorem}[section]
\newenvironment{theoremBox}[1][]{
    \refstepcounter{innerthm}%
    \begin{tcolorbox}[
        left=5pt, right=5pt,  
        colback=gray!1,     
        colframe=gray!150,    
        colbacktitle=gray!20,
        coltitle=black,      
        fonttitle=\bfseries, 
        title={ \color{Black} Theorem~\theinnerthm}%
          {\color{Black}\if{\color{Black}\relax}\detokenize{#1}\relax
          \else: #1\fi }    
    ]
}{
    \end{tcolorbox}
}

\newtheorem{asmp}{Assumption}

\newtheorem{remark}{Remark}[section]

\newcommand {\commentout}[1] {}

\def\ints{{{\rm Z} \kern -.35em {\rm Z} }}  
\def\smallints{{{\rm Z} \kern -.3em {\rm Z} }}  
\def\pints{{{\rm I} \kern -.15em {\rm N} }}      
\newcommand{\reals}{\mathbb R}

\def\cplx{{{\rm I} \kern -.45em {\rm C} }}       
\def\l2{\rm {\mathcal L}^{2}(\reals)}            

\newcommand{\be}{\begin{eqnarray}}
\newcommand{\ee}{\end{eqnarray}}
\newcommand{\bea}{\begin{eqnarray}}
\newcommand{\eea}{\end{eqnarray}}
\newcommand{\beaa}{\begin{eqnarray*}}
\newcommand{\eeaa}{\end{eqnarray*}}
\newcommand{\bnad}{\begin{nad}}
\newcommand{\enad}{\end{nad}}

\newcommand{\bbr}{\mathbb{R}}

\newcommand{\eps}{\varepsilon}

\newcommand{\bbR}{\mathbb{R}}
\newcommand{\cO}{\mathcal{O}}

\newcommand{\IGNORE}[1]{}

﻿

\renewcommand{\E} {\operatornamewithlimits{\ensuremath{\mathbb{E}}}} 

{\begin{array}{l@{\hspace{8mm}}l@{\hspace{8mm}}l}}%
{\end{array}}
\newlength{\lplb}
\newcommand{\reg}{\ensuremath{\mathtt{Reg_{CB}}}}

\DeclareMathSymbol{\mhyphen}{\mathord}{AMSa}{"39}

\newcommand{\Crv}{C_{\cM, \cN}(\bxi)}

\newcommand{\sketch}{\mathscr{S}}

\newcommand{\skLinUCB}{\color{Green}\ensuremath{\mathtt{sk\mhyphen LinUCB}}}
\newcommand{\skLinTS}{\color{Green}\ensuremath{\mathtt{sk\mhyphen LinTS}}}

\newcommand{\boldS}{\mathbf{S}}
\newcommand{\cU}{\mathcal{U}}
\newcommand{\cV}{\mathcal{V}}

\newtheorem{definition}{Definition}[section]
\newtheorem{nad}{Notation and Definitions}[section]

\newtheorem{theorem}[innerthm]{Theorem}
\newtheorem{lemma}[innerthm]{Lemma}

\newtheorem{proposition}{Proposition}[section]

\usepackage[colorlinks=true]{hyperref}

\hypersetup{
  pdffitwindow=true,
  pdfstartview={FitH},
  pdfnewwindow=true,
  colorlinks,
  linktocpage=true,
  urlcolor=Green,
  linkcolor=Red,
  citecolor=Blue
}
\usepackage[capitalize,noabbrev]{cleveref}

\usepackage{url}
\usepackage{cleveref}
\usepackage{enumitem}
\setlist{leftmargin=25pt,labelindent=25pt}
\setlist[enumerate]{align=left}

\begin{document}

%

%

\twocolumn[

\aistatstitle{Beyond Johnson-Lindenstrauss: Uniform Bounds for Sketched Bilinear Forms}

\aistatsauthor{ Rohan Deb \And Qiaobo Li \And Mayank Shrivastava \And Arindam Banerjee }

\aistatsaddress{Siebel School of Computing and Data Science, University of Illinois, Urbana Champaign } ]

\begin{abstract}

Uniform bounds on sketched inner products of vectors or matrices underpin several important computational and statistical results in machine learning and randomized algorithms, including the Johnson-Lindenstrauss (J-L) lemma, the Restricted Isometry Property (RIP), randomized sketching, and approximate linear algebra.  
However, many modern analyses involve \emph{sketched bilinear forms}, for which existing uniform bounds either do not apply or are not sharp on general sets. 
In this work, we develop a general framework to analyze such sketched bilinear forms and derive uniform bounds in terms of geometric complexities of the associated sets.  
Our approach relies on generic chaining and introduces new techniques for handling suprema over pairs of sets.  
We further extend these results to the setting where the bilinear form involves a sum of $T$ independent sketching matrices and show that the deviation scales as $\sqrt{T}$.  
This unified analysis recovers known results such as the J-L lemma as special cases, while extending RIP-type guarantees.  
Additionally, we obtain improved convergence bounds for sketched Federated Learning algorithms where such cross terms arise naturally due to sketched gradient compression, and design sketched variants of bandit algorithms with sharper regret bounds that depend on the geometric complexity of the action and parameter sets, rather than the ambient dimension.
 
\end{abstract}

\section{Introduction}
\label{sec:intro}
Sketching is a powerful technique for dimensionality reduction, widely used across many areas of numerical linear algebra and machine learning (ML). 
In linear regression, it enables solving large-scale least squares problems efficiently by reducing the size of the data matrix while preserving key geometric properties \citep{CW13, MM13, SWZ17, ALS18} 
while in distributed or federated learning, it facilitates communication-efficient protocols for matrix multiplication and aggregation \citep{Song_sketching,shrivastava2024sketching,BWZ16}. 
In recent times, sketching has also been applied to several other key areas, such as reinforcement learning \citep{WZD20,SSX23,Kuzborskij}, training deep networks \citep{SYZ21, SZZ21, GQSW22}, and convex programming \citep{JSWZ21, SY21, QSZZ23} (see Appendix~\ref{sec:related} for a detailed related works).

A foundational result, the Johnson–Lindenstrauss (J-L) Lemma formalizes such a geometric guarantee, showing that random projections approximately preserve pairwise distances \citep{joli84,aich06}. 
More generally, the \emph{Restricted Isometry Property} (RIP) \citep{bcfs14,nrwy12,vers14} provides a framework for understanding when a random matrix acts as an approximate isometry over a structured set, and underpins theoretical guarantees in compressed sensing and high-dimensional estimation. A matrix $\boldS \in \mathbb{R}^{b \times d}$ is said to satisfy RIP for a sparse set $\cU \subseteq \R^{d}$ with constant $\delta_r$ if the following holds for all $u \in  \cU$:
\begin{align}
\label{eq:RIP}
(1 - \delta_r) \| u \|_2^2 \leq \| \boldS u \|_2^2 \leq (1 + \delta_r) \| u \|_2^2.
\end{align}
In the context of sketching, the matrix $\boldS$ sketches the $d$-dimensional vector $u$ to a $b$ dimensional vector $\boldS u$. The above condition can be equivalently expressed as $\sup_{u\in\cU} | \|\boldS u\|_2^2 - \|u\|_2^2 | \leq \delta_r  \| u \|_2^2$. However, in many applications one encounters \emph{sketched bilinear forms} such as $(u^\top \boldS^\top \boldS \; v)$, for vectors $u \in \cU, v \in \cV$, 
and is required to provide uniform bounds for $\sup_{u\in\cU, v \in \cV} | u^\top\boldS^\top\boldS\; v - u^\top v |$.

For example in linear regression the response is given by $y_i =  \beta^\top \x_i + \eta_i$, where $\beta \in \cB \subseteq \R^d$ is the unknown parameter, $\x_i \in \cX \subseteq \R^d$ is the input and $\eta_i$ is some noise. In linear bandits \citep{abbasi2011improved, chu2011contextual} the reward is given by $r_i =  \theta_*^\top a_i + \eta_i$, $\theta_* \in \Theta \subseteq \R^d$ is the unknown parameter, $a_i \in \cA \subseteq \R^d$ is the chosen action and $\eta_i$ is a sub-gaussian noise. In high dimensional cases, i.e., when $d$ is very large, one might sketch both the unknown parameter and the input, and would naturally want to bound $|\theta_*^\top \boldS^\top \boldS a_i -  \theta_*^\top a_i |$. In Section~\ref{sec:app} and \ref{sec:extension}, we develop sketched variants of linear regression and linear bandit algorithms that leverage bilinear sketching of both the unknown parameter and the input vectors and yields sharper bounds on the error and regret respectively, which scale with the geometric complexity of the parameter and input sets—such as their Gaussian width (see Section~\ref{sec:prelim}) rather than the ambient dimension $d$. 

In distributed or federated learning, clients perform local gradient descent for multiple steps at every communication round, sketch their updates using a random matrix, and send the sketched updates to the server. The server aggregates and returns the sketched updates, which clients then de-sketch using the transpose of the sketching matrix to recover and apply the update to their local parameters \citep{Song_sketching,shrivastava2024sketching}. Obtaining finite-time convergence guarantees for such sketched algorithms requires controlling the deviation $\left| g^\top \boldS^\top \boldS h - g^\top h \right|$, where $g$ ranges over the set of possible loss gradients and $h$ over the eigenvectors of the loss Hessian (see Section~\ref{subsec:skDL}).

Although such bilinear forms have been studied, using a covering based analysis, that control bilinear forms via Dudley’s entropy integral (e.g., \citet{sarlos}, \citet{wood14}), in contrast, our approach uses generic chaining \citep{tala05,tala14}, yielding bounds in terms of $\gamma_2$ functional and Gaussian-width (see Section~\ref{sec:prelim}), which are strictly sharper on many sets than Dudley’s entropy (cf. \cite{tala05,tala14, vers18}), while recovering the classical finite-set rates.

A general framework to analyze such deviations is the \emph{random quadratic form} \citep{krahmer2014suprema,banerjee2019random}. Given a set of $m \times n$ matrices $\cA$ and a sub-gaussian random vector $\bxi \in \R^n$
\begin{align}
\label{eq:RQF_single}
    C_{\cA}(\bxi) &= \sup_{A \in \cA} \left| \|A \bxi\|_2^2 - \mathbb{E} \|A \bxi\|_2^2 \right| \notag\\
    &= \sup_{A \in \cA} \left| \bxi^\top A^\top A \bxi - \mathbb{E} \; \bxi^\top A^\top A \bxi \right|.
\end{align}
Note that the term $\sup_{u\in\cU} | \|\boldS u\|_2^2 - \|u\|_2^2 |$ can be equivalently expressed as in \eqref{eq:RQF_single} by converting the matrix $A$ into a vector $u = \text{vec}(A)$ and converting $\bxi$ into the matrix $\boldS$ (see Section~\ref{sec:JL-RIP} for details). In this work we develop deviation bounds for the following random variable with cross inner products:
\begin{align}
\label{eq:RQF_double}
    C_{\cM,\cN}(\bxi) = \!\!\! \sup_{M \in \cM, N \in \cN}\! \left| \bxi^\top M^\top N \bxi - \mathbb{E} \bxi^\top M^\top N \bxi \right|
\end{align}
and show that this yields sharp statistical guarantees for a variety of ML problems where such cross-terms naturally arise—particularly in settings involving bilinear sketches of the form $(u^\top \boldS^\top \boldS \; v)$.

We outline our \emph{main technical contributions} below.
\vspace{-1.5ex}
\begin{enumerate}[left=0em]
    \item \textbf{Uniform Deviation Bounds.} Our main result is a large deviation bound on $ C_{\cM,\cN}(\bxi)$ defined in \eqref{eq:RQF_double} (cf. Theorem~\ref{thm:cross_product}). This significantly generalizes existing results in \cite{krahmer2014suprema} that provide such bounds for $C_{\cA}(\bxi)$ in \eqref{eq:RQF_single}. Our approach relies on Generic Chaining \citep{tala05,tala14} and introduces new techniques for handling suprema over pairs of sets, which would be of independent interest (more details in  Section~\ref{sec:proof_main_result}). 
    One such result we develop is a \emph{double chaining bound} (Theorem~\ref{thm:doubleTree}) for stochastic processes indexed by pairs $(u,v) \in \cU \times \cV$. It provides a uniform deviation bound by chaining separately over $\cU$ and $\cV$, yielding a bound that scales with the sum of their $\gamma_2$ functional (see Remark~\ref{rem:doubleTree} for details and Section~\ref{sec:prelim} for definitions).
    
    \item \textbf{Applications.} This unified analysis recovers known results such as the J-L lemma as special cases, while also extending RIP-type guarantees to preserving inner products over arbitrary sets (see Proposition~\ref{prop:J-L} and \ref{prop:restricted_inner_prod} respectively). We also apply our results to obtain improved finite time convergence guarantees for sketched Federated Learning algorithms (see section~\ref{subsec:skDL}) and to obtain error bounds for linear regression that scale with the geometric complexity of the parameter and input sets rather than the ambient dimension (see Theorem~\ref{sec:app_proof_sumTtheorem}).
    \item \textbf{Sum of Random Quadratic Forms.} We extend our deviation bound to the random variable $C_{\cM,\cN}(\bxi^{1:T}) \triangleq \sup_{M \in \cM, N \in \cN} | \sum_{t=1}^{T} (\bxi_t^\top M^\top N \bxi_t) - \E(\bxi_t^\top M^\top N \bxi_t) |~,$ where $\bxi_t, t\in [T]$ are i.i.d., and show that the deviation scales by an extra $\sqrt{T}$ factor (see Theorem~\ref{theo:sumTtheorem}). This result is especially useful when the sketching matrix needs to change at every iteration, and is required in our bandit analysis. We note two things about this result. (1) A standard Hoeffding type argument does not apply here because of the $\sup$ and one has to extend the Generic chaining analysis. We provide a simpler expected analysis in Appendix~\ref{sec:sum_version_expected} to show how the scale $\sqrt{T}$ appears. (2) Such an analysis for sum of random quadratic forms does not exist even for the single set case as in \eqref{eq:RQF_single}.
    \item \textbf{Sketched Bandits.} We develop sketched versions of linear bandit algorithms, under assumptions on the action set - $\skLinUCB$ (sketched Linear UCB) and $\skLinTS$ (sketched Linear Thompson Sampling) and provide sharper regret bounds that depend on the geometric complexity of the action and parameter sets. 
    Our analysis introduces a novel way to decompose the regret into sum of the `regret in the sketched space' and a `restricted isometry term', and the latter is bounded using our new results (see Section~\ref{sec:extension} for details). We also empirically show that our sketched algorithms achieve lower regret and are significantly faster than their un-sketched counterparts when the true parameter has some structure (see Appendix~\ref{sec:app_bandits}).
\end{enumerate}
\vspace{-2ex}

\section{Preliminaries}
\label{sec:prelim}
\paragraph{Notation.} The notation \( a \lesssim b \) means that the inequality holds up to a multiplicative constant; that is, \( a \lesssim b \) implies the existence of a constant \( C > 0 \) such that \( a \leq C b \). The notations $y = \Theta(x)$ (respectively $y = O(x)$, $y = \Omega(x)$) imply there exist absolute constants $c_1, c_2, c_3, c_4$ such that $c_1 \cdot x \leq y \leq c_2 \cdot x$ (respectively $y \leq c_3 \cdot x$, $y \geq c_4 \cdot x$), and $\tilde{\Theta}(\cdot)$, $\tilde{\Omega}(\cdot)$ and $\tilde{O}(\cdot)$ notations hide the dependence on logarithm terms. We interchangeably use $x^\top y$ and $\langle x, y\rangle$ to denote the inner product.

\textbf{Random Quadratic Forms:} Quadratic forms involving random vectors arise naturally in high-dimensional statistics, machine learning, and random matrix theory. Consider $\|A\bxi\|_2^2 = \bxi^\top (A^\top A) \bxi$ where $\bxi = (\bxi_1, \ldots, \bxi_n) \in \mathbb{R}^n$ is a random vector and $A \in \mathbb{R}^{n \times n}$ is a fixed matrix. When $\bxi$ has independent coordinates, this expression is referred to as a \emph{random chaos} of order 2. A major result in this area is the \emph{Hanson-Wright inequality} \citep{hanson1971bound,vers18}, which gives exponential tail bounds for deviations of $ \|A\bxi\|_2^2$ from its mean.
\begin{theorem}[\textbf{Hanson–Wright}]
Let $\bxi = (\bxi_1, \ldots, \bxi_n) \in \mathbb{R}^n$ be a random vector with independent, mean-zero, sub-Gaussian entries. Let $A \in \mathbb{R}^{n \times n}$ be any fixed matrix. Then, $ \epsilon \geq 0$,
\begin{align*}
\mathbb{P} \Big( \big| \|A\bxi\|_2^2 &- \mathbb{E}[\|A\bxi\|_2^2] \big| \geq \epsilon \Big) \\
&\lesssim \exp \left( - c \min \left( \frac{\epsilon^2}{\|A^\top A\|_F^2}, \frac{\epsilon}{ \|A^\top A\|_{2 \rightarrow 2}} \right) \right),
\end{align*}
where $\|A\|_F$ is the Frobenius norm, and $\|A\|_{2 \rightarrow 2}$ is the operator norm of $A$. \qed
\end{theorem}
\vspace{-2ex}
However, many applications in high-dimensional statistics and signal processing, such as restricted isometry property (RIP) or sketching analysis, involve the supremum of such quadratic forms over a \emph{set} of matrices. That is, instead of bounding deviations of $\|A\bxi\|_2^2$ for fixed $A$, one is interested in bounding $\sup_{A \in \mathcal{A}} \left| \|A \boldsymbol{\xi}\|_2^2 - \mathbb{E} \|A \boldsymbol{\xi}\|_2^2 \right|,$ where $\mathcal{A}$ is a class of matrices. \cite{krahmer2014suprema} significantly generalize the Hanson-Wright result by analyzing such \emph{suprema of chaos processes}. They show using chaining arguments, one can control this supremum in terms of geometric complexity measures of $\mathcal{A}$. The results depend on two types of \emph{complexity measures} for the collection $\cA$. The first type consists of the radii of $\cA$ under different matrix norms and is defined below.
\begin{definition}[\textbf{Radii of $\cA$}]
 \label{def:radii}
    The radii of $\cA$ under the Frobenius norm and the operator norm, denoted by $d_F(\cA)$ and $d_{2 \to 2}(\cA)$, respectively. The Frobenius norm is given by $\|A\|_F = \sqrt{\mathrm{Tr}(A^\top A)}$, and the operator norm is defined as $\|A\|_{2 \to 2} = \sup_{\|\mathbf{x}\|_2 \leq 1} \|A \mathbf{x}\|_2$. For a given set $\cA$, we define:
    \vspace{-1ex}
    \begin{align*}
        d_F(\cA) &= \sup_{A \in \cA} \|A\|_F,\\
        d_{2 \to 2}(\cA) &= \sup_{A \in \cA} \|A\|_{2 \to 2}.
    \end{align*}
\end{definition}
\vspace{-2ex}
The second type of complexity measure involves Talagrand's $\gamma_2$ functional \citep{tala05,tala14}, written as $\gamma_2(\cA, \mathbf{d})$, with distance metric $\mathbf{d}$, and is defined as follows.
\begin{definition}[\textbf{Talagrand's $\gamma_2$ functional} \citep{tala05,tala14}]
\label{def:gamma_2}
    {For a metric space $(\cA, \mathbf{d})$, an admissible sequence of $\cA$ is a collection of subsets of $\cA$, $\{T_r(\cA) : r \geq 0\}$, such that for every $s \geq 1$, $|T_r(\cA)| \leq 2^{2^r}$ and $|T_0| = 1$. For $\beta \geq 1$, define}
    \vspace{-2ex}
    \begin{align}
        \gamma_\beta(\cA, d) = \inf \sup_{t \in \cA} \sum_{r=0}^{\infty} 2^{r/\beta} \mathbf{d}(t, T_r(\cA)),
        \label{eq:gamma2}
    \end{align}
    {where the infimum is taken with respect to all admissible sequences of $\cA$.} \qed
\end{definition}
The next theorem generalizes Hanson-Wright by giving deviation bound uniformly for all $A \in \cA$.
\begin{theorem}[Theorem 3.1, \citep{krahmer2014suprema}]
\label{thm:krahmerTheorem}
Let $\mathcal{A}$ be a set of matrices, and let $\boldsymbol{\xi}$ be a random vector whose entries $\xi_j$ are independent, mean-zero, unit variance, and $L$-subgaussian random variables. Set
\begin{align*}
{\color{Red}W} &{\color{Red} = \gamma_2(\mathcal{A}, \|\cdot\|_{2 \to 2}) \left( \gamma_2(\mathcal{A}, \|\cdot\|_{2 \to 2}) + d_F(\mathcal{A}) \right)} \\
& \qquad\qquad  {\color{Red} + d_F(\mathcal{A}) d_{2 \to 2}(\mathcal{A})}, \\
{\color{Blue}V} &{\color{Blue} = d_{2 \to 2}(\mathcal{A}) \left( \gamma_2(\mathcal{A}, \|\cdot\|_{2 \to 2}) + d_F(\mathcal{A}) \right)}, \;
{\color{Green}U = d_{2 \to 2}^2(\mathcal{A})}.
\end{align*}
Then, for $\epsilon > 0$,
\(
\mathbb{P} \left( \mathcal{C}_{\mathcal{A}}(\boldsymbol{\xi}) \geq c_1 {\color{Red}W} + \epsilon \right)
\leq 2 \exp\big(\!\!-c_2 \min( \frac{\epsilon^2}{\color{Blue}V^2}, \frac{\epsilon}{\color{Green}U})\big),
\)
\emph{where $c_1, c_2$ depend only on $L$.}
\end{theorem}
Finally, we define another geometric complexity of a set -- Gaussian width. Indeed \( \gamma_2(\mathcal{A}, \|\cdot\|_{2 \to 2}) \) can be upper bounded by the Gaussian width up to a constant factor \citep{tala14}.
\begin{definition}[\textbf{Gaussian Width}]
\label{def:gaussian_width}
Let $\cT \subseteq \mathbb{R}^d$ be a bounded subset. The \emph{Gaussian width} of $\cT$ is defined as $w(\cT) := \mathbb{E}_{g} \left[ \sup_{t \in \cT} \langle g, t \rangle \right]$, where $g \sim \mathcal{N}(0, I_d)$. Further let $\cA \subseteq \mathbb{R}^{m \times n}$ be a bounded set. The {Gaussian width} of $\cA$ is defined as $w(\cA) := \mathbb{E}_{G} \left[ \sup_{A \in \cA} \left| \operatorname{Tr}(G^\top A) \right| \right],$ where $G = [g_{i,j}] \in \mathbb{R}^{m \times n}$ is a random matrix with i.i.d.\ standard normal entries, i.e., $g_{i,j} \sim \mathcal{N}(0,1)$.
\end{definition}

\vspace{-1ex}

\section{Main Result}
\label{sec:main}
Let $\cM$ be a set of $(m \times n)$ matrices and $\cN$ be a set of $(n \times m)$ matrices. We make the following assumption on the stochastic process $\bxi$.

\begin{asmp}
\label{asmp:xi}
Suppose $\bxi$ be a random vector with independent coordinates $\bxi_i$, each of which is a mean zero L-subgaussian random variable.
\end{asmp}
\vspace{1ex}
We want to develop large deviation bound for $\Crv$ defined in \eqref{eq:RQF_single}.
Our analysis and results would rely on radii of $\cM,\cN$ under the Frobenius and operator norms, denoted by $d_F(\cdot)$ and $d_{2 \to 2}(\cdot)$, respectively (see Definition~\ref{def:radii}) and Talagrand's $\gamma_2$ functional (see Definition~\ref{def:gamma_2}).
\vspace{2ex}
\begin{theoremBox}[ (Uniform Deviation Bounds for Cross Inner Products) ]
{Let $\mathcal{\cM}$ and $\mathcal{\cN}$ be set of $(m\times n)$ and $(n\times m)$ matrices respectively, and let $\bxi$ be a random vector satisfying Assumption~\ref{asmp:xi}. We define}
\begin{align*}
    {\color{Red} W} &{\color{Red} = \gamma_2(\cM,\| \cdot \|_{2 \rightarrow 2}) \cdot \Big[\gamma_2(\cN, \| \cdot \|_{2 \rightarrow 2}) + d_F(\cN) ~\Big]} \\
    & \;\;\; {\color{Red} +\; \gamma_2(\cN,\| \cdot \|_{2 \rightarrow 2}) \cdot \Big[\gamma_2(\cM, \| \cdot \|_{2 \rightarrow 2}) + d_F(\cM)\Big]} \\
    {\color{Blue} V}& {\color{Blue}=\gamma_2(\cM,\| \cdot \|_{2 \rightarrow 2})d_{2 \rightarrow 2}(\cN) }\\
    & \;\;\;\;{\color{Blue}+\; \gamma_2(\cN,\| \cdot \|_{2 \rightarrow 2})d_{2 \rightarrow 2}(\cM)} \\
    & \;\;\;\;{\color{Blue} +\; \min \Big\{ d_F(\cM) \cdot d_{2 \rightarrow 2}(\cN), d_F(\cN) \cdot  d_{2 \rightarrow 2}(\cM) \!\Big\}} \\
    {\color{darkgreen} U} &{\color{darkgreen} = d_{2 \to 2}(\cM) \; d_{2 \to 2}(\cN).}
\end{align*}
{Then, for $\epsilon > 0$,} $\displaystyle \mathbb{P}\big(C_{\mathcal{\cM,\cN}}(\bxi) \geq c_1 {\color{Red} W} + \epsilon\big) 
\leq 2 \exp\left( -c_2 \min\left\{ \frac{\epsilon^2}{{\color{Blue} V^2}}, \frac{\epsilon}{{\color{darkgreen} U}} \right\} \right),$
\textit{where the constants $c_1, c_2$ depend on the sub-gaussian parameter $L$.}
\label{thm:cross_product}
\end{theoremBox}
\vspace{2ex}
\begin{remark}
    Comparing Theorem~\ref{thm:cross_product} and \ref{thm:krahmerTheorem} we observe that $W, V$ and $U$ in our case depend on product of geometric complexities of $\cM$ and $\cN$ and reduce to the terms in Theorem~\ref{thm:krahmerTheorem} when $M = N$. Interestingly the terms do not depend on the complexity of a joint set such as $\{M^\top N : M \in \cM, N \in \cN\}$, but rather on a symmetric combination of the individual complexities of $\cM$ and $\cN$, and is precisely because our analysis decouples them. This decoupled structure allows sharper bounds in applications where one set has small $\gamma_2$ functional or radii, even if the other is large.
\end{remark}

\subsection{Proof of Main Result}
\label{sec:proof_main_result}
We outline the main steps in the proof of Theorem~\ref{thm:cross_product} along with the novel aspects of our analysis here. A detailed version and proofs of intermediate technical results provided in Appendix~\ref{sec:app_proof}.

Our approach to getting a large deviation bound for $C_{\cM.\cN}(\bxi)$ defined in \eqref{eq:RQF_double} is based on bounding
$\| C_{\cM, \cN}(\bxi) \|_{L_p}$, where $\|X\|_{L_p}$ is the $L_p$ norm of $X$ defined as $\|X\|_{L_p} = \left( \mathbb{E}|X|^p \right)^{1/p}.$ Specifically, our objective is to show that $\| C_{\cM, \cN}(\bxi)\|_{L_p} \leq a + \sqrt{p} \cdot b + p \cdot c~,  \forall p \geq 1~$,
where $a,b,c$ are constants that do not depend on $p$, which using the moment-generating function and Markov's inequality \citep{will91,vers12} immediately implies $P(\left| C_{\cM, \cN}(\bxi) \right|  \geq a + u ) \leq \exp\left\{-\min\left(\frac{u^2}{4b^2}, \frac{u}{2c}\right)\right\}~.$

The proof consists of the following three main steps:
\begin{enumerate}[left=0em]
    \item \textbf{Decomposing $\Crv$ into off-diagonal and diagonal terms.} We define the following terms corresponding to the off-diagonal and diagonal terms of $M^\top N$ respectively:
    \begin{align*}
        B_{\cM, \cN}(\bxi) &\triangleq \sup_{\substack{M \in \mathcal{M} \\ N \in \mathcal{N}}} \Big| \sum_{\substack{j,k=1\\j \neq k}}^n \bxi_j \bxi_k \langle M_j, N_k \rangle \Big| ,\\
        D_{\cM, \cN}(\bxi) &\triangleq \sup_{\substack{M \in \mathcal{M} \\ N \in \mathcal{N}}} \Big| \sum_{j=1}^n (|\bxi_j|^2 - \E|\bxi_j|^2) \langle M_j, N_j \rangle \Big| ,
    \end{align*}
    where $M_i$ and $N_i$ are the $i$-th row of $M$ and $N$ respectively. The next lemma relates the $L_p$ norm of $\Crv$ to that of $B_{\cM, \cN}(\bxi)$ and $D_{\cM, \cN}(\bxi)$ using symmetrization \citep{vers18}.
    \begin{restatable}{lemma}{Decomposition}
    \label{lemm:Decomposition}
        For $B_{\cM, \cN}(\bxi)$ and $D_{\cM, \cN}(\bxi)$ as defined above, the following holds
        \begin{align}
            \|C_{\cM, \cN}(\bxi)\|_{L_p} \leq \|B_{\cM, \cN}(\bxi)\|_{L_p} + \|D_{\cM, \cN}(\bxi)\|_{L_p}
        \end{align}
    \end{restatable}
    \item\textbf{Bounding the off-diagonal term.} To bound the $L_p$ norm of $B_{\cM, \cN}(\bxi)$ we consider $\bxi'$ to be an independent copy of $\bxi$ and use symmetrization (eg. Lemma 6.3 \citep{leta91}) to get the following upper bound:
    \begin{align*}
        \| B_{\cM, \cN}(\bxi) \|_{L_p} & \leq  4 \big\| \sup_{M \in \cM, N \in \cN} \left| \langle M \bxi, N \bxi' \rangle \right| \big\|_{L_p}~.
    \end{align*}
    Next, unlike \cite{krahmer2014suprema} the inner product above contains two different matrices $M$ and $N$, and therefore we consider two separate admissible sequences (cf. definition ~\ref{def:gamma_2}) $\{T_r(\cM)\}_{r=0}^{\infty}$ and $\{T_r(\cN)\}_{r=0}^{\infty}$ of $\cM$ and $\cN$ respectively. We then use a generic chaining argument by creating two separate increment sequences for $\cM$ and $\cN$ to give the following bound:
        \begin{align*}
        & \left\|\! \sup_{M \in \cM, N \in \cN} \langle M \bxi, N \bxi' \rangle \right\|_{L_p} \!\!\leq \!\! \underbrace{\sup_{M \in \cM, N \in \cN} \Big\| \langle M \bxi, N \bxi' \rangle \Big\|_{L_p}}_{I} \\
        & + \gamma_2(\cM,  \|\cdot \|_{2 \rightarrow 2}) \cdot \Big\|\sup_{N \in \cN} \| N \bxi' \|_2 \Big\|_{L_p} ~ \\
        & \underbrace{\qquad \qquad \qquad+ ~\gamma_2(\cN,\|\cdot \|_{2 \rightarrow 2}) \cdot \Big\|\sup_{M \in \cM} \| M \bxi \|_2 \Big\|_{L_p}}_{II}
    \end{align*}
    In term \( I \), the supremum has been pulled outside the \( L_p \)-norm, which allows us to apply standard concentration bounds for sub-Gaussian random variables. Since \( \bxi \) and \( \bxi' \) are independent and sub-Gaussian, the quantity \( \| \langle M \bxi, N \bxi' \rangle \|_{L_p} \) can be bounded in terms of the operator norms of \( M \) and \( N \). The appearance of the cross-terms in term \( II \), arises specifically from the application of generic chaining to \(\langle M \bxi, N \bxi' \rangle\), where we create separate admissible sequences for \(\cM\) and \(\cN\). Specifically, the chaining decomposition of \(M\) contributes a sequence of approximations whose increments are paired with the worst-case realization over \(N\), and vice versa. Finally $\|\sup_{N \in \cN} \| N \bxi' \|_2 \|_{L_p}$ can be handled using existing techniques in \citep{krahmer2014suprema} to give the following upper bound on the off-diagonal term.
    \begin{restatable}{theorem}{theoremOffDiagonal}
    \label{theo:offDiagonalLpBound}
    Let $\bxi$ be a stochastic process satisfying Assumption~\ref{asmp:xi}. Then, for all $p \geq 1$, we have
    \begin{align*}
        &\| B_{\cM, \cN}(\bxi)\|_{L_p} \\
        &\lesssim \Bigg[{\color{Red}\gamma_2(\cM,\| \cdot \|_{2 \rightarrow 2})} \cdot \Big({\color{Red}\gamma_2(\cN, \| \cdot \|_{2 \rightarrow 2}) + d_F(\cN)} + \\&\qquad\qquad{\color{Blue}\sqrt{p}~ d_{2 \rightarrow 2}(\cN)}~\Big)\nonumber \\
        & + {\color{Red}\gamma_2(\cN,\| \cdot \|_{2 \rightarrow 2})} \cdot \Big({\color{Red}\gamma_2(\cM, \| \cdot \|_{2 \rightarrow 2}) + d_F(\cM)} + \\&\qquad\qquad{\color{Blue}\sqrt{p}~ d_{2 \rightarrow 2}(\cM)}~\Big) \nonumber
        \\ &  + {\color{Blue}~ \sqrt{p}~ \min \Big\{  d_F(\cM) \cdot d_{2 \rightarrow 2}(\cN),d_F(\cN) \cdot d_{2 \rightarrow 2}(\cM)} \Big\} \nonumber \\
        &+ {\color{Green}p \cdot d_{2 \rightarrow 2}(\cM) \cdot d_{2 \rightarrow 2}(\cN)}~.\Bigg]   
    \end{align*}
    \end{restatable}
    \item \textbf{Bounding the diagonal term.} We use symmetrization (\cite[Lemma 6.3]{leta91}) and contraction (\cite[Lemma~{4.6}]{leta91}) to get
\begin{align}
    \|& D_{\cM, \cN}(\bxi)) \|_{L_p}  \leq \underbrace{\left\| D_{\cM, \cN}(\g) \right\|_{L_p}}_{I} \notag\\
    &+ \quad \quad \underbrace{\left\| \sup_{M\in\cM, N\in\cN} \Big| \sum_{j=1}^n \eps_j   \langle M_j, N_j \rangle \Big| \right\|_{L_p}}_{II}~,
\label{eq:diagonal_1}
\end{align}
where $\g$ is a Gaussian random vector with independent entries $\g_j$ and $\{ \eps_j\}$ is a set of independent Rademacher variables independent of $\bxi$. Note that term $I$ is the same diagonal term with the Gaussian random vector $\g$ instead of the sub-gaussian random vector $\bxi$, and is arguably easier to bound. However, because of the cross terms $\langle M_j, N_j \rangle$, we cannot use a standard decoupling argument as in Theorem 2.5 in \cite{krahmer2014suprema}. Instead we we prove a stronger version of decoupling for Gaussian chaos that holds even for non-hermitian matrices to upper bound term $I$.

\vspace{1ex}
Bounding term $II$ is even more challenging. In the single set case \citep{krahmer2014suprema} the term inside the $\sup$ reduces to $\sum_{j}\varepsilon_j \|M_j\|_2^2$ and is a sub-gaussian process relative to the metric $d(A,B) = (\sum_{j}\|A_j\|_2^2 - \|B_j\|_2^2)^{1/2}$ and therefore a standard chaining argument follows (see Proof of Theorem 3.5 in \cite{krahmer2014suprema}). Since we cannot use such a technique, we device a new double tree argument for generic chaining in the following theorem. 
\begin{restatable}{theorem}{doubleTree}
\label{thm:doubleTree}
    Consider a stochastic process $\{ X_{(u,v)}\}$ where $u \in \mathcal{U}$, $v \in \mathcal{V}$, and let $d$ be the metric. Suppose for any $t > 0$, with probability at least $1 - c_0 \exp\left(-\frac{t^2}{2}\right)$, $\{ X_{(u,v)}\}$ satisfies 
    \begin{align*}
        | X_{(u_1, v)} - X_{(u_2, v)} | &\leq C_u t \cdot d(u_1, u_2), \forall v \in \cV\quad \\
        | X_{(u, v_1)} - X_{(u, v_2)} | &\leq C_v t \cdot d(v_1, v_2), \forall u \in \cU.
    \end{align*}
    Then, w.p. $1 - 2c_0 \exp\left( - \frac{t^2}{2}\right),$
        $\sup_{u \in \mathcal{U}, \, v \in \mathcal{V}} \left| X_{u,v} \right| \leq 4\sqrt{2}t\left(C_{u} \gamma_2(\mathcal{U}, d) + C_{v}\gamma_2(\mathcal{V}, d) \right).$
\end{restatable}
\begin{remark}[\textbf{Double Chaining Bound}]
\label{rem:doubleTree}
    In Theorem~\ref{thm:doubleTree} we consider a double indexed stochastic process $X$ indexed by elements from sets $\cU$ and $\cV$. We assume that the process varies smoothly in each coordinate, i.e., for a fixed $v$, the process is Lipschitz in $u$, and vice versa, (with high probability) and provide a uniform deviation bound on the suprema of the double indexed process that scales with the sum of $\gamma_2$ functional of $\cU$ and $\cV$ . The proof idea involves applying Generic Chaining by creating two separate chains (or trees) over $\cU$ and $\cV$ and relating it to the displacement in both the indices.
\end{remark}
The final result follows by defining the stochastic process $ X_{(M,N)} = \left| \sum_{j=1}^n \g_j \langle M_j, N_j \rangle \right|$ and invoking Theorem~\ref{thm:doubleTree} and is as given below.
\begin{restatable}{theorem}{theoremDiagonalbound} 
\label{theo:diagonalLpBound}
Let $\bxi$ be a stochastic process satisfying Assumption~\ref{asmp:xi}. Then, for all $p \geq 1$, we have
    \begin{align*}
    &\left\| D_{\cM,\cN}(\bxi) \right\|_{L_p} \\ &\lesssim \Bigg[{\color{Red}\gamma_2(\cM,\| \cdot \|_{2 \rightarrow 2})} \cdot \Big({\color{Red}\gamma_2(\cN, \| \cdot \|_{2 \rightarrow 2}) + d_F(\cN)}  \\ & \qquad\qquad + {\color{Blue}\sqrt{p}~ d_{2 \rightarrow 2}(\cN)}~\Big)\nonumber \\
        & + {\color{Red}\gamma_2(\cN,\| \cdot \|_{2 \rightarrow 2})} \cdot \Big({\color{Red}\gamma_2(\cM, \| \cdot \|_{2 \rightarrow 2}) + d_F(\cM)}  \\ & \qquad\qquad +{\color{Blue} \sqrt{p}~ d_{2 \rightarrow 2}(\cM)}~\Big) \nonumber
        \\ & + {\color{Blue}~ \sqrt{p}~ \min \Big\{  d_F(\cM) \cdot d_{2 \rightarrow 2}(\cN),d_F(\cN) \cdot d_{2 \rightarrow 2}(\cM)} \Big\} \nonumber
        \\ & {\color{Green} + p \cdot d_{2 \rightarrow 2}(\cM) \cdot d_{2 \rightarrow 2}(\cN)}~.\Bigg]    
    \end{align*}
\end{restatable}
\end{enumerate}
The proof of Theorem~\ref{thm:cross_product} now follows by combining the three steps (namely Lemma~\ref{lemm:Decomposition}, Theorem~\ref{theo:offDiagonalLpBound} and Theorem~\ref{theo:diagonalLpBound}) and collecting terms corresponding to ${\color{Red}W}, {\color{Blue}V}$ and ${\color{Green}U}$.


\section{Applications}
\label{sec:app}
\subsection{John-Lindenstrauss Lemma}
\label{subsec:JL}
We show how Theorem~\ref{thm:cross_product} can be used to recover the J-L lemma \citep{joli84}. The JLT is a randomized embedding that maps high-dimensional vectors into a lower-dimensional space while approximately preserving pairwise Euclidean distances~\citep{CW13,wood14}.
Let \( X \in \mathbb{R}^{n \times p} \), with \( n < p \), and let \( \mathcal{A} \) be any set of \( N \) vectors in \( \mathbb{R}^p \). We say that \( X \) satisfies the \emph{Johnson--Lindenstrauss Transform (JLT)} if for every \( \varepsilon > 0 \),
\begin{equation}
    (1 - \varepsilon)\|u\|_2^2 \leq \|Xu\|_2^2 \leq (1 + \varepsilon)\|u\|_2^2 \quad \text{for all } u \in \mathcal{A}.
\end{equation}
In Appendix~\ref{sec:JL-RIP} we show that with appropriate choices of the matrices $M$ and $N$ we ensure that a random matrix $\frac{1}{\sqrt{n}}\tilde{X}$ whose entries are i.i.d. standard gaussian would satisfy the JLT condition.
\begin{proposition}
\label{prop:J-L}
    Let \( X \in \mathbb{R}^{n \times p} \) be a matrix defined as \( X = \frac{1}{\sqrt{n}} \tilde{X} \), where the entries of \( \tilde{X} \) are i.i.d.\ standard normal. If we choose \( n = \Omega(\varepsilon^{-2} \log N) \), then \( X \) is a Johnson--Lindenstrauss Transform (JLT) with probability at least \( 1 - \frac{1}{N^c} \), for some constant \( c > 0 \).
\end{proposition}

\subsection{Restricted Isometry Property}
\label{subsec:RIP}
A closely related property to the J-L lemma is the \emph{Restricted Isometry Property}. Let $X \in \mathbb{R}^{n \times p}$ and let $\mathcal{A}$ denote the collection of all $s$-sparse vectors in $\mathbb{R}^p$. We say that $X$ satisfies the RIP with constant $\delta_s \in (0,1)$ if, for all $u \in \mathcal{A}$,
\begin{align}
    (1 - \delta_s)\|u\|_2^2 \leq \|Xu\|_2^2 \leq (1 + \delta_s)\|u\|_2^2.
\end{align}
Such matrices are of fundamental importance in high-dimensional statistics and compressed sensing, where the objective is to recover a sparse signal $\theta^* \in \mathbb{R}^p$ from a limited number of noisy linear measurements \citep{plve13,krahmer2014suprema}. We can extend this to preserving inner products as follows. 

Let $\mathcal{U}, \mathcal{V} \subseteq \mathbb{R}^p$ be two subsets. Consider the following condition: for every $\epsilon > 0$ we have
\begin{align}
\label{eq:restricted_inner_prod}
     |\langle Xu, Xv \rangle - \langle u, v \rangle| \leq  \epsilon \|u\|_2 \|v\|_2 \quad \text{for all } u \in \cU, v \in \cV.
\end{align}
For finite sets with $f = |\cU| = |\cV|$, \cite{wood14} shows that \( X \in \mathbb{R}^{n \times p} \) be a matrix defined as \( X = \frac{1}{\sqrt{n}} \tilde{X} \), where the entries of \( \tilde{X} \) are i.i.d.\ standard normal satisfies \eqref{eq:restricted_inner_prod} for $n = \Omega(\varepsilon^{-2} \log(f/\delta))$ (see Theorem 4 in \cite{wood14}). Their analysis uses a standard concentration plus a union bound. We can significantly generalize this result using Theorem~\ref{thm:cross_product}, by considering arbitrary sets $\cU, \cV$ in the following proposition (see Appendix~\ref{sec:JL-RIP} for the proof).
\begin{proposition}
\label{prop:restricted_inner_prod}
    Consider arbitrary sets $\cU$ and $\cV$ such that $\|u\|_2 = \|v\|_2 = 1,$ for all $u \in \cU, v \in \cV$ and let \( X \in \mathbb{R}^{n \times p} \) be a matrix defined as \( X = \frac{1}{\sqrt{n}} \tilde{X} \), where the entries of \( \tilde{X} \) are i.i.d.\ standard normal. Then for every $0<\epsilon\leq 1$, if $ n = \Omega(\frac{1}{\epsilon^2} \big(\omega(\cU) + \omega(\cV))^2\big)$, then \eqref{eq:restricted_inner_prod} holds with probability $1-e^{-c (\omega(\cU) + \omega(\cV))^2}$ for constant $c>0$, where $\omega(\cT)$ is the Gaussian width of $\cT$ (cf. Definition~\ref{def:gaussian_width}) 
\end{proposition}
For infinite sets, a standard approach is to discretize $\cU, \cV$ via multiscale coverings and control $\sup_{u\in\cU,\,v\in\cV} |\langle Xu, Xv\rangle - \langle u, v\rangle|$ by concentration plus a union bound on the nets, yielding Dudley-type complexities through the entropy integral $\mathcal{D}(\cT) = \int_{0}^{\mathrm{diam}(\cT)} \sqrt{\log N(\cT,\tau)} \, \mathrm{d}\tau$ (cf. \citet{sarlos}; \citet{wood14}). Note that, $\omega(\cT) \asymp \gamma_{2}(\cT, \| \cdot \|_{2})$ (see \cite{tala14}), whereas $\mathcal{D}(\cT)$ is an upper bound for $\gamma_{2}$, and there exist sets where $\mathcal{D}(\cT)$ is not sharp \citep{tala05,tala14}. On such sets, our Gaussian-width sample complexity $n \gtrsim \varepsilon^{-2}(\omega(\cU) + \omega(\cV))^{2}$ is strictly sharper, while on finite sets we recover $\omega(\cT) \asymp \sqrt{\log |\cT|}$ and thus match the classical rates of \citet{wood14} and the cases analyzed by \citet{sarlos}.

\subsection{Sketching-based Distributed Learning}
\label{subsec:skDL}
We consider the setup of Sketching-based Distributed Learning~(sketch-DL) as described in~\citep{shrivastava2024sketching,Song_sketching}.  For a complete treatment, we refer the reader to the Appendix~\ref{sec:app_distributed}. 

Consider a distributed learning framework with  $C$ clients. Each client $c \in C$ has access to a local dataset $\mathcal{D}_{c}\coloneqq\{x_{i,c},y_{i,c}\}_{i=1}^{N}$. The goal is to learn $\theta \in \R^{p}$ that minimises the joint loss function:
$
\cL(\theta) = \frac{1}{C}\sum_{c=1}^{C} \cL_{c}(\theta)
$, where $\cL_{c}(\theta) = \frac{1}{N}\sum_{i=1}^{n} \ell(f(\theta,\x_{i,c}),y_{i,c})$, with $\ell(\cdot,\cdot)$ being some loss function and $f$ being the output of a deep learning model. 

The sketch-DL algorithm (see Appendix~\ref{sec:app_distributed}) proceeds as follows: in each round, every client computes local gradients and sends their \textit{sketched} (\texttt{sk}) versions to the server. The server aggregates these \texttt{sk}-gradients and broadcasts the result. Each client then \textit{desketches} (\texttt{desk}) the aggregated gradient and uses it to update its local model. The \texttt{sk} and \texttt{desk} operations reduce communication by mapping $p$-dimensional gradients to a lower $b$-dimensional space (with $b \ll p$) via a shared sketching matrix. The sketch de-sketch operations are $\texttt{sk}(\x): \R^{p} \rightarrow \R^{b} \coloneqq R \x , \texttt{desk}(\x): \R^{b} \rightarrow \R^{p} \coloneqq R^{\top} \x $.

The objective is to provide provide finite time convergence guarantee by bounding the loss difference $\mathcal{L}(\theta_{T}) - \mathcal{L}(\theta^{*})$ after $T$ rounds, where $\theta^* = \argmin \cL(\theta)$. Towards this we provide the following theorem. The precise statement along with the proof and all details can be found in Appendix~\ref{sec:app_distributed}.

We make the following assumptions on the loss function $\cL$ and the loss Hessian $\mathbf{H}$:
\begin{asmp}[\textbf{PL condition}]\label{assump:pl}
The loss function $\mathcal{L}(\cdot)$ satisfies the Polyak--Łojasiewicz (PL) condition with constant~$\mu$, i.e. $\norm{\nabla \cL(\w)}^2 \ge 2\mu (\cL(\w) - \cL(\w_*))$, where $\w_* = \argmin_{\w} \cL(\w)$.
\end{asmp}
\begin{asmp}[\textbf{Anistropic Loss Hessian}]\label{assump:hessian-spectrum}
For any loss Hessian $H_{i,c,t}$, assume there exists a fixed positive definite $\mathbf{H}$ such that $-\mathbf{H}\preceq H_{i,c,t}\preceq \mathbf{H}$. Let $\Lambda_{1},\;\dots,\;\Lambda_{p}$ be the eigenvalues of $\mathbf{H}$
and define $\Lambda_{\max} =\;\max_j \bigl|\Lambda_{j}\bigr|$.
Then there exists a constant $\kappa = \mathcal{O}(1)$ such that
$\sum_{j=1}^p \bigl|\Lambda_{j}\bigr|
\;\le\;
\kappa\,\Lambda_{\max}$.
\end{asmp}
With the above assumptions, we can state the convergence guarantee of K-step sketch-DL as:
\begin{restatable}[]{theorem}{federated}
    \label{thm:federated}
Suppose Assumptions~\ref{assump:pl} and~\ref{assump:hessian-spectrum} hold and let Let $\|\nabla_{\theta}\ell(\theta)\|\le G$. For suitable constant $\varepsilon<1,$ width of the network $m$, learning rate $\eta$, sketching dimension $b \;=\;\Omega\Bigl(\tfrac{1} {\varepsilon^{2}}\;\mathrm{polylog}\!\bigl(\tfrac{T\,N\,p^{2}}{\delta}\bigr)\Bigr),$ and $C_{2}(m,\kappa):= \mathcal{O}\Bigl(\frac{\varepsilon\kappa}{\sqrt{m}}+\frac{1}{\sqrt{m}}\Bigr),$ with probability at least $1-\delta$, we have 
\begin{align}
\mathcal{L}(\theta_{T}) & - \mathcal{L}(\theta^{*})
\le \bigl(\mathcal{L}(\theta_{0}) - \mathcal{L}(\theta^{*})\bigr)\,
e^{-2\mu\eta KT} \notag \\
&+
\frac{\left(\eta KC_{2}(m,\kappa)+C_{3}\right)\left(G^{2}+\epsilon^{2}\right)}{2\mu}\,.
\end{align}
\end{restatable}
\begin{remark}
Our bound relies on Gaussian width of two sets: the predictor gradients \( \mathcal{G} \coloneqq \{\nabla f(\theta_{c,i,t})\} \) and the loss-Hessian eigenvectors \( \mathcal{H} \coloneqq \{v_j\} \). Using Theorem~\ref{thm:cross_product} we bound $\sup_{g\in\mathcal{G},h\in\mathcal{H}}\left|g^{\top}R^{\top}Rh-g^{\top}h\right|\lesssim Z(\mathcal{G},\mathcal{H},\delta)$ in the analysis, where $Z(\mathcal{G},\mathcal{H},\delta)$ is a function of geometric complexities of $\cG$ and $\cH$ (see Appendix~\ref{sec:app_distributed} for details).By combining these with the gaussian-width estimates of \citet{banerjee2024loss}, we obtain the stated inequality. In contrast to \citet{shrivastava2024sketching}, who introduce a fresh sketching matrix at every iteration to avoid dependence with the gradient set, our approach allows reusing a single sketching matrix throughout. This is permissible because we bound the error directly via the Gaussian width of the  predictor-gradient set, which falls outside the finite-set framework employed in their analysis.
\end{remark}

\subsection{Linear Regression with Sketching}
\label{subsec:skLR}
We observe $n$ i.i.d. samples $(\x_i, y_i)$ from the linear model $y_i = \x_i^\top \beta^* + \varepsilon_i$, where $\x_i \in \mathbb{R}^d$ are sub-Gaussian covariates with covariance $\Sigma$ and $\varepsilon_i$ are independent sub-Gaussian noise terms. Further $\beta \in \cB \subseteq \R^d$. 
We solve a least squares problem to compute $\hat{\beta}^s$ using the sketched inputs and corresponding responses, i.e., using $(\mathbf{S}\x_i,y_i)$. Subsequently we de-sketch the estimate using $\mathbf{S}^\top \hat{\beta}^s$ and use this to make predictions. Note that solving the least squares problem in a lower dimensional sketched space is computationally faster. Our goal is to bound the error $(\mathbf{S}^\top \hat{\beta}^s - \beta^*)^\top \x$, for $\x \in \cX$. In Appendix~\ref{sec:regression_app} we show that one can provide error bounds that do not depend on the ambient dimension $d$ but on geometric complexities of $\cB$ and $\cX$.

\section{Sum of Random Quadratic Forms}
\label{sec:extension}
In this section we consider sum of random quadratic forms. Specifically, we want to develop large deviation bound for the following random variable.
\begin{equation}
    C_{\cM,\cN}(\bxi^{1:T})\triangleq \sup_{M \in \cM, N \in \cN} \big| \sum_{t=1}^{T} (\bxi_t^\top M^\top N \bxi_t) - \E(\bxi_t^\top M^\top N \bxi_t) \big|~ 
    \label{eq:caxi} 
\end{equation}
\begin{restatable}{theorem}{sumTtheorem}
\label{theo:sumTtheorem}
    Let $\mathcal{\cM}$ and $\mathcal{\cN}$ be set of $(m\times n)$ and $(n\times m)$ matrices respectively, and $\bxi_t$, $t \in [T]$ be i.i.d. random vectors satisfying Assumption~\ref{asmp:xi}. Further let $W, V$ and $U$ be as defined in Theorem~\ref{thm:cross_product}; then, for $\epsilon > 0$, $\mathbb{P}\left(C_{\cM,\cN}(\bxi^{1:T}) \geq c_1 {W}\sqrt{T} + \epsilon\right) \leq \exp\left( -c_2 \min\left\{ \frac{\epsilon^2}{{ T V^2}}, \frac{\epsilon}{{\sqrt{T}\;U}} \right\} \right),$
where the constants $c_1, c_2$ depend only on the sub-gaussian parameter $L$.
\end{restatable}
\subsection{Sketched Contextual Bandits}
In this section, we present a simplified bandit setup to illustrate a concrete setting where Theorem~\ref{theo:sumTtheorem} naturally applies. We briefly describe our results here and have pushed detailed descriptions to Appendx~\ref{sec:app_bandits}. In a bandit problem \cite{auer,lattimore_szepesvári_2020} a learner needs to make sequential decisions over $T$ time steps.
We consider a simplified setup where at any round $t \in [T]$, the learner picks an action $a_t$ from the action set $\cA$, and then the associated reward of the arm $r(a_t) \in [0,1]$ is observed. We make the following assumption on the reward.
\begin{asmp}[\textbf{Linear Reward}]
The reward $r(a_t)$ is given by
\(
r(a_t) = \langle a_t, \theta^* \rangle + \eta_t,
\) 
where $\theta^* \in \Theta^*$ is an unknown parameter vector and $\eta_t$ is a conditionally sub-Gaussian noise, i.e., $\forall \lambda \in \bbR, \E[e^{\lambda \eta_t} | a_1,\ldots a_t, \eta_1,\ldots,\eta_{t-1}] \leq \exp(\frac{\lambda^2}{2})$.
Further $\|a\|_2 = 1$ for all $a \in \cA$.
\label{asmp:realizability_bandit}
\end{asmp}

\begin{definition}[\textbf{Regret}] For actions $a_t, t\in[T]$, the learner wants to minimize the regret defined as
\begin{align}
\reg(T) \!& =\! \E\Big[\sum_{t=1}^T \left( r(a^*) \!-\! r(a_t) \right)\Big] \nonumber \!=\! \sum_{t=1}^T \langle \theta^*, a^* \rangle \!-\! \langle \theta^*, a_t \rangle \;,
\end{align}
where $a^* = {\argmax_{a \in \cA}} \; \langle \theta^*, a \rangle $ .
\end{definition}
\begin{algorithm}[t]
\caption{$\skLinUCB$\; (Sketched Linear UCB)}
\begin{algorithmic}[1]
\FOR {$t = 1, 2, ..., T$} 
\STATE Solve the least squares regression problem:
\vspace{-1.5ex}
\begin{align}
    \!\!\!\!\!\!\hat{\theta}^{s}_{t} = \!\! \underset{\theta \in \R^b, \|\theta\|_{2} \leq 1}{\argmin} \!\sum_{i=1}^{t-1} \left( \langle \theta , \sketch_t a_i \rangle - r_{i}\right)^2 + \lambda \|\theta\|_2
    \label{eq:theta-hat}
\end{align}
\vspace{-2ex}
\STATE Construct the Confidence set $\cC^s_t$ as in \eqref{eq:confidence}.
\STATE Compute the the optimistic estimates: $(\tilde{\theta}^s_t, a^s_t) = \argmax_{\theta \in \cC_{t}^s, a \in \sketch_t \cA} \langle \theta, a \rangle$
\STATE De-sketch and play the action $a_t = \sketch_t^\top a^{{s}}_t$; observe the reward $r_t$.
\ENDFOR
\end{algorithmic}
\label{algo:sketch-LinUCB}
\end{algorithm}

We develop a sketched version of the popular algorithm LinUCB (Linear Upper Confidence Bound \citep{abbasi2011improved}) and is summarized in Algorithm~\ref{algo:sketch-LinUCB}. At every round $t$ the learner sketches the inputs using $\sketch_t \R^{b \times d}$ whose entries are drawn i.i.d. from $N(0,1/b)$. It then solves a regularized least-squares problem (cf \eqref{eq:theta-hat} in Algorithm~\ref{algo:sketch-LinUCB}) in the $b$-dimensional sketched space to obtain an estimate \(\hat{\theta}_t^s\) of the unknown parameter. Subsequently, with $\bar{V}^{s}_{t} = \sum_{i=1}^{t} a_{i}^{s} {a_{i}^{s}}^\top + \lambda I$, the learner constructs a confidence set \(\cC^s_t\) around the estimate:
\begin{align}
&\cC_t^s = \Bigg\{ 
        \theta \in \bbr^b : \|\hat{\theta}^{s}_t - \sketch_t \theta^*\|_{\bar{V}^s_{t}} \notag
        \\& \!\!\!\le\!
        \sqrt{%
            \log\!\Bigl(
                \frac{\det(\bar{V}^s_{t})^{1/2}\,\det(\lambda I)^{-1/2}}{\delta}
            \Bigr)}
        \!+\! \lambda^{1/2} \|\sketch_t \theta_*\|_2 \!\!
      \Bigg\}
      \label{eq:confidence}
\end{align}
The algorithm then uses this confidence set to compute an \emph{optimistic} parameter and action pair \(\bigl(\tilde{\theta}^s_t, a^s_t\bigr)\) (Line~4) where $\tilde{\theta}^s_t \in \cC_t^s$ and the action is in the \emph{sketched} action space $a \in \sketch_t \cA$. 
Once this sketched action is identified, it is ``\emph{de-sketched}'' (Line~5) to recover the corresponding action \(a_t\) in the original space. Finally, the chosen action \(a_t\) is played, and the observed reward \(r_t\) is used in subsequent rounds to refine future estimates. Our primary result in this section is the following decomposition for the regret.

\begin{tcolorbox}[
  width=0.5 \textwidth,
  colback=gray!1,        
  colframe=black,         
  arc=5pt,               
  boxrule=1pt,            
  left=5pt, right=5pt,  
  top=5pt,  bottom=2pt  
]
    \begin{restatable}[\textbf{(Informal) Regret Decomposition for $\skLinUCB$}]{theorem}{ucb}
        Suppose Assumption~\ref{asmp:realizability_bandit} holds and the de-sketched actions selected by Algorithm~\ref{algo:sketch-LinUCB} are in $\cA$. Then with high probability
        \begin{align*}
            \reg(T)\! &= \underbrace{\tilde{O} \Bigg(\sqrt{bT}\bigg[\sqrt{b} + \frac{1}{\sqrt{b}} \omega(\Theta_*)\bigg]\Bigg)}_{I} \\ &+ \underbrace{\sum_{t=1}^{T} {\theta^*}^\top (I - \sketch_t^\top \sketch_t)a^*}_{II}
        \end{align*}
        where $\omega(\Theta_*)$, is the Gaussian width of the set $\Theta_*$ (cf. Definition~\ref{def:gaussian_width}).
        \label{thm:sk-LinUCB}
    \end{restatable}
\end{tcolorbox}
\begin{remark}
    Term $I$ captures the regret in the sketched $b$ dimensional space while term $II$ captures the restricted isometry term due to random sketching. Term $II$ is exactly in the bilinear form that involves a sum of $T$ independent sketch matrices. Therefore with suitable choices of $\cM$ and $\cN$ we can use Theorem~\ref{theo:sumTtheorem} to bound term $II$. We also develop a similar sketched version of Thompson sampling. We report the full results and associated details in Appendix~\ref{sec:app_bandits}.
\end{remark}

\section{Conclusion}
We presented a unified theory for analyzing uniform deviation bounds of \emph{sketched bilinear forms}, extending classical results on random quadratic forms to the setting of cross-inner products over pairs of structured sets. Our approach introduces new chaining techniques for controlling suprema over product spaces, and provides tight control in terms of the geometric complexity of the underlying sets. We also developed a uniform bound for the sum of random quadratic forms with i.i.d. random vectors, showing that the deviation scales as $\sqrt{T}$; notably, such a bound does not exist even for the single-set case considered in prior works. We apply our results on several ML problems and derive improved bounds. Our work highlights several scenarios in which sketched bilinear forms arise in modern ML algorithms and opens up several future directions, including extensions to adaptive sketches, analysis of non-linear predictors.

\bibliography{references}
\bibliographystyle{plainnat}


\clearpage
\appendix
\thispagestyle{empty}

\onecolumn
\aistatstitle{Beyond Johnson-Lindenstrauss:\\ Uniform Bounds for Sketched Bilinear Forms}

\section{Related Works}
\label{sec:related}
\vspace{-1ex}
\textbf{Sketching.} Sketching is a key dimensionality reduction technique with applications in machine learning, including federated learning \citep{ivkin2020communicationefficientdistributedsgdsketching} and low-rank approximation \citep{Tropp_2017}. The linearity property of sketching makes it an attractive tool in communication efficient training~\citep{Song_sketching, 10.1145/3183713.3196894}. Matrix sketching techniques have also been applied to linear bandits to improve computational efficiency. using  online sketching technique,~\citet{Sketching_nicolo} reduce the update time for OFUL and Thompson Sampling from \(O(d^2)\) to \(O(md)\). Randomized sketching has become a unifying primitive for fast numerical linear-algebra.  \citet{CW13} first showed that a sparse subspace embedding lets one obtain $(1+\varepsilon)$ low-rank approximation and over-constrained least-squares solvers in time nearly proportional to $\mathrm{nnz}(A)$.  Their idea was sharpened in several directions: OSNAP matrices of \citet{NN13} make the embedding \emph{ultra-sparse} (constant non-zeros per column) while guaranteeing subspace preservation; \citet{MM13} proved that the same input-sparsity bounds extend to low-distortion embeddings for \emph{robust} regression; and \citet{BW14} established optimal relative-error CUR decompositions within the same computational budget.  Follow-up work moved beyond $\ell_2$ objectives—\citet{SWZ17} give the first entrywise $\ell_1$ low-rank approximation—and beyond the centralized RAM model: \citet{WZ16} develop communication-optimal sketches for distributed low-rank approximation of implicit matrix functions, while \citet{BWZ16} match information-theoretic limits for distributed and streaming PCA.

\citet{SWZ19} extended sketching guarantees to tensor low-rank approximation, opening the door to high-order data analysis.  In data mining and clustering, sketch-based algorithms accelerate density-level-set estimation (\citealp{EMZ21}) and yield nearly-optimal coresets for $(k,z)$-clustering (\citealp{DSWY22}).  Sketching has also entered decision-making: \citet{WZD+20} embed planning with general objective functions, and \citet{SSX23} design value-iteration routines for linear MDPs whose runtime is sublinear in the action space.  On the optimization side, leverage-score maintenance lets \citet{LSZ19} solve empirical-risk-minimization problems in \emph{matrix-multiplication} time; fast inverse maintenance from \citet{JSWZ21} accelerates interior-point and cutting-plane methods, complemented by the oblivious central-path sketch of \citet{SY21} and the multi-layered cutting-plane scheme of \citet{JLSW20}.  Finally, \citet{QSZZ23} provide an online, unified framework for projection matrix–vector multiplication, yielding faster online ERM with either data-oblivious or adaptive sketches.%

 \textbf{Contextual Bandits.} Contextual bandits generalize multi-armed bandits by incorporating feature vectors, allowing decisions to be conditioned on context. The problem has been widely studied under linear payoffs \citep{chu2011contextual,abbasi2011improved,bandit_bake,lattimore_szepesvári_2020}. Linear stochastic bandits have been extensively studied as a fundamental setting for decision-making under uncertainty. \citet{10.1023/A:1013689704352} introduced an early upper confidence bound (UCB) algorithm for linear bandits, with subsequent improvements in regret bounds by \citet{Dani2008StochasticLO} and \citet{10.1287/moor.1100.0446}. The widely used OFUL (Optimism in the Face of Uncertainty Learning) algorithm (also called LinUCB) \cite{abbasi2011improved} provides an \(\mathcal{O}(d\sqrt{T})\) regret under the assumption that the action and parameter sets are \(d\)-dimensional Euclidean balls. \citet{agrawal2013thompson} extended Thompson Sampling to contextual bandits with linear payoffs, proving a regret bound of \(\tilde{O}(d^{3/2}\sqrt{T})\). More recently, neural contextual bandits have been proposed to handle non-linearity in reward functions. \citet{zhou2020neural_ucb} use neural networks with UCB-based exploration and achieve near-optimal \(\tilde{O}(\sqrt{T})\) regret. \citet{ban2022eenet} introduce a novel exploration strategy using an additional neural network to estimate potential gains, outperforming traditional linear bandit baselines, while \cite{deb2024contextual,deb2025conservative} extend the inverse gap weighting idea with neural networks.

\section{Detailed proof of Main Result (Theorem~\ref{thm:cross_product})}
\label{sec:app_proof}
To develop large deviation bounds on $C_{\cM, \cN}(\bxi)$, we decompose the quadratic form into terms depending on the off-diagonal and the diagonal elements of $M^\top N$ respectively as follows.
    \begin{align*}
        B_{\cM, \cN}(\bxi) ~ & \triangleq ~\sup_{M\in\cM, N\in\cN} \Big| \sum_{\substack{j,k=1\\j \neq k}}^n \bxi_j \bxi_k \langle M_j, N_k \rangle \Big| ~,
    \end{align*}
    \begin{align*}
        D_{\cM, \cN}(\bxi) ~ & \triangleq ~\sup_{M\in\cM, N\in\cN} \left| \sum_{j=1}^n (|\bxi_j|^2 - \E|\bxi_j|^2) \langle M_j, N_j \rangle \right| ~,
    \end{align*}
    where $M_i$ and $N_i$ are the $i$-th row of $M$ and $N$ respectively.

Our approach to getting a large deviation bound for $C_{\cM,\cN}(\bxi)$ is based on bounding
$\| C_{\cM, \cN}(\bxi) \|_{L_p}$, where  where for a random variable $X$, the $L_p$ norm is defined as:
\begin{align*}
\|X\|_{L_p} = \left( \mathbb{E}|X|^p \right)^{1/p}.
\end{align*} 
This in turn is based on bounding $\| B_{\cM, \cN}(\bxi) \|_{L_p}$ and $\| D_{\cM, \cN}(\bxi) \|_{L_p}$. Such bounds lead to a bound on $\| C_{\cM, \cN}(\bxi)\|_{L_p}$ of the form
\begin{equation}
    \| C_{\cM, \cN}(\bxi)\|_{L_p} \leq {\color{Red}W} + \sqrt{p} \cdot {\color{Blue}V} + p \cdot {\color{Green}U}~, \quad \forall p \geq 1~,
\end{equation}
where $a,b,c$ are constants which do not depend on $p$.
Note that by using the moment-generating function and Markov's inequality \citep{will91,vers12}, these $L_p$-norm bounds imply, for all $\epsilon >0$
\begin{equation}
    P(\left| C_{\cM, \cN}(\bxi) \right|  \geq {\color{Red}W} + {\color{Blue}V} \cdot \sqrt{\epsilon} + {\color{Green}U} \cdot \epsilon ) \leq e^{-\epsilon}~,
    \label{eq:mark1}
\end{equation}
or, equivalently 
\begin{equation}
    P(\left| C_{\cM, \cN}(\bxi) \right|  \geq {\color{Red}W} + \epsilon ) \leq \exp\left\{-\min\left(\frac{\epsilon^2}{4{\color{Blue}V}^2}, \frac{u}{2{\color{Green}U}}\right)\right\}~,
    \label{eq:mark2}
\end{equation}
which yields the desired large deviation bound. 

The analysis for bounding the $L_p$ norms of $C_{\cM, \cN}(\bxi)$ for any $p \geq 1$ will thus be based on bounding the $L_p$ norms of $B_{\cM, \cN}(\bxi)$, a term based on the off-diagonal elements of $M^\top N$, and that of $D_{\cM, \cN}(\bxi)$, a term based on the diagonal elements of $M^\top N$. 

\Decomposition*
\begin{proof}
    Note that the contributions from the off-diagonal terms of $M^\top N$ to $\E[(\bxi^\top M^\top N \bxi)]$ is 0. To see this, by linearity of expectation we have
\begin{align*}
    \E_{\bxi} \left[ \sum_{\substack{j,k=1\\j\neq k}}^n \bxi_j \bxi_k \langle M_j, N_k \rangle \right]
    & =  \sum_{\substack{j,k=1\\j\neq k}}^n  \E_{\bxi_j,\bxi_k} [\bxi_j \bxi_k] \langle M_j, N_k \rangle 
     =  0~,
\end{align*}
where the last equality follows by Assumption~\ref{asmp:xi}.

Now, using Jensen's inequality, we have 
\begin{align*}
    C_{\cM, \cN}(\bxi) & = ~ \sup_{M \in \cM, N \in \cN} \left| (\bxi^\top M^\top N \bxi) - \E[(\bxi^\top M^\top N \bxi)] \right|~. \\
    & =~ \sup_{M \in \cM, N \in \cN} \left| \sum_{\substack{j,k=1\\j \neq k}}^n \bxi_j \bxi_k \langle M_j, N_k \rangle + \sum_{j=1}^n (|\bxi_j|^2 - \E|\bxi_j|^2) \langle M_j, N_j \rangle \right| \\
    & \leq~ \sup_{M \in \cM, N \in \cN} \left| \sum_{\substack{j,k=1\\j \neq k}}^n \bxi_j \bxi_k \langle M_j, N_k \rangle \right| + \sup_{M \in \cM, N \in \cN} \left| \sum_{j=1}^n (|\bxi_j|^2 - \E|\bxi_j|^2) \langle M_j, N_j \rangle \right| \\
    & =~ B_{\cM, \cN}(\bxi) +  D_{\cM, \cN}(\bxi) 
\end{align*}    
Therefore, for any $p \in [1,\infty)$, we have 
\begin{align}
    \| C_{\cM, \cN}(\bxi) \|_{L_p} & \leq \| B_{\cM, \cN}(\bxi)\|_{L_p}  +  \| D_{\cM, \cN}(\bxi) \|_{L_p}~.
\end{align}
\end{proof}
We bound $\| B_{\cM, \cN}(\bxi)\|_{L_p}$ and $\| D_{\cM, \cN}(\bxi) \|_{L_p}$ using Theorem~\ref{theo:offDiagonalLpBound} and Theorem~\ref{theo:diagonalLpBound} respectively. Note that collecting terms corresponding to {\color{Red}W}, {\color{Blue}V} and {\color{Green}U} and combining them with the observation in \eqref{eq:mark2} completes the proof of Theorem~\ref{thm:cross_product}. The rest of the proof is devoted to bounding the off-diagonal term $B_{\cM, \cN}({\boldsymbol{\xi}})$ and the diagonal term $D_{\cM, \cN}({\boldsymbol{\xi}})$ in section~\ref{subsection:Off-diagonal} and \ref{subsection:diagonal} respectively.
\subsection{The Off-diagonal Term \texorpdfstring{$B_{\cM, \cN}({\boldsymbol{\xi}})$}{textxi}}
\label{subsection:Off-diagonal}
The main result for the off-diagonal term is the following:


\noindent\textbf{Theorem 3.3.}
\emph{Let $\bxi$ be a stochastic process satisfying Assumption~\ref{asmp:xi}. Then, for all $p \geq 1$, we have}
\begin{align*}
        &\| B_{\cM, \cN}(\bxi)\|_{L_p} \lesssim \Bigg[{\color{Red}\gamma_2(\cM,\| \cdot \|_{2 \rightarrow 2})} \cdot \Big({\color{Red}\gamma_2(\cN, \| \cdot \|_{2 \rightarrow 2}) + d_F(\cN)} + {\color{Blue}\sqrt{p}~ d_{2 \rightarrow 2}(\cN)}~\Big)\nonumber \\
        & \quad\quad\quad\quad\quad + {\color{Red}\gamma_2(\cN,\| \cdot \|_{2 \rightarrow 2})} \cdot \Big({\color{Red}\gamma_2(\cM, \| \cdot \|_{2 \rightarrow 2}) + d_F(\cM)} + {\color{Blue}\sqrt{p}~ d_{2 \rightarrow 2}(\cM)}~\Big) \nonumber
        \\ & \quad\quad + {\color{Blue}~ \sqrt{p}~ \min \Big\{  d_F(\cM) \cdot d_{2 \rightarrow 2}(\cN),d_F(\cN) \cdot d_{2 \rightarrow 2}(\cM)} \Big\} \nonumber + {\color{Green}p \cdot d_{2 \rightarrow 2}(\cM) \cdot d_{2 \rightarrow 2}(\cN)}~.\Bigg]    
    \end{align*}

\subsubsection{Proof of Theorem~\ref{theo:offDiagonalLpBound}}

    We start by using a decoupling inequality from \citep{krahmer2014suprema}.
    \begin{lemma}[Theorem 2.4, \cite{krahmer2014suprema}]. Let $\bxi = (\bxi_1, \ldots, \bxi_n)$ be a sequence of independent, centered random variables, and let $F$ be a convex function. If $\mathcal{B}$ is a collection of matrices and $\bxi'$ is an independent copy of $\bxi$, then
        \[
        \mathbb{E} \sup_{B \in \mathcal{B}} F\left( \sum_{\substack{j,k=1\\j \neq k}}^n \xi_j \xi_k B_{j,k} \right)
        \leq \mathbb{E} \sup_{B \in \mathcal{B}} F\left( 4 \sum_{j,k=1}^n \xi_j {\xi_k'} B_{j,k} \right).
        \]
        \label{lemma:decouplingOffDiag}
    \end{lemma}
    We use Lemma~\ref{lemma:decouplingOffDiag} with $F(x) = |x|^p, p \geq 1$ as the convex function and set $B_{j,k} = \langle M_j, N_k \rangle$. Then
    \begin{align}
        \| B_{\cM, \cN}(\bxi) \|_{L_p} & = \E \sup_{M\in\cM, N \in \cN} F\left( \sum_{\substack{j,k=1\\j \neq k}}^n \xi_j \xi_k \langle M_j, N_k \rangle  \right)\nonumber\\
        & \leq E \sup_{M\in\cM, N \in \cN} F\left( 4 \sum_{\substack{j,k=1\\j \neq k}}^n \xi_j \xi'_k \langle M_j, N_k \rangle  \right)\nonumber\\
        &\lesssim \left\| \sup_{M \in \cM, N \in \cN} \left| \langle M \bxi, N \bxi' \rangle \right| \right\|_{L_p}~.
        \label{eq:diag_first_upper_bound}
    \end{align}
    Note that for fixed $M,N,$ the term $ \left| \langle M \bxi, N \bxi' \rangle \right|$ conditioned on $\bxi'$ is sub-gaussian and therefore its $L_p$ norm can be bounded. However, the $\sup_{M \in \cM, N \in \cN}$ inside the $\| \cdot \|_{L_p}$ does not let us use this approach. The next Theorem therefore upper bounds $\left\| \sup_{M \in \cM, N \in \cN} \left| \langle M \bxi, N \bxi' \rangle \right| \right\|_{L_p}$ by $\sup_{M \in \cM, N \in \cN} \left\|  \langle M \bxi, N \bxi' \rangle \right\|_{L_p}$ plus some additional complexity terms. Unlike \cite{krahmer2014suprema} the inner product contains two different matrices $M$ and $N$, and therefore we consider two separate admissible sequences (cf. definition ~\ref{def:gamma_2}) $\{T_r(\cM)\}_{r=0}^{\infty}$ and $\{T_r(\cN)\}_{r=0}^{\infty}$ of $\cM$ and $\cN$ respectively. We then use a generic chaining argument by creating two separate increment sequences for $\cM$ and $\cN$ and is detailed in the following theorem.
    \begin{lemma}
    Let $\bxi$ be a stochastic process satisfying Assumption~\ref{asmp:xi}, and $\bxi'$ be a decoupled tangent sequence to $\bxi$. Then, for every $p \geq 1$,
        \begin{align}
            \left\| \sup_{M \in \cM, N \in \cN} \langle M \bxi, N \bxi' \rangle \right\|_{L_p}
            &\lesssim \;\gamma_2(\cM,\| \cdot \|_{2 \rightarrow 2}) \cdot \Big[\gamma_2(\cN, \| \cdot \|_{2 \rightarrow 2}) + d_F(\cN) + \sqrt{p} d_{2 \rightarrow 2}(\cN)~\Big]\nonumber \\
            & + \;\gamma_2(\cN,\| \cdot \|_{2 \rightarrow 2}) \cdot \Big[\gamma_2(\cM, \| \cdot \|_{2 \rightarrow 2}) + d_F(\cM) + \sqrt{p} d_{2 \rightarrow 2}(\cM)~\Big] \nonumber
            \\ & \qquad\qquad\qquad\qquad + \sup_{M \in \cM, N \in \cN} \| \langle M \bxi, N \bxi' \rangle \|_{L_p}~,
        \end{align}
    \label{lemm:offd-lp}
    \end{lemma}

    \begin{proof}[\textbf{Proof of Lemma~\ref{lemm:offd-lp}}]
             Without loss of generality, assume $\cM$ and $\cN$ are finite~\cite{tala14}. 
                Let $\{T_r(\cM)\}_{r=0}^{\infty}$ and $\{T_r(\cN)\}_{r=0}^{\infty}$ be admissible sequences for $\cM$ and $\cN$ for which the minimum in the definition
                of $\gamma_2(\cM, \| \cdot \|_{2 \rightarrow 2})$ and $\gamma_2(\cN, \| \cdot \|_{2 \rightarrow 2})$ are attained respectively. Let 
                \begin{align*}
                    \pi_r M &= d_{2 \rightarrow 2}(M,T_r(\cM)) = \underset{B \in T_r(\cM)}{\argmin} ~\| B - A\|_{2 \rightarrow 2} \qquad \text{and} \qquad \Delta_r M = \pi_r M - \pi_{r-1} M~.\\
                    \pi_r N &= d_{2 \rightarrow 2}(N,T_r(\cN)) \;=\; \underset{B \in T_r(\cN)}{\argmin} ~\| B - A\|_{2 \rightarrow 2} \qquad \text{and} \qquad \Delta_r N = \pi_r N - \pi_{r-1} N~.
                \end{align*}
                For any given $p \geq 1$, let $\ell$ be the largest integer for which $2^{\ell} \leq 2 p$.
                Then,
                \begin{align*}
                    &\langle M \bxi, N \bxi'\rangle - \langle (\pi_{\ell} M) \bxi, (\pi_{\ell} N) \bxi' \rangle = \sum_{r = \ell}^{\infty}\langle (\pi_{r+1} M) \bxi, (\pi_{r+1} N) \bxi'\rangle - \langle (\pi_{r} M) \bxi, (\pi_{r} N) \bxi' \rangle \\
                    & = \sum_{r = \ell}^{\infty}\langle (\pi_{r}M +  \Delta_{r+1}M) \bxi, (\pi_{r+1}N) \bxi'\rangle - \langle (\pi_{r} M) \bxi, (\pi_{r+1}N - \Delta_{r+1} N) \bxi' \rangle \\
                    &= \sum_{r = \ell}^{\infty}\langle (\pi_{r} M) \bxi, (\pi_{r+1} N) \bxi'\rangle + \langle (\Delta_{r+1} M) \bxi, (\pi_{r+1} N) \bxi'\rangle - \langle (\pi_{r} M) \bxi, (\pi_{r+1} N) \bxi' \rangle + \langle (\pi_{r} M) \bxi, (\Delta_{r+1} N) \bxi' \rangle   \\
                    &= \sum_{r = \ell}^{\infty} \langle (\Delta_{r+1} M) \bxi, (\pi_{r+1} N) \bxi'\rangle + \langle (\pi_{r} M) \bxi, (\Delta_{r+1} N) \bxi' \rangle   \\
                \end{align*}
                Now by applying triangle inequality, we have
                \beq
                \left| \langle M \bxi, N \bxi'\rangle - \langle (\pi_{\ell} M) \bxi, (\pi_{\ell} N) \bxi' \rangle \right|
                \leq \underbrace{\left| \sum_{r=\ell}^{\infty} \langle (\Delta_{r+1} M) \bxi, (\pi_{r+1} N) \bxi' \rangle \right|}_{S_1}  
                + \underbrace{\left| \sum_{r=\ell}^{\infty} \langle (\pi_{r} M) \xi, (\Delta_{r+1} N) \xi' \rangle \right|}_{S_2}~. 
                \label{eq:l3tri}
                \eeq
                We first consider $S_1$. Let us define
                \begin{equation*}
                    X_r(M,N) = \langle (\Delta_{r+1} M) \bxi, (\pi_{r+1} N) \bxi' \rangle~. 
                \end{equation*}
                Conditioning $X_r(M,N)$ on $\bxi'$, we note 
                \begin{equation*}
                 X_r(M,N)~ \big|~\bxi' =   \langle (\Delta_{r+1} M) \bxi, (\pi_{r+1} N) \bxi' \rangle ~\big|~\bxi' = \langle  \bxi, (\Delta_{r+1} M)^T (\pi_{r+1} N) \bxi' \rangle ~\big|~\bxi'
                \end{equation*}
                is a sub-Gaussian random variable and therefore using Azuma-Hoeffding bound~\citep{bolm13,vers18} gives
                \begin{align*}
                    P\left( |X_r(M,N) | > u \| (\Delta_{r+1} M)^T (\pi_{r+1} N) \bxi' \|_2 ~ \Big|~ \bxi'~ \right) \leq 2 \exp(-u^2/2)~.
                \end{align*}
                Using $u = t 2^{r/2}$, we get
                \begin{align*}
                P\left( |X_r(M,N)
                | > t 2^{r/2} \| (\Delta_{r+1} M)^T (\pi_{r+1} N) \bxi' \|_2 ~ \Big|~ \bxi'~ \right) \leq 2 \exp(-t^2 2^r/2)~.
                \label{eq:azho}
                \end{align*}
                Since
                \begin{align*}
                \left| (\Delta_{r+1} M)^T (\pi_{r+1} N) \bxi' \right| 
                \leq \| \Delta_{r+1} M \|_{2 \rightarrow 2} \sup_{N \in \cN} \| N \bxi' \|_2~.
                \end{align*}
                we have 
                \begin{align*}
                P\left( |X_r(M,N) | > t 2^{r/2} \| \Delta_{r+1} M \|_{2 \rightarrow 2} \sup_{N \in \cN} \| N \bxi' \|_2 ~ \Big|~ \bxi'~ \right) \leq 2 \exp(-t^2 2^r/2)~.
                \end{align*}
                Now, since $|\{\pi_r M : M \in \cM \}| = |T_r(\cM)| \leq 2^{2^r}$ and $|\{\pi_r N : N \in \cM \}| = |T_r(\cN)| \leq 2^{2^r}$, by union bound, we get 
                 \begin{align*}
                P\bigg( \sup_{M \in \cM, N \in \cN}  &  ~\sum_{r=\ell}^{\infty} |X_r(M,N)| > t \left( \sup_{M \in \cM} \sum_{r=\ell}^{\infty} 2^{r/2} \| \Delta_{r+1} M \|_{2 \rightarrow 2} \right) \cdot \sup_{N \in \cN} \| N \bxi' \|_2 ~ \Big|~ \bxi'~ \bigg) \\  
                 & \leq 2 \sum_{r=\ell}^{\infty} |T_{r}(\cM)| \cdot |T_{r+1}(\cM)| \cdot |T_{r+1}(\cN)| \cdot \exp(-t^2 2^r/2) \\
                 & \leq 2 \sum_{r=\ell}^{\infty} 2^{2^{r+2}} \cdot \exp (-t^2 2^r/2) \\
                 & \leq 2 \exp(-2^{\ell} t^2)~,
                \end{align*}
                for all $t \geq t_0$, a constant. Next, note that
                \begin{align*}
                    \sup_{M \in \cM} \sum_{r=\ell}^{\infty} 2^{r/2} \| \Delta_{r+1} M \|_{2 \rightarrow 2} & = \gamma_2(\cM, \| \cdot \|_{2 \rightarrow 2} ).
                \end{align*}
                Therefore we have 
                \begin{align*}
                P\bigg( \sup_{M \in \cM, N \in \cN}  &  ~\sum_{r=\ell}^{\infty} |X_r(M,N)| > t \gamma_2(\cM,  \|\cdot \|_{2 \rightarrow 2}) \sup_{N \in \cN} \| N \bxi' \|_2 ~ \Big|~ \bxi'~ \bigg) 
                \leq 2 \exp(-p t^2)~,
                \end{align*}
                since $p \leq 2^{\ell}$ by construction which implies with $V(\bxi') = \gamma_2(\cM,  \|\cdot \|_{2 \rightarrow 2}) \sup_{N \in \cN} \| N \bxi' \|_2$, for $t \geq t_0$ we have
                \begin{align*}
                    P\left( S_1 \geq t V(\bxi') ~\big|~\bxi' \right) \leq 2 \exp(- p t^2)~.
                \end{align*}
                Note that
                \begin{align*}
                    \| S_1 \|_{L_p}^p & = \E_{\bxi,\bxi'} S_1^p = E_{\bxi'} \int_0^{\infty} p t^{p-1} P(S_1 > t ~\big|~ \bxi') dt~,
                \end{align*}
                and
                \begin{align*}
                    \int_0^{\infty} p t^{p-1} P(S_1 > t ~\big|~ \bxi') dt & \leq c^p V(\bxi')^p + \int_{cV(\bxi')}^{\infty} p t^{p-1} P(S_1 > t ~\big|~ \bxi') dt \\
                    & \leq c^p V(\bxi')^p + V(\bxi')^p \int_{c}^{\infty} p \tau^{p-1} P( S_1 > \tau V(\bxi') | \bxi') d\tau \\
                    & \leq c_1^p V(\bxi')^p~,
                \end{align*}
                where $c \geq t_0,c_1$ are suitable constants that depend on $L$. As a result, $\| S_1 \|_{L_p} \leq c_1 V(\bxi') = c_1 \|V(\bxi)\|_{L_p}$, i.e., we have the following bound on $S_1$.
                \begin{align*}
                    \| S_1 \|_{L_p} \lesssim \gamma_2(\cM,  \|\cdot \|_{2 \rightarrow 2}) \Big\|\sup_{N \in \cN} \| N \bxi \|_2 \Big\|_{L_p}
                \end{align*}
                Note that a similar analysis follows for $S_2$, and we can bound $\| S_2 \|_{L_p}$. As a result
                \begin{equation}
                    \| S_1 + S_2 \|_{L_p}  \lesssim \Big(\gamma_2(\cM,  \|\cdot \|_{2 \rightarrow 2}) \cdot \Big\|\sup_{N \in \cN} \| N \bxi \|_2 \Big\|_{L_p} + \gamma_2(\cN,  \|\cdot \|_{2 \rightarrow 2}) \cdot \Big\|\sup_{M \in \cM} \| M \bxi \|_2 \Big\|_{L_p}\Big)
                    \label{eq:s1s2}
                \end{equation}
                Further, since $| \{ \pi_{\ell} M : M \in \cM \} | \leq 2^{2^{\ell}} \leq \exp(2p)$, and $| \{ \pi_{\ell} N : N \in \cN \} | \leq 2^{2^{\ell}} \leq \exp(2p)$ we have 
                \begin{align*}
                    \E \sup_{M \in \cM, N \in \cN} & | \langle (\pi_{\ell} M) \bxi, (\pi_{\ell} N) \bxi \rangle |^p  \leq \sum_{M \in T_{\ell}(M), N \in T_{\ell}(N)} \E| \langle M\bxi, N \bxi\rangle |^p \\
                    &\leq 2^{2p} 2^{2p} \sup_{M \in \cM, N \in \cN} E | \langle A \bxi, A \bxi \rangle |^p = 2^{4p} \sup_{M \in \cM, N \in \cN}  | \langle M \bxi, N \bxi \rangle |^p~,
                \end{align*}
                so that
                \begin{equation}
                    \left\| \sup_{M \in \cM, N \in \cN} | \langle (\pi_{\ell} M) \bxi, (\pi_{\ell} N) \bxi \rangle \right\|_{L_p} \leq 16~  \sup_{M \in \cM, N \in \cN} \Big\| \langle M \bxi, N \bxi \rangle \Big\|_{L_p}~.
                    \label{eq:s0}
                \end{equation}
                Combining \eqref{eq:l3tri}, \eqref{eq:s1s2}, and \eqref{eq:s0} and using triangle inequality we have
                \begin{align*}
                    \left\| \sup_{M \in \cM, N \in \cN} \langle M \bxi, N \bxi' \rangle \right\|_{L_p} \!\!\!\! &\lesssim \gamma_2(\cM, \|\cdot \|_{2 \rightarrow 2}) \cdot \Big\|\sup_{N \in \cN} \| N \bxi' \|_2 \Big\|_{L_p} \!\!\!\! + \gamma_2(\cN,  \|\cdot \|_{2 \rightarrow 2}) \cdot \Big\|\sup_{M \in \cM} \| M \bxi \|_2 \Big\|_{L_p}\\
                    & \qquad\qquad + \sup_{M \in \cM, N \in \cN} \Big\| \langle M \bxi, N \bxi' \rangle \Big\|_{L_p}~.
                \end{align*}
                
                Using Theorem 3.5 from \citep{krmr14} we have that 
                \begin{align*}
                    \Big\|\sup_{M \in \cM} \| M \bxi \|_2 \Big\|_{L_p} & \lesssim \cdot \Big(\gamma_2(\cM, \| \cdot \|_{2 \rightarrow 2}) + d_F(\cM) + \sqrt{p} d_{2 \rightarrow 2}(\cM) \Big)\\
                    \Big\|\sup_{N \in \cN} \| N \bxi \|_2 \Big\|_{L_p} & \lesssim \cdot \Big( \gamma_2(\cN, \| \cdot \|_{2 \rightarrow 2}) + d_F(\cN) + \sqrt{p} d_{2 \rightarrow 2}(\cN)\Big)
                \end{align*}
                Combining all these completes the proof of Lemma~\ref{lemm:offd-lp}.
        \end{proof}
    Therefore using \eqref{eq:diag_first_upper_bound} and Lemma~\ref{lemm:offd-lp} we get
    \begin{align}
        \| B_{\cM, \cN}(\bxi) \|_{L_p} &\lesssim \gamma_2(\cM,\| \cdot \|_{2 \rightarrow 2}) \cdot \Big[\gamma_2(\cN, \| \cdot \|_{2 \rightarrow 2}) + d_F(\cN) + \sqrt{p} d_{2 \rightarrow 2}(\cN)~\Big]\nonumber \\
            & + \;\gamma_2(\cN,\| \cdot \|_{2 \rightarrow 2}) \cdot \Big[\gamma_2(\cM, \| \cdot \|_{2 \rightarrow 2}) + d_F(\cM) + \sqrt{p} d_{2 \rightarrow 2}(\cM)~\Big] \nonumber
            \\ & \qquad\qquad\qquad\qquad + \sup_{M \in \cM, N \in \cN} \| \langle M \bxi, N \bxi' \rangle \|_{L_p}
            \label{eq:diagBound2}
    \end{align}
    Next we consider $\displaystyle\sup_{M \in \cM, N \in \cN}\Big\| \langle M \bxi, N \bxi' \rangle \Big\|_{L_p}$ and bound it using the following Lemma.
    \begin{lemma}
        Let $\bxi$ be a stochastic process satisfying Assumption~\ref{asmp:xi}, and let $\bxi'$ be a decoupled tangent sequence. Then, for every $p \geq 1$,
        \begin{align}
            \sup_{M \in \cM, N \in \cN}\Big\| \langle M \bxi, N \bxi' \rangle \Big\|_{L_p} &\lesssim \min \Big\{ \sqrt{p} \cdot d_F(\cM) \cdot d_{2 \rightarrow 2}(\cN), \sqrt{p} \cdot d_F(\cN) \cdot d_{2 \rightarrow 2}(\cM) \Big\}  + p \cdot d_{2 \rightarrow 2}(\cM) \cdot d_{2 \rightarrow 2}(\cN)~.
            \label{eq:offd-lp-2nd}
        \end{align}
    \label{lemm:offd-lp-2nd}
    \end{lemma} 
    \begin{proof}[\textbf{Proof of Lemma~\ref{lemm:offd-lp-2nd}}]
    Fix $M \in \cM, N \in \cN$ and let $S = \{M^\top N x : x \in B^n_2, M \in \cM, N\in \cN\}$, where $B^n_2 = \{x \in \R^n: \|x\|_2 \leq 1 \}$  \rdcomment{Define $B^n_2$}. Conditioned on $\bxi$, the random variable $\langle \bxi, M^\top N \bxi'\rangle$ is sub-gaussian. Therefore for some global constant $\tilde{C}>0$, we have \rdcomment{cite Vershynin?}
    \begin{align*}
        \Big\| \langle M \bxi, N \bxi' \rangle \Big\|_{L_p} &\leq \tilde{C} \Bigg( \E_{\bxi'}\big((\E_{\bxi} \big| \bxi,M^\top N \bxi' \big|^p)^{1/p}\big)^p \Bigg)^{1/p} \leq \tilde{C} (\E_{\bxi'}(L\sqrt{p})^p \|M^\top N\bxi'\|_2^p)^{1/p}\\
        &\lesssim L\sqrt{p} \Bigg( \E_{\bxi'} \sup_{y \in S} |\langle y, \bxi' \rangle|^p \Bigg)^{1/p}
    \end{align*}
    Now using Theorem 2.3 from \cite{krmr14} we have for every $p\geq 1$
    \begin{align*}
        \Bigg( \E_{\bxi'} \sup_{y \in S} |\langle y, \bxi' \rangle|^p \Bigg)^{1/p} \lesssim \Big(\E_{\mathbf{g}} \sup_{y \in S} |\langle \mathbf{g},y \rangle| + \sup_{y \in S} (\E_{\bxi'} |\langle \bxi',y \rangle|^p)^{1/p}\Big)
    \end{align*}
    where $\mathbf{g}$ is a standard Gaussian vector. The first term in the rhs can be bounded as follows:
    \begin{align*}
        \E_{\mathbf{g}} \sup_{y \in S} |\langle \mathbf{g},y \rangle| &= \E_{\mathbf{g}} \|M^\top N g\|_2 \leq (\E_{\mathbf{g}} \|M^\top N g\|_2^2)^{1/2} = \|M^\top N\|_{F} \\
        &\leq \min \Big\{ \|M\|_{2\rightarrow 2} \|N\|_F, \|N\|_{2\rightarrow 2} \|M\|_F \Big\}
    \end{align*}
    Next the second term in the rhs can be bounded as follows:
    \begin{align*}
        \sup_{y \in S} (\E_{\bxi'} |\langle \bxi',y \rangle|^p)^{1/p} \leq L \sup_{z \in B^p_2} \sqrt{p} \|M^\top N z\|_2 \leq L \sqrt{p} \|M\|_{2 \rightarrow 2} \|N\|_{2 \rightarrow 2}.
    \end{align*}
Combining all the above bounds we get:
\begin{align}
     & \Big\| \langle M \bxi, N \bxi' \rangle \Big\|_{L_p} \lesssim \sqrt{p} \min \Big\{ \|M\|_{2\rightarrow 2} \|N\|_F, \|N\|_{2\rightarrow 2} \|M\|_F \Big\} +  {p} \|M\|_{2 \rightarrow 2} \|N\|_{2 \rightarrow 2}
\end{align}
Now recall that for the set $\cM$, we have $d_F(\cM) = \sup_{M \in \cM} \| M \|_F$, and $d_{2 \rightarrow 2}(\cM) = \sup_{A \in \cM} \| A \|_{2 \rightarrow 2}$, which implies
\begin{align*}
    \sup_{M \in \cM, N \in \cN} \Big\| \langle M \bxi, N \bxi' \rangle \Big\|_{L_p} & \lesssim \Bigg( \sqrt{p }\min \Big\{ d_{2\rightarrow 2}(\cM) d_F(\cN), d_{2\rightarrow 2}(\cN) d_F(\cM) \Big\} + p d_{2\rightarrow 2}(\cM) d_{2\rightarrow 2}(\cN)\Bigg).
\end{align*}
\end{proof}
Combining Lemma~\ref{eq:offd-lp-2nd} with \eqref{eq:diagBound2} completes the proof of Theorem~\ref{theo:diagonalLpBound}.

\subsection{The Diagonal Term \texorpdfstring{$D_{\cM, \cN}({\boldsymbol{\xi}})$}{textxi}}
\label{subsection:diagonal}
For the diagonal terms, we have the following main result:
\textbf{Theorem 3.5} 
\emph{Let $\bxi$ be a stochastic process satisfying Assumption~\ref{asmp:xi}. Then, for all $p \geq 1$, we have}
\begin{align*}
        &\left\| D_{\cM,\cN}(\bxi) \right\|_{L_p} \lesssim \Bigg[{\color{Red}\gamma_2(\cM,\| \cdot \|_{2 \rightarrow 2})} \cdot \Big({\color{Red}\gamma_2(\cN, \| \cdot \|_{2 \rightarrow 2}) + d_F(\cN)} + {\color{Blue}\sqrt{p}~ d_{2 \rightarrow 2}(\cN)}~\Big)\nonumber \\
            & \quad\quad\quad\quad\quad + {\color{Red}\gamma_2(\cN,\| \cdot \|_{2 \rightarrow 2})} \cdot \Big({\color{Red}\gamma_2(\cM, \| \cdot \|_{2 \rightarrow 2}) + d_F(\cM)} + {\color{Blue}\sqrt{p}~ d_{2 \rightarrow 2}(\cM)}~\Big) \nonumber
            \\ & \quad\quad + {\color{Blue}~ \sqrt{p}~ \min \Big\{  d_F(\cM) \cdot d_{2 \rightarrow 2}(\cN),d_F(\cN) \cdot d_{2 \rightarrow 2}(\cM)} \Big\} \nonumber + {\color{Green}p \cdot d_{2 \rightarrow 2}(\cM) \cdot d_{2 \rightarrow 2}(\cN)}~.\Bigg]    
    \end{align*}

\subsubsection{Proof of Theorem~\ref{theo:diagonalLpBound}}


By definition of $D_{\cM, \cN}(\bxi)$ and {from Lemma 9 in \cite{banerjee2019random}}
,  we have
\begin{align*}
    \| D_{\cM, \cN}(\bxi)) \|_{L_p} & = \left\| \sup_{M\in\cM, N\in\cN} \Big| \sum_{j=1}^n (|\bxi_j|^2 - \E|\bxi_j|^2) \langle M_j, N_j \rangle \Big| \right\|_{L_p}\!\! 
     \\
     & \leq 2 \left\| \sup_{M\in\cM, N\in\cN} \Big| \sum_{j=1}^n \eps_j |\bxi_j|^2  \langle M_j, N_j \rangle \Big| \right\|_{L_p},
\end{align*}
where $\{ \eps_j\}$ is a set of independent Rademacher variables independent of $\bxi$. Let $\{ g_j \}$ be a sequence of independent Gaussian random variables. Since $\bxi_j$ is a $L$-sub-Gaussian random variable~\citep{vers18}, there is an absolute constant $c$ such that for all $t > 0$
\begin{align*}
    \P \left( |\xi_j|^2 \geq t L^2 \right) & \leq c \P(g_j^2 \geq t)~.
\end{align*}
Then, from contraction of stochastic processes (\cite[Lemma~{4.6}]{leta91}), we have
\begin{align}
    \| & D_{\cM, \cN}(\bxi)) \|_{L_p}  \leq 2 \left\| \sup_{M\in\cM, N\in\cN} \Big| \sum_{j=1}^n \eps_j |\bxi_j|^2  \langle M_j, N_j \rangle \Big| \right\|_{L_p} \notag\\
    & \leq 2c{L^{2}} \left\| \sup_{M\in\cM, N\in\cN} \Big| \sum_{j=1}^n \eps_j |g_j|^2  \langle M_j, N_j \rangle \Big| \right\|_{L_p}\notag \\
    & \overset{(a)}{\leq} 2c{L^{2}}\left\| \sup_{M\in\cM, N\in\cN} \Big| \sum_{j=1}^n \eps_j (|g_j|^2 - 1)  \langle M_j, N_j \rangle \Big| \right\|_{L_p}  + 2c{L^{2}}\left\| \sup_{M\in\cM, N\in\cN} \Big| \sum_{j=1}^n \eps_j   \langle M_j, N_j \rangle \Big| \right\|_{L_p}\notag \\
    & \overset{(b)}{\leq} 4c{L^{2}}\left\| \sup_{M\in\cM, N\in\cN} \Big| \sum_{j=1}^n (|g_j|^2 - 1)  \langle M_j, N_j \rangle \Big| \right\|_{L_p}  + 2c{L^{2}}\left\| \sup_{M\in\cM, N\in\cN} \Big| \sum_{j=1}^n \eps_j   \langle M_j, N_j \rangle \Big| \right\|_{L_p} \notag \\ 
    & \leq {4cL^{2}} \left\| D_{\cM, \cN}(\g) \right\|_{L_p} + 2c{L^{2}}\left\| \sup_{M\in\cM, N\in\cN} \Big| \sum_{j=1}^n \eps_j   \langle M_j, N_j \rangle \Big| \right\|_{L_p}~,
\label{eq:diagonal_1_app}
\end{align}
where (a) follows from Jensen's inequality and since $E|g_j|^2 = 1$, and (b) follows by de-symmetrization following {\cite[Lemma 11]{banerjee2019random}} 
and since the convex function here is 1-Lipschitz.

By triangle inequality, we have 
\begin{equation}
\begin{split}
    \left\| D_{\cM,\cN}(\g) \right\|_{L_p} & \leq \left\| C_{\cM,\cN}(\g) \right\|_{L_p} + \left\| B_{\cM,\cN}(\g) \right\|_{L_p}
 \end{split}
 \label{eq:diag11}
\end{equation}
In order to handle $\left\|C_{\mathcal{M},\mathcal{N}}(\g)\right\|_{L_{p}}$, we require a stronger decoupling inequality for an order 2 Gaussian chaos than the one used in Theorem~2.5 in \cite{krahmer2014suprema}.
\begin{theorem}
\label{thm:decouple}
There exists an absolute constant $C$ such that the following holds for all $p \geq 1$.  Let $\g=(\g_1,\dots,\g_n)$ be a sequence of independent standard normal random variables. If $\mathcal{B}$ is a collection of matrices and $\g'$ is an independent copy of $\g$, then
\begin{align*}
\mathbb{E}\sup_{B\in\mathcal B}
\left|\sum_{\substack{j,k=1\\j\neq k}}^n \g_j \g_k B_{j,k}
+\sum_{j=1}^n\left(\g_j^2-1\right)\,B_{j,j}\right|^p
\;\le\;
C^p\,\mathbb{E}\sup_{B\in\mathcal B}
\left|\sum_{j,k=1}^n \g_j\,\g'_k\,B_{j,k}\right|^p.
\end{align*}
\end{theorem}
\begin{proof}
For each $B\in\mathcal{B}$, denote $C^{B}=(C^{B}_{j,k})_{n\times n}$ such that $C^{B}_{j,k}=\frac{B_{j,k}+B_{k,j}}{2},\forall j,k\in[n]\times[n]$, and denote $\mathcal{C}^{\mathcal{B}}$ is the collection of $C^{B}$ for all $B\in\mathcal{B}$. Then $\mathcal{C}^{\mathcal{B}}$ is a collection of Hermitian matrices. According to Theorem 2.5 in \cite{krahmer2014suprema}, we have
\begin{align}
\mathbb{E}\sup_{C^{B}\in\mathcal{C}^{\mathcal{B}}}
\left|\sum_{\substack{j,k=1\\j\neq k}}^n \g_j \g_k C^{B}_{j,k}
+\sum_{j=1}^n\left(\g_j^2-1\right)\,C^{B}_{j,j}\right|^p
\;\le\;
c^p\,\mathbb{E}\sup_{C^{B}\in\mathcal C^{\mathcal{B}}}
\left|\sum_{j,k=1}^n \g_j\,\g'_k\,C^{B}_{j,k}\right|^p.\label{eq:decouple_1}
\end{align}
In addition, the left-hand side is
\begin{align}
    &\mathbb{E}\sup_{C^{B}\in\mathcal{C}^{\mathcal{B}}}
\left|\sum_{\substack{j,k=1\\j\neq k}}^n \g_j \g_k C^{B}_{j,k}
+\sum_{j=1}^n\left(\g_j^2-1\right)\,C^{B}_{j,j}\right|^p\notag\\
=&\mathbb{E}\sup_{B\in\mathcal{B}}
\left|\sum_{\substack{j,k=1\\j\neq k}}^n \g_j \g_k \frac{B_{j,k}+B_{k,j}}{2}
+\sum_{j=1}^n\left(\g_j^2-1\right)\,B_{j,j}\right|^p\notag\\
=&\mathbb{E}\sup_{B\in\mathcal{B}}
\left|\sum_{\substack{j,k=1\\j\neq k}}^n \g_j \g_k B_{j,k}
+\sum_{j=1}^n\left(\g_j^2-1\right)\,B_{j,j}\right|^p\label{eq:decouple_2}.
\end{align}
Further the right-hand side can be upper bounded as follow.
\begin{align}
    c^p\,\mathbb{E}\sup_{C^{B} \in\mathcal C^{\mathcal{B}}}
\left|\sum_{j,k=1}^n \g_j\,\g'_k\,C^{B}_{j,k}\right|^p = & \; c^p\,\mathbb{E}\sup_{B\in\mathcal B}
\left|\sum_{j,k=1}^n \g_j\,\g'_k\,\frac{B_{j,k}+B_{k,j}}{2}\right|^p\notag\\
=&\frac{c^{p}}{2^{p}}\,\mathbb{E}\sup_{B\in\mathcal B}
\left|\sum_{j,k=1}^n \g_j\,\g'_k\,\left(B_{j,k}+B_{k,j}\right)\right|^p\notag\\
=&\frac{c^{p}}{2^{p}}\,\mathbb{E}\sup_{B\in\mathcal B}
\left|\sum_{j,k=1}^n \g_j\,\g'_k\,B_{j,k}+\sum_{j,k=1}^n \g_j\,\g'_k\,B_{k,j}\right|^p\notag\\
\overset{(a)}{\leq}&\frac{c^{p}}{2^{p}}\mathbb{E}\sup_{B\in\mathcal{B}}
2^{p}\left(\left|\sum_{\substack{j,k=1\\j\neq k}}^n \g_j \g'_k B_{j,k}
\right|^{p}+\left|\sum_{j,k=1}^n \g_j\,\g'_k\,B_{k,j}\right|^p\right)\notag\\
\end{align}
where $(a)$ utilizes the inequality $\left|a+b\right|^{p}\leq 2^{p}(\left|a\right|^{p}+\left|b\right|^{p})$. Therefore
\begin{align}
c^p\,\mathbb{E}\sup_{C^{B} \in\mathcal C^{\mathcal{B}}} & \leq c^{p}\mathbb{E}\sup_{B\in\mathcal{B}}\left|\sum_{\substack{j,k=1\\j\neq k}}^n \g_j \g'_k B_{j,k}
\right|^{p}+c^{p}\mathbb{E}\sup_{B\in\mathcal{B}}\left|\sum_{j,k=1}^n \g_j\,\g'_k\,B_{k,j}\right|^p\notag\\
=&c^{p}\mathbb{E}\sup_{B\in\mathcal{B}}\left|\sum_{\substack{j,k=1\\j\neq k}}^n \g_j \g'_k B_{j,k}
\right|^{p}+c^{p}\mathbb{E}\sup_{B\in\mathcal{B}}\left|\sum_{j,k=1}^n \g'_j\,\g_k\,B_{j,k}\right|^p\notag\\
\overset{(b)}{=}&2c^{p}\mathbb{E}\sup_{B\in\mathcal{B}}\left|\sum_{\substack{j,k=1\\j\neq k}}^n \g_j \g'_k B_{j,k}
\right|^{p}\label{eq:decouple_3}
\end{align}
where $(b)$ holds since $\g'$ is an independent copy of $\g$. Substituting \eqref{eq:decouple_2} and \eqref{eq:decouple_3} into \eqref{eq:decouple_1}, we can get that
\begin{align*}
    \mathbb{E}\sup_{B\in\mathcal{B}}
\left|\sum_{\substack{j,k=1\\j\neq k}}^n \g_j \g_k B_{j,k}
+\sum_{j=1}^n\left(\g_j^2-1\right)\,B_{j,j}\right|^p\leq2c^{p}\mathbb{E}\sup_{B\in\mathcal{B}}\left|\sum_{\substack{j,k=1\\j\neq k}}^n \g_j \g'_k B_{j,k}
\right|^{p}
\end{align*}
By setting $C=2^{\frac{1}{p}}c$, we finish the proof.
\end{proof}

Now 
\begin{align*}
\left\| C_{\cM,\cN}(\mathbf{g}) \right\|_{L_p} 
&= 
\sup_{M \in \cM, N \in \cN} \left| \sum_{\substack{j,k=1\\j \neq k}}^n \mathbf{g}_j \mathbf{g}_k \langle M_j, N_k \rangle + \sum_{j=1}^n (|\mathbf{g}_j|^2 - \E|\mathbf{g}_j|^2) \langle M_j, N_j \rangle \right|
\\
&\overset{(a)}{\leq} C
\left\| 
\sup_{M \in \cM, N \in \cN} 
\left| \sum_{\substack{j,k=1}}^n \mathbf{g}_j \mathbf{g'}_k \langle M_j, N_k \rangle\right| \right\|_{L_p}
= 
\left\| 
\sup_{M \in \cM, N \in \cN}  
\left| 
\langle M\mathbf{g}, N\mathbf{g'} \rangle 
\right| 
\right\|_{L_p}
\\
&\overset{(b)}{\lesssim}  \;\gamma_2(\cM,\| \cdot \|_{2 \rightarrow 2}) \cdot \Big[\gamma_2(\cN, \| \cdot \|_{2 \rightarrow 2}) + d_F(\cN) + \sqrt{p} d_{2 \rightarrow 2}(\cN)~\Big]\nonumber \\
    & \quad + \;\gamma_2(\cN,\| \cdot \|_{2 \rightarrow 2}) \cdot \Big[\gamma_2(\cM, \| \cdot \|_{2 \rightarrow 2}) + d_F(\cM) + \sqrt{p} d_{2 \rightarrow 2}(\cM)~\Big] \nonumber
    \\ & \quad +  \min \Big\{ \sqrt{p} \cdot d_F(\cM) \cdot d_{2 \rightarrow 2}(\cN), \sqrt{p} \cdot d_F(\cN) \cdot d_{2 \rightarrow 2}(\cM) \Big\} \nonumber\\
        & \qquad\qquad + p \cdot d_{2 \rightarrow 2}(\cM) \cdot d_{2 \rightarrow 2}(\cN)~.
\end{align*}
{where (a) uses Theorem \ref{thm:decouple}, and (b) holds because of Lemma~\ref{lemm:offd-lp} and Lemma~\ref{lemm:offd-lp-2nd}}. Term $\left\| B_{\cM,\cN}(\g) \right\|_{L_p}$ can be bounded using Theorem~\ref{theo:offDiagonalLpBound} thus giving
\begin{align}
    \left\| D_{\cM,\cN}(\g) \right\|_{L_p} \lesssim \Bigg[&\gamma_2(\cM,\| \cdot \|_{2 \rightarrow 2}) \cdot \Big[\gamma_2(\cN, \| \cdot \|_{2 \rightarrow 2}) + d_F(\cN) + \sqrt{p} d_{2 \rightarrow 2}(\cN)~\Big]\nonumber \\
    & \quad + \gamma_2(\cN,\| \cdot \|_{2 \rightarrow 2}) \cdot \Big[\gamma_2(\cM, \| \cdot \|_{2 \rightarrow 2}) + d_F(\cM) + \sqrt{p} d_{2 \rightarrow 2}(\cM)~\Big] \nonumber
    \\ & \quad + \min \Big\{ \sqrt{p} \cdot d_F(\cM) \cdot d_{2 \rightarrow 2}(\cN), \sqrt{p} \cdot d_F(\cN) \cdot d_{2 \rightarrow 2}(\cM) \Big\} \nonumber\\
        & \qquad\qquad + p \cdot d_{2 \rightarrow 2}(\cM) \cdot d_{2 \rightarrow 2}(\cN)~.\Bigg] \label{eq:diagonal_2}   
\end{align}
Next we bound the second term $\displaystyle \left\| \sup_{M\in\cM, N\in\cN} \Big| \sum_{j=1}^n \eps_j   \langle M_j, N_j \rangle \Big| \right\|_{L_p}$. We use the following theorem to bound this term.

\begin{theorem}
    Consider a stochastic process $\{ X_{(u,v)}\}$ where $u \in \mathcal{U}$, $v \in \mathcal{V}$, and let $d(\cdot,\cdot)$ is the metric. Suppose for any $t > 0$, with probability at least $1 - c_0 \exp\left(-\frac{t^2}{2}\right)$, $\{ X_{(u,v)}\}$ satisfies  the following: 
    \begin{align*}
        \left| X_{(u_1, v)} - X_{(u_2, v)} \right| &\leq C_u t \cdot d(u_1, u_2), \\
        \left| X_{(u, v_1)} - X_{(u, v_2)} \right| &\leq C_v t \cdot d(v_1, v_2).
    \end{align*}
    
    Then, with probability $1 - 2c_0 \exp\left( - \frac{t^2}{2}\right)$
    \begin{align*}
    \sup_{u \in \mathcal{U}, \, v \in \mathcal{V}} \left| X_{u,v} \right|
    &\leq 4\sqrt{2}t\left(C_{u} \gamma_2(\mathcal{U}, d) + C_{v}\gamma_2(\mathcal{V}, d) \right).
    \end{align*}
    \label{theo:double_tree}
\end{theorem}
{\begin{proof}
Let $(\mathcal{U}_{k})$ be an admissible sequence of subsets of $\mathcal{U}$, and denote $\mathcal{U}_{0} = {u_{0}}$. Let $(\mathcal{V}_{k})$ be an admissible sequence of subsets of $\mathcal{V}$, and denote $\mathcal{V}_{0} = {v_{0}}$. We now walk from $u_0$ to a general point $u \in \mathcal{U}$ along the chain
\begin{align*}
u_0 \;=\;\pi_0(u)\;\to\;\pi_1(u)\;\to\;\cdots\;\to\;\pi_{K_{1}}(u)\;=\;u,
\end{align*}
of points $\pi_k(u)\in \mathcal{U}_k$ that are chosen as best approximations to $u$ in $\mathcal{U}_k$, i.e.
\begin{align*}
d\bigl(u,\pi_k(u)\bigr)\;=\;d\bigl(u,\mathcal{U}_k\bigr).
\end{align*}
Similarly, We walk from $v_0$ to a general point $v \in \mathcal{V}$ along the chain
\begin{align*}
v_0 \;=\;\pi_0(v)\;\to\;\pi_1(v)\;\to\;\cdots\;\to\;\pi_{K_{2}}(v)\;=\;v,
\end{align*}
of points $\pi_k(v)\in \mathcal{V}_k$ that are chosen as best approximations to $v$ in $\mathcal{V}_k$, i.e.
\begin{align*}
d\bigl(v,\pi_k(v)\bigr)\;=\;d\bigl(v,\mathcal{V}_k\bigr).
\end{align*}
The displacement $X_{(u,v)} - X_{(u_{0},v_{0})}$ can be expressed as a telescoping sum:
\begin{align}
\left|X_{(u,v)} - X_{(u_{0},v_{0})}\right|
&=
\left|\sum_{k=1}^{K_{2}} \bigl(X_{(u,\pi_k(v))} - X_{(u,\pi_{k-1}(t))}\bigr)+\sum_{k=1}^{K_{1}}\bigl(X_{(\pi_{k}(u),v_{0})}-X_{(\pi_{k-1}(u),v_{0})}\bigr)\right|\notag\\
&\leq\left|\sum_{k=1}^{K_{2}} \bigl(X_{(u,\pi_k(v))} - X_{(u,\pi_{k-1}(t))}\bigr)\right|+\left|\sum_{k=1}^{K_{1}}\bigl(X_{(\pi_{k}(u),v_{0})}-X_{(\pi_{k-1}(u),v_{0})}\bigr)\right|.\label{eq:generic_1}
\end{align}
According to the assumption, with probability $1-c_{0}\exp\left(-4t^{2}2^{k}d\left(\pi_{k}(v),\pi_{k-1}(v)\right)\right)$, we have
\begin{align*}
    \left|X_{(u,\pi_k(v))} - X_{(u,\pi_{k-1}(v))}\right|\leq 2\sqrt{2}C_{v}t2^{k/2}d\left(\pi_{k}(v),\pi_{k-1}(v)\right).
\end{align*}
We can now unfix $v\in\mathcal{V}$ by taking a union bound over
\begin{align*}
    \left|\mathcal{V}_{k}\right|\left|\mathcal{V}_{k-1}\right|\leq\left|\mathcal{V}_{k}\right|^{2}=2^{2^{k+1}}
\end{align*}
possible pairs $\left(\pi_{k}(v),\pi_{k-1}(v)\right)$. Similarly, we can unfix $k$ by a union bound over all $k\in\mathbb{N}$. Then with probability at least
\begin{align*}
    1-\sum_{k=1}^{\infty}2^{2^{k+1}}\cdot c_{0}\exp\left(-4t^{2}2^{k}\right)\geq1-c_{0}\exp\left(-\frac{t^{2}}{2}\right),
\end{align*}
for all $v\in\mathcal{V}$ and $k\in\mathbb{N}$,
\begin{align}
    \left|\sum_{k=1}^{K_{2}} \bigl(X_{(u,\pi_k(v))} - X_{(u,\pi_{k-1}(t))}\bigr)\right|&\leq\sum_{k=1}^{K_{2}}\left|X_{(u,\pi_k(v))} - X_{(u,\pi_{k-1}(t))}\right|\notag\\&\leq2\sqrt{2}C_{v}t\sum_{k=1}^{\infty}2^{k/2}d\left(\pi_{k}(v),\pi_{k-1}(v)\right)\notag\\
    &\leq2\sqrt{2}C_{v}t\sum_{k=1}^{\infty}2^{k/2}\left(d\left(v,\pi_{k}(v)\right)+d\left(v,\pi_{k-1}(v)\right)\right)\notag\\
    &=2\sqrt{2}C_{v}t\sum_{k=1}^{\infty}2^{k/2}d\left(v,\mathcal{V}_{k}\right)+4C_{u}a\sum_{k=0}^{\infty}2^{k/2}d\left(v,\mathcal{V}_{k}\right)\notag\\
    &\leq4\sqrt{2}C_{v}t\gamma_{2}\left(\mathcal{V},d\right).\label{eq:generic_2}
\end{align}
Following a similar analysis, with probability at least $1-c_{0}\exp\left(-\frac{t^{2}}{2}\right)$,
\begin{align}
    \left|\sum_{k=1}^{K_{1}}\bigl(X_{(\pi_{k}(u),v_{0})}-X_{(\pi_{k-1}(u),v_{0})}\bigr)\right|\leq4\sqrt{2}C_{u}t\gamma_{2}\left(\mathcal{U},d\right).\label{eq:generic_3}
\end{align}
Therefore, substituting \eqref{eq:generic_2} and \eqref{eq:generic_3} into \eqref{eq:generic_1}, with probability at least $1-2c_{0}\exp\left(-\frac{t^{2}}{2}\right)$,
\begin{align*}
    \left|X_{(u,v)} - X_{(u_{0},v_{0})}\right|
&\leq
\left|\sum_{k=1}^{K_{2}} \bigl(X_{(u,\pi_k(v))} - X_{(u,\pi_{k-1}(t))}\bigr)\right|+\left|\sum_{k=1}^{K_{1}}\bigl(X_{(\pi_{k}(u),v_{0})}-X_{(\pi_{k-1}(u),v_{0})}\bigr)\right|\\
&\leq4\sqrt{2}C_{u}t\gamma_{2}\left(\mathcal{U},d\right)+4\sqrt{2}C_{v}t\gamma_{2}\left(\mathcal{V},d\right)
\end{align*}
Then we finish the proof.
\end{proof}}

{According to \cite[Lemma 7]{banerjee2019random}, 
\begin{align}
    &\left\| \sup_{M \in \mathcal{M}, N \in \mathcal{N}} 
\left| \sum_{j=1}^n \varepsilon_j \langle M_j, N_j \rangle \right| \right\|_{L_p}\notag\\
=&\left(\mathbb{E}_{\varepsilon}\left[\sup_{M \in \mathcal{M}, N \in \mathcal{N}} 
\left| \sum_{j=1}^n \varepsilon_j \langle M_j, N_j \rangle \right|\right]^{p}\right)^{1/p}\notag\\
=&\left(\mathbb{E}_{\varepsilon}\left[\sup_{M \in \mathcal{M}, N \in \mathcal{N}} 
\left| \sum_{j=1}^n \varepsilon_j \langle M_j, N_j \rangle \right|^{p}\right]\right)^{1/p}\notag\\
\lesssim&\mathbb{E}_{\g}\left[\sup_{M \in \mathcal{M}, N \in \mathcal{N}} 
\left| \sum_{j=1}^n \g_j \langle M_j, N_j \rangle \right|\right]+\sup_{M\in\mathcal{M},N\in\mathcal{N}}\left(\mathbb{E}_{\varepsilon}\left[\left| \sum_{j=1}^n \varepsilon_j \langle M_j, N_j \rangle \right|^{p}\right]\right)^{1/p}\label{eq:rest_1}
\end{align}
in which $\g_{j}\sim\mathcal{N}(0,1)$ are independent. For the first term, define $\displaystyle X_{(M,N)} = \left| \sum_{j=1}^n \g_j \langle M_j, N_j \rangle \right|$. Then
\begin{align*}
\left| X_{(M^1, N)} - X_{(M^2, N)} \right|
&= \left| \left|\sum_{j=1}^n \g_j \left\langle M_j^1, N_j \right\rangle\right| - \left|\sum_{j=1}^n \g_j \left\langle M_j^2, N_j \right\rangle \right| \right| \\
&= \left| \sum_{j=1}^n \g_j \left\langle M_j^1 - M_j^2, N_j \right\rangle \right| \\
&\leq \sum_{j=1}^n \varepsilon_j \left\langle M_j^1 - M_j^2, N_j \right\rangle
\end{align*}
Since $\g_j$ are standard normal random variables, therefore with probability $1 - 2e^{-u^2/2}$,
\begin{align*}
\left| X_{(M^1, N)} - X_{(M^2, N)} \right|
&\leq u \left(\sum_{j=1}^{n}\langle M_j^1 - M_j^2, N_j \rangle^{2}\right)^{\frac{1}{2}}\\
&\leq u\left(\sum_{j=1}^{n}\left\|M_j^1 - M_j^2\right\|_{2}^{2}\left\|N_j \right\|_{2}^{2}\right)^{\frac{1}{2}}\\
&\leq ud_{F}\left(\mathcal{N}\right)\left\|M^{1}-M^{2}\right\|_{2\to2}
\end{align*}
Similarly with probability $1 - 2e^{-u^2/2}$,
\begin{align*}
\left| X_{(M, N^1)} - X_{(M, N^2)} \right|\leq ud_{F}\left(\mathcal{M}\right)\left\|N^{1}-N^{2}\right\|_{2\to2}
\end{align*}
Using Theorem~\ref{theo:double_tree}, with probability $1 - 4e^{-u^2/2}$, 
\begin{align*}
\sup_{M \in \mathcal{M}, N \in \mathcal{N}} 
\left| \sum_{j=1}^n \g_j \langle M_j, N_j \rangle \right|
\lesssim u \left( d_{F} (\cN)\gamma_2(\mathcal{M}, \|\cdot\|_{2\rightarrow 2}) + d_{F} (\cM)\gamma_2(\mathcal{N}, \|\cdot\|_{2\rightarrow 2}) \right).
\end{align*}
so we have
\begin{align}
    \mathbb{E}_{\g}\left[\sup_{M \in \mathcal{M}, N \in \mathcal{N}} 
\left| \sum_{j=1}^n \g_j \langle M_j, N_j \rangle \right|\right]\lesssim d_{F} (\cN)\gamma_2(\mathcal{M}, \|\cdot\|_{2\rightarrow 2}) + d_{F} (\cM)\gamma_2(\mathcal{N}, \|\cdot\|_{2\rightarrow 2}) .\label{eq:rest_2}
\end{align}
For the second term, for each $M\in\mathcal{M},N\in\mathcal{N}$, since $\varepsilon_{j}$ are Rademacher random variables, therefore with probability at least $1-2e^{-u^{2}}$,
\begin{align*}
    \left| \sum_{j=1}^n \varepsilon_j \langle M_j, N_j \rangle \right|&\leq u\left(\sum_{j=1}^{n}\left\langle M_{j},N_{j}\right\rangle^{2}\right)^{\frac{1}{2}}\leq u\left(\sum_{j=1}^{n}\left\|M_{j}\right\|_{2}^{2}\left\|N_{j}\right\|_{2}^{2}\right)^{\frac{1}{2}}\\&\leq\min\left\{d_{F}\left(\mathcal{M}\right)d_{2\to2}\left(\mathcal{N}\right),d_{F}\left(\mathcal{N}\right)d_{2\to2}\left(\mathcal{M}\right)\right\}.
\end{align*}
According to \cite[Proposition 2.5.2]{vers18},
\begin{align*}
    \left(\mathbb{E}_{\varepsilon}\left[\left| \sum_{j=1}^n \varepsilon_j \langle M_j, N_j \rangle \right|^{p}\right]\right)^{1/p}\lesssim \sqrt{p}\min\left\{d_{F}\left(\mathcal{M}\right)d_{2\to2}\left(\mathcal{N}\right),d_{F}\left(\mathcal{N}\right)d_{2\to2}\left(\mathcal{M}\right)\right\},
\end{align*}
so
\begin{align}
    \sup_{M\in\mathcal{M},N\in\mathcal{N}}\left(\mathbb{E}_{\varepsilon}\left[\left| \sum_{j=1}^n \varepsilon_j \langle M_j, N_j \rangle \right|^{p}\right]\right)^{1/p}\lesssim \sqrt{p}\min\left\{d_{F}\left(\mathcal{M}\right)d_{2\to2}\left(\mathcal{N}\right),d_{F}\left(\mathcal{N}\right)d_{2\to2}\left(\mathcal{M}\right)\right\}\label{eq:rest_3}
\end{align}
Substituting \eqref{eq:rest_2} and \eqref{eq:rest_3} into \eqref{eq:rest_1}, we have
\begin{align}
&\left\| \sup_{M \in \mathcal{M}, N \in \mathcal{N}} 
\left| \sum_{j=1}^n \varepsilon_j \langle M_j, N_j \rangle \right| \right\|_{L_p}\notag\\
\lesssim&\mathbb{E}_{\g}\left[\sup_{M \in \mathcal{M}, N \in \mathcal{N}} 
\left| \sum_{j=1}^n \g_j \langle M_j, N_j \rangle \right|\right]+\sup_{M\in\mathcal{M},N\in\mathcal{N}}\left(\mathbb{E}_{\varepsilon}\left[\left| \sum_{j=1}^n \varepsilon_j \langle M_j, N_j \rangle \right|^{p}\right]\right)^{1/p}\notag\\
\lesssim& d_{F} (\cN)\gamma_2(\mathcal{M}, \|\cdot\|_{2\rightarrow 2}) + d_{F} (\cM)\gamma_2(\mathcal{N}, \|\cdot\|_{2\rightarrow 2}) +\sqrt{p}\min\left\{d_{F}\left(\mathcal{M}\right)d_{2\to2}\left(\mathcal{N}\right),d_{F}\left(\mathcal{N}\right)d_{2\to2}\left(\mathcal{M}\right)\right\}\label{eq:diagonal_3}
\end{align}
Substituting \eqref{eq:diagonal_2} and \eqref{eq:diagonal_3} into \eqref{eq:diagonal_1_app}, we get 
\begin{align*}
    \left\| D_{\cM,\cN}(\bxi) \right\|_{L_p} \lesssim \Bigg[&\gamma_2(\cM,\| \cdot \|_{2 \rightarrow 2}) \cdot \Big[\gamma_2(\cN, \| \cdot \|_{2 \rightarrow 2}) + d_F(\cN) + \sqrt{p} d_{2 \rightarrow 2}(\cN)~\Big]\nonumber \\
    & \quad + \gamma_2(\cN,\| \cdot \|_{2 \rightarrow 2}) \cdot \Big[\gamma_2(\cM, \| \cdot \|_{2 \rightarrow 2}) + d_F(\cM) + \sqrt{p} d_{2 \rightarrow 2}(\cM)~\Big] \nonumber
    \\ & \quad + \sqrt{p} \min \Big\{  d_F(\cM) \cdot d_{2 \rightarrow 2}(\cN),d_F(\cN) \cdot d_{2 \rightarrow 2}(\cM) \Big\} \nonumber\\
        & \qquad\qquad + p \cdot d_{2 \rightarrow 2}(\cM) \cdot d_{2 \rightarrow 2}(\cN)~.\Bigg]    
\end{align*}
That completes the proof of Theorem~\ref{theo:diagonalLpBound}.}

\section{Johnson Lindenstrauss and Restricted Isometry Property}
\label{sec:JL-RIP}
\subsection{Proof of Proposition~\ref{prop:J-L}}
\begin{proof}
    Suppose the entries of $\tilde{X}$ be i.i.d.\ standard normal. For any $u \in \cU$ we define both $M \in \cM$ and $N \in \cN$ as follows:
    \begin{align*}
        M = N = \frac{1}{\sqrt{n}}\begin{bmatrix}
                                u^\top & 0 & \cdots & 0 \\
                                0 & u^\top & \cdots & 0 \\
                                \vdots & \ddots & \ddots & \vdots \\
                                0 & 0 & \cdots & u^\top
                               \end{bmatrix},
        \end{align*}
\end{proof}
Further we define the random vector $\bxi = [\tilde{X}_{1,:}, \tilde{X}_{2,:}, \ldots, \tilde{X}_{n,:}]^\top$ where $\tilde{X}_{j,:}$ is the $j$-th row of $\tilde{X}$. Observe that 
    \begin{align*}
        \sup_{u \in \cU}|\|Xu\|_2^2 - \E \|Xu\|_2^2| &= \sup_{M \in \cM, N \in \cN}|\bxi^\top M ^\top N \bxi - \E \bxi^\top M ^\top N \bxi|. \\
        &= \sup_{M \in \cM} |\; \|M \bxi\|_2^2 - \E\|M \bxi\|_2^2\;|
    \end{align*}
where $\cU$ is a set of $N$ vectors. Then,
\begin{align*}
        d_{F}(\cM) &= \sup_{M \in \cM} \|M\|_F =  \sup_{u \in \cU} \frac{1}{\sqrt{n}} \Bigg(\sqrt{\sum_{i=1}^{n} \|u\|^2_2}\Bigg) = \sup_{u \in \cU} \|u\|_2 \\
        d_{2 \rightarrow 2}(\cM) &= \sup_{M \in \cM} \|M\|_{2 \to 2} = \sup_{M \in \cM} \sup_{w \in S^d} M w = \sup_{u \in \cU} \frac{1}{\sqrt{n}} \|u\|_2
    \end{align*}
Further using the fact that the $\gamma_2$ functional can be upper bounded by the Gaussian width up to constants \citep{tala05,tala14} we get
    \begin{align*}
        \gamma_{2}(\cM, \|\cdot\|_2) \lesssim \omega(\cM) \lesssim \frac{1}{\sqrt{n}} \omega(\cU)
    \end{align*}
Now the Gaussian width of $\cU$ is $\omega(\cU) \lesssim \sup_{u \in \cU}\|u\|_2\sqrt{\log N }$ \citep{vers18}. Therefore,

\begin{align*}
    {\color{Red} W} &{\color{Red} = 2\gamma_2(\cM,\| \cdot \|_{2 \rightarrow 2}) \cdot \Big[\gamma_2(\cM, \| \cdot \|_{2 \rightarrow 2}) + d_F(\cM) ~\Big]} \\
    &{\color{Red}\lesssim \frac{2}{\sqrt{n}} \omega(\cU) \left[ \frac{1}{\sqrt{n}} \omega(\cU) + \sup_{u \in \cU}\|u\|_2 \right]}\\
    &{\color{Red}\lesssim 2\sup_{u \in \cU}\|u\|_2^2 \left[ \frac{1}{{n}} \log N + \frac{1}{\sqrt{n}}\sqrt{\log N} \right]}
    \end{align*}
    \begin{align*}
    {\color{Blue} V}&{\color{Blue} = 2\gamma_2(\cM,\| \cdot \|_{2 \rightarrow 2})d_{2 \rightarrow 2}(\cN) +  d_F(\cM) \cdot d_{2 \rightarrow 2}(\cN)} \\
    &{\color{Blue} \lesssim \frac{1}{\sqrt{n}}\omega(\cU)\; \frac{1}{\sqrt{n}}\sup_{u \in \cU}\|u\|_2 \;+\; \frac{1}{\sqrt{n}}\sup_{u \in \cU}\|u\|_2} \\
    &{\color{Blue} \lesssim \frac{1}{n} \sup_{u \in \cU}\|u\|_2^2 \sqrt{\log N} \;\;+\;\; \frac{1}{\sqrt{n}} \sup_{u \in \cU}\|u\|_2}
    \end{align*}
\begin{align*}
    {\color{darkgreen} U} &{\color{darkgreen} = d_{2 \to 2}(\cM) \; d_{2 \to 2}(\cN).}{\color{darkgreen} = \frac{1}{\sqrt{n}} \sup_{u \in \cU}\|u\|_2 \; \frac{1}{\sqrt{n}} \sup_{u \in \cU}\|u\|_2 = \frac{1}{n}\sup_{u \in \cU}\|u\|_2^2.}
\end{align*}
Note that $\bxi$ satisfies Assumption~\ref{asmp:xi} and therefore invoking Theorem ~\ref{thm:cross_product} we have
    \begin{align*}
        &P\Bigg\{\sup_{u \in \cU}|\|Xu\|_2^2 - \E \|Xu\|_2^2| \gtrsim \color{Red}\sup_{u \in \cU}\|u\|_2^2 \left[ \frac{1}{{n}} \log N + \frac{1}{\sqrt{n}}\sqrt{\log N} \right] + \epsilon \sup_{u \in \cU}\|u\|_2^2 \Bigg\}\\
        & \lesssim \exp \left(  - \min \left\{ \frac{\epsilon^2 \sup_{u \in \cU}\|u\|_2^4}{\big({\color{Blue}  \frac{1}{n} \sup_{u \in \cU}\|u\|_2^2 \sqrt{\log N} + \frac{1}{\sqrt{n}} \sup_{u \in \cU}\|u\|_2}\big)^2}, \frac{\epsilon \sup_{u \in \cU}\|u\|_2^2}{{\color{Green}\frac{1}{n}\sup_{u \in \cU}\|u\|_2^2}} \right\} \right)
    \end{align*}
Now choosing $n = \Omega(\varepsilon^{-2} \log N)$ immediately proves the theorem.
\subsection{Proof of Proposition~\ref{prop:restricted_inner_prod}}
    Suppose the entries of $\tilde{X}$ be i.i.d.\ standard normal. For any $u \in \cU$ and $v \in \cV$ we define the corresponding $M \in \cM$ and $N \in \cN$ as follows:
    \begin{align*}
        M = \frac{1}{\sqrt{n}}\begin{bmatrix}
                                u^\top & 0 & \cdots & 0 \\
                                0 & u^\top & \cdots & 0 \\
                                \vdots & \ddots & \ddots & \vdots \\
                                0 & 0 & \cdots & u^\top
                               \end{bmatrix},
        \quad N = \frac{1}{\sqrt{n}}\begin{bmatrix}
                                v & 0 & \cdots & 0 \\
                                0 & v & \cdots & 0 \\
                                \vdots & \ddots & \ddots & \vdots \\
                                0 & 0 & \cdots & v
                               \end{bmatrix}
    \end{align*}
    Further we define the random vector $\bxi = [\tilde{X}_{1,:}, \tilde{X}_{2,:}, \ldots, \tilde{X}_{n,:}]^\top$ where $\tilde{X}_{j,:}$ is the $j$-th row of $\tilde{X}$. Observe that 
    \begin{align*}
        \sup_{u \in \cU, v \in \cV}|\langle Xu, Xv \rangle - \E \langle Xu, Xv \rangle| = \sup_{M \in \cM, N \in \cN}|\bxi^\top M ^\top N \bxi - \E \bxi^\top M ^\top N \bxi|. 
    \end{align*}

    We can now compute the various complexity measures required in Theorem~\ref{thm:cross_product} as follows:
    \begin{align*}
        d_{F}(\cM) &= \sup_{M \in \cM} \|M\|_F =  \sup_{u \in \cU} \frac{1}{\sqrt{n}} \Bigg(\sqrt{\sum_{i=1}^{n} \|u\|^2_2}\Bigg) = \sup_{u \in \cU} \|u\|_2 = 1\\
        d_{2 \rightarrow 2}(\cM) &= \sup_{M \in \cM} \|M\|_{2 \to 2} = \sup_{M \in \cM} \sup_{w \in S^d} M w = \sup_{u \in \cU} \frac{1}{\sqrt{n}} \|u\|_2 = \frac{1}{\sqrt{n}}
    \end{align*}
    Further using the fact that the $\gamma_2$ functional can be upper bounded by the Gaussian width up to constants \citep{tala05,tala14} we get
    \begin{align*}
        \gamma_{2}(\cM, \|\cdot\|_2) \lesssim \omega(\cM) \lesssim \frac{1}{\sqrt{n}} \omega(\cU)
    \end{align*}
    Similarly, $d_{F}(\cN) = 1$, $d_{2}(\cN) = \frac{1}{\sqrt{n}}$ and $\gamma_{2}(\cN, \|\cdot\|_2) \lesssim \frac{1}{\sqrt{n}} \omega(\cV)$. We can now compute ${\color{Red} W}, {\color{Blue} V}$ and ${\color{darkgreen} U}$ as follows:
    \begin{align*}
    {\color{Red} W} &{\color{Red} = \gamma_2(\cM,\| \cdot \|_{2 \rightarrow 2}) \cdot \Big[\gamma_2(\cN, \| \cdot \|_{2 \rightarrow 2}) + d_F(\cN) ~\Big]} \\
    & \qquad\qquad\qquad\qquad\qquad\qquad  {\color{Red} + \gamma_2(\cN,\| \cdot \|_{2 \rightarrow 2}) \cdot \Big[\gamma_2(\cM, \| \cdot \|_{2 \rightarrow 2}) + d_F(\cM) ~\Big],} \\
    &{\color{Red}\lesssim \frac{1}{\sqrt{n}} \omega(\cU) \left[ \frac{1}{\sqrt{n}} \omega(\cV) + 1 \right] + \frac{1}{\sqrt{n}} \omega(\cU) \left[ \frac{1}{\sqrt{n}} \omega(\cV) + 1 \right]}\\
    &{\color{Red}= \frac{2}{{n}} \omega(\cU)\omega(\cV) + \frac{1}{\sqrt{n}}(\omega(\cU) + \omega(\cV))}
    \end{align*}
    \begin{align*}
    {\color{Blue} V}&{\color{Blue} = \gamma_2(\cM,\| \cdot \|_{2 \rightarrow 2})d_{2 \rightarrow 2}(\cN) + \gamma_2(\cN,\| \cdot \|_{2 \rightarrow 2})d_{2 \rightarrow 2}(\cM)} \\
    & \qquad\qquad\qquad\qquad\qquad\qquad {\color{Blue} + \min \Big\{ d_F(\cM) \cdot d_{2 \rightarrow 2}(\cN), d_F(\cN) \cdot  d_{2 \rightarrow 2}(\cM) \Big\}} \\
    &{\color{Blue} = \frac{1}{\sqrt{n}}\omega(\cU)\; \frac{1}{\sqrt{n}} \;+\; \frac{1}{\sqrt{n}}\omega(\cV) \; \frac{1}{\sqrt{n}} \; + \; \min \Bigg\{\frac{1}{\sqrt{n}},\frac{1}{\sqrt{n}}\Bigg\}} \\
    &{\color{Blue} = \frac{1}{n} (\omega(\cU) + \omega(\cV)) \;\;+\;\; \frac{1}{\sqrt{n}}}
    \end{align*}
\begin{align*}
    {\color{darkgreen} U} &{\color{darkgreen} = d_{2 \to 2}(\cM) \; d_{2 \to 2}(\cN).}\\
    &{\color{darkgreen} = \frac{1}{\sqrt{n}} \; \frac{1}{\sqrt{n}} = \frac{1}{n}.}
\end{align*}
    
    Note that $\bxi$ satisfies Assumption~\ref{asmp:xi} and therefore invoking Theorem ~\ref{thm:cross_product} we have
    \begin{align*}
        P\Bigg\{\sup_{u \in \cU, v \in \cV}|\langle Xu, Xv \rangle - \E \langle Xu, Xv \rangle | &\geq {\color{Red} \frac{1}{{n}} \omega(\cU)\omega(\cV) + \frac{1}{\sqrt{n}}(\omega(\cU) + \omega(\cV))} + \epsilon \Bigg\}\\
        & \lesssim \exp \left(  - \min \left\{ \frac{\epsilon^2}{\big({\color{Blue} \frac{1}{n} (\omega(\cU) + \omega(\cV)) +  \frac{1}{\sqrt{n}}}\big)^2}, \frac{\epsilon}{{\color{Green}\frac{1}{n}}} \right\} \right)
    \end{align*}

    Choosing $\displaystyle n = \Omega\bigg(\frac{1}{\epsilon^2} \big( \omega(\cU) + \omega(\cV) \big)^2\bigg)$ in the above we get
    \begin{align*}
        \P\Bigg\{\sup_{u \in \cU, v \in \cV}|\langle Xu, Xv \rangle - \E \langle Xu, Xv \rangle | &\gtrsim \frac{\epsilon^2}{\Big( \omega(\cU) + \omega(\cV)\Big)^2} {\color{Red} \omega(\cU)\omega(\cV)} + {\frac{\epsilon({\color{Red}\omega(\cU) + \omega(\cV)})}{{( \omega(\cU) + \omega(\cV))}}} + \epsilon \Bigg\}\\
        & \lesssim \exp \left(  - \min \left\{ \frac{\epsilon^2}{\big({\color{Blue} \frac{1}{n} (\omega(\cU) + \omega(\cV)) +  \frac{1}{\sqrt{n}}}\big)^2}, \frac{\epsilon}{{\color{Green}\frac{1}{n}}} \right\} \right)\\
        & \lesssim \exp \left(- \min \left\{ \frac{\epsilon^2}{\big({\color{Blue} \epsilon^2 +  \frac{\epsilon}{\omega(\cU) + \omega(\cV)}}\big)^2}, \frac{\epsilon}{{\color{Green}\frac{\epsilon^2}{(\omega(\cU) + \omega(\cV))^2}}} \right\} \right)\\
        & \lesssim \exp \left(- (\omega(\cU) + \omega(\cV))^2 \right)
    \end{align*}
    Now note that 
    \begin{align*}
        \frac{\epsilon^2}{\Big( \omega(\cU) + \omega(\cV)\Big)^2} {\color{Red} \omega(\cU)\omega(\cV)} + {\frac{\epsilon({\color{Red}\omega(\cU) + \omega(\cV)})}{{( \omega(\cU) + \omega(\cV))}}} + \epsilon \geq \epsilon^2 + 2\epsilon \gtrsim \epsilon
    \end{align*}
    where the first inequality uses the fact that $\big(\omega(\cU) + \omega(\cV)\big)^2 \geq \omega(\cU)\omega(\cV)$. Therefore with probability $1 - \exp \left(- c(\omega(\cU) + \omega(\cV))^2 \right)$
    \begin{align*}
        |\langle Xu, Xv \rangle - \E \langle Xu, Xv \rangle| \leq \epsilon \|u\|_2\|v\|_2 \quad \text{holds for all $u \in \cU, v \in \cV$.}
    \end{align*}

\section{Application in Distributed Learning}
\label{sec:app_distributed}

\subsection{Sketching-based Distributed Learning}
We outline the sketching‐based distributed learning framework in Algorithm~\ref{alg:sketch}. Each client receives a random seed from the server to initialize the local parameters $\theta_{c,1}$, and generate a sketching matrix $R$. At each local step $k\in[1,\dots,K]$, each client performs local gradient descent (GD) over their local dataset $\mathcal{D}_c$. At each communication round, the client accumulates the changes over $K$-local steps, sketches the local updates, and sends the sketched update to the server. The server then aggregates the sketched changes and sends the aggregated sketched updates back to the clients. To update the local parameters, each client needs to recover an unbiased estimate of the true vector from the aggregated sketched update. We call this the desk (desketching) operation (Line 9), for which we use the transpose of the sketching matrix $R$. Each client then desketches the received aggregated sketched updates by applying desk and updates their local parameters. We refer to the sketching and desketching operations using the \text{sk} and \text{desk} operators defined as:
\begin{align*}
\texttt{sk}   &:= R \in \mathbb{R}^{b\times p}
\quad(\text{Sketching}),\\
\texttt{desk} &:= R^\top \in \mathbb{R}^{p\times b}
\quad(\text{Desketching}).
\end{align*}
Denote $R_{t}\in\mathbb{R}^{b\times p}$ the sketching matrix at round $t\in[T]$.

\textbf{Choice of sketching matrix:} We use a $(1/\sqrt b)$‐sub‐Gaussian matrix as the choice of sketching matrix. We say $R\in\mathbb{R}^{b\times p}$ is a $(1/\sqrt b)$‐sub‐Gaussian matrix if each row $R_i$ is an independent mean‐zero, sub‐Gaussian isotropic random‐vector such that $\|R_i\|_{\psi_2}\le1/\sqrt b$. We assume $\mathbb{E}[R^\top R]=I_{p\times p}$. From the above definition, we can see that for $g_1,g_2\in\mathbb{R}^p$
\begin{align*}
R(g_1+g_2)                          &= Rg_1 + Rg_2
\quad(\text{Linearity}),\notag\\
 \mathbb{E}_{R\sim\Pi}\left[R^\top R\,g\right] &= g
 \quad(\text{Unbiasedness}).
 \end{align*}
 \begin{algorithm}[t]
 \caption{Sketching‐Based Distributed Learning.}
 \label{alg:sketch}
 \textbf{Hyperparameters:} server learning rate $\eta_{\text{global}}$, local learning rate $\eta_{\text{local}}$.\\
 \textbf{Inputs:} local datasets $\mathcal{D}_c$ of size $n_c$ for clients $c=1,\dots,C$, number of communication rounds $T$.\\
 \textbf{Output:} final model $\theta_T$.
 \begin{algorithmic}[1]
 \STATE Broadcast a random SEED to the clients.
 \FOR{$t = 0,\dots,T$}
   \STATE \textbf{On Client Nodes:}
   \FOR{$c = 1,\dots,C$}
     \IF{$t = 0$}
       \STATE Initialize the local model $\theta_{0}=\theta_{c,0,0}\in\mathbb{R}^p$.
     \ELSE
       \STATE Receive sk$(\bar\Delta_{t-1})$ from the server.
       \STATE Desketch and update the model parameters $\theta_t \leftarrow \theta_{t-1} + \texttt{desk}(\texttt{sk}(\bar\Delta_{t-1}))$.
       \STATE Assign the local model’s parameters $\theta_{c,t,0} \leftarrow \theta_t$ to be updated locally.
     \ENDIF
     \FOR{$k = 1,\dots,K$}
       \STATE $\theta_{c,t,k}\leftarrow \theta_{c,t,k-1} - \eta_{\text{local}}\cdot\nabla\mathcal{L}_{c}(\theta_{c,t,k})$
     \ENDFOR
     \STATE $\Delta_{c,t}\leftarrow \theta_{t} - \theta_{c,t,K}$
     \STATE Send $\texttt{sk}(\Delta_{c,t})$ to the server.
   \ENDFOR
   \vspace{0.5em}
   \STATE \textbf{On the Server Node:}
   \STATE Receive $\texttt{sk}(\Delta_{c,t})$ from clients $c=1,\dots,C$.
   \STATE Aggregate: $\texttt{sk}(\bar\Delta_t)\leftarrow \eta_{\text{global}}\cdot\frac{1}{C}\sum_{c=1}^C\texttt{sk}(\Delta_{c,t})$
   \STATE Broadcast $\texttt{sk}(\bar\Delta_t)$ to the clients.
 \ENDFOR
 \end{algorithmic}
 \end{algorithm}

In line with standard works~\cite{shrivastava2024sketching,banerjee2023restricted}, we study a fully-connected feedforward neural network $f$ of depth $L$, where each layer $l\in[L]:={1,\dots,L}$ has associated activations $\alpha^{(l)}$, defined recursively as:
\begin{align*}
  \alpha^{(0)}(x) &= x,\\
  \alpha^{(l)}(x)
  &= \phi\left(\tfrac{1}{\sqrt{m_{l-1}}}\,W_t^{(l)}\,\alpha^{(l-1)}(x)\right),
  \quad \forall\,l\in[L],\\
  f(\theta; x)
  &= \alpha^{(L+1)}(x)
  = \frac{1}{\sqrt{m_L}}\;v_t^{\!\top}\,\alpha^{(L)}(x),
  \end{align*}
where $W_t^{(l)}\in\mathbb{R}^{m_l\times m_{l-1}}$ denotes the weight matrix at the $l^{\text{th}}$ layer, $v_t\in\mathbb{R}^{m_L}$ is the output-layer vector at time $t$, and $\phi$ is a smooth pointwise activation function. The input dimension is $m_0 = \dim(x) = d$. The complete parameter vector at iteration $t$ is given by
\begin{align*}
\theta_t = \left(\left(\vec{W}_t^{(1)}\right)^\top, \dots, \left(\vec{W}_t^{(L)}\right)^\top, v_t^\top\right)^\top \in \mathbb{R}^p.
\end{align*}
For notational simplicity, we assume all hidden layers have uniform width $m$, i.e., $m_l = m$ for every $l\in[L]$, so that the total number of parameters is $p = (L-1)m^2 + m d + m$. We restrict our focus to scalar-valued networks $f(\theta;x)\in\mathbb{R}$, though the analysis extends naturally to multi-output settings.

We now state the standard assumptions on the activation function, loss function, and initialization, which are satisfied by most commonly used choices in practice.

\begin{asmp}[Activation function]
The activation $\phi$ is 1-Lipschitz and $\beta_\phi$-smooth, that is, $|\phi'|\le 1$ and $|\phi''|\le \beta_\phi$.
\end{asmp}

\begin{asmp}[Initialization]
The initial parameters are drawn as $w^{(l)}_{0,ij} \sim \mathcal{N}(0, \sigma_0^2)$ for each layer $l\in[L]$, where $\sigma_0 = \frac{\sigma_1}{2\left(1 + \frac{\sqrt{\log m}}{\sqrt{2m}}\right)}$ with $\sigma_1 > 0$. The final layer vector $v_0$ is a random unit vector with $|v_0|_2 = 1$.
\end{asmp}

\begin{asmp}[Loss function]
Let $\ell_{i,c} = \ell(y_{i,c}, \hat{y}{i,c})$ denote the per-example loss, with first and second derivatives $\ell'{i,c} = \frac{d\ell_{i,c}}{d\hat{y}{i,c}}$ and $\ell''{i,c} = \frac{d^2\ell_{i,c}}{d\hat{y}{i,c}^2}$. The loss satisfies:
(i) Lipschitz continuity: $|\ell'{i,c}| \le c_\ell$, and
(ii) Smoothness: $\ell''{i,c} \le c_s$,
for some constants $c\ell, c_s > 0$.
\end{asmp}
 
 \subsection{Convergence Analysis for Sketching-based Distributed Learning}
 We begin by presenting a set of standard assumptions that are widely adopted in the literature on first-order stochastic methods. 
 \begin{asmp}[Bounded Loss Gradients]\label{asmp:gradient_norm}
 There exists a constant $G \geq 0$, such that for every $\left(\mathbf{x},y\right)$, $\left\|\nabla \ell(\theta;(x,y))\right\|_{2} \le G$.
 \label{asmp:bcg}
 \end{asmp}
 \begin{asmp}[\textbf{Anistropic Loss Hessian}]\label{assump:hessian-spectrum-app}
For any loss Hessian $H_{i,c,t}$, assume there exists a fixed positive definite $\mathbf{H}$ such that $-\mathbf{H}\preceq H_{i,c,t}\preceq \mathbf{H}$. Let $\Lambda_{1},\;\dots,\;\Lambda_{p}$ be the eigenvalues of $\mathbf{H}$
and define $\Lambda_{\max} =\;\max_j \bigl|\Lambda_{j}\bigr|$.
Then there exists a constant $\kappa = \mathcal{O}(1)$ such that
$\sum_{j=1}^p \bigl|\Lambda_{j}\bigr|
\;\le\;
\kappa\,\Lambda_{\max}$.
\end{asmp}
 From Theorem~\ref{thm:cross_product}, we can derive the following sketching guarantee.
 \begin{theorem}\label{thm:sketching_guarantee}
 Let $\mathcal{G}$ and $\mathcal{H}$ be set of $d$-dimensional vectors respectively, and let $R \in \R^{b \times d}$ denote a random $\frac{1}{\sqrt{b}}$-sub-Gaussian matrix. Denote $d_{2}\left(\mathcal{G}\right)=\sup_{g\in\mathcal{G}}\left\|g\right\|_{2}$ and $d_{2}\left(\mathcal{H}\right)=\sup_{h\in\mathcal{H}}\left\|h\right\|_{2}$. We define
 \begin{align*}
     A&= \frac{w(\mathcal{G})w(\mathcal{H})}{b}+\frac{w(\mathcal{G})d_{2}(\mathcal{H})+w(\mathcal{H})d_{2}(\mathcal{G})}{\sqrt{b}}, \\
     B&= \frac{w(\mathcal{G})d_{2}(\mathcal{H})+w(\mathcal{H})d_{2}(\mathcal{G})}{b}+\frac{d_{2}(\mathcal{G})d_{2}(\mathcal{H})}{\sqrt{b}}\\
     C & = \frac{d_{2}(\mathcal{G})d_{2}(\mathcal{H})}{b}.
 \end{align*}
 Then, for $\epsilon > 0$,
 \begin{align*}
 \mathbb{P}\left(\sup_{g\in\mathcal{G},h\in\mathcal{H}}\left|g^{\top}R^{\top}Rh-g^{\top}h\right|\geq c_1 A + \epsilon\right) 
 \leq 2 \exp\left( -c_2 \min\left\{ \frac{\epsilon^2}{B^2}, \frac{\epsilon}{C} \right\} \right).
 \end{align*}
 where the constants $c_1, c_2$ are absolute constants. Equivalently, with probability at least $1-2\delta$,
 \begin{align}
     \sup_{g\in\mathcal{G},h\in\mathcal{H}}\left|g^{\top}R^{\top}Rh-g^{\top}h\right|\lesssim Z(\mathcal{G},\mathcal{H},\delta):= A+\sqrt{\log(1/\delta)}B+\log(1/\delta)C\label{ineq:distributed_upper}
 \end{align}
 \end{theorem}
 \begin{proof}
 Denote $r=\left[\sqrt{b}R_{1,:},\sqrt{b}R_{2:},\dots,\sqrt{b}R_{b:}\right]^{\top}$ be a $1$-sub-Gaussian random vector of
 length $bd$ by concatenating the rows of $R$ and rescaling by $\sqrt{b}$. For any $g\in\mathcal{G}$ and $h\in\mathcal{H}$, define
 \begin{align*}
     M_{g}:=\frac{1}{\sqrt{b}}\begin{bmatrix}
 g^T & 0       & \cdots & 0       \\
 0       & g^T & \cdots & 0       \\
 \vdots  & \ddots  & \ddots & \vdots  \\
 0       & 0       & \cdots & g^T
 \end{bmatrix}, N_{h}=\frac{1}{\sqrt{b}}\begin{bmatrix}
 h^T & 0       & \cdots & 0       \\
 0       & h^T & \cdots & 0       \\
 \vdots  & \ddots  & \ddots & \vdots  \\
 0       & 0       & \cdots & h^T
 \end{bmatrix}
 \end{align*}
 so $M_{g}\in\mathbb{R}^{b\times bd}$ and $N_{h}\in\mathbb{R}^{b\times bd}$ are block diagonal matrices, and $g^{\top}R^{\top}Rh=r^{\top}M_{g}^{\top}N_{h}r$. Denote $\mathcal{M}_{\mathcal{G}}$ and $\mathcal{N}_{\mathcal{H}}$ be the sets of such $M_{g}$ and $N_{h}$ respectively. According to results in \cite{tala14}, we can see that $\gamma_{2}\left(\mathcal{M}_{\mathcal{G}},\left\|\cdot\right\|_{2\to2}\right)\leq C\frac{w(\mathcal{G})}{\sqrt{b}}$, $\gamma_{2}\left(\mathcal{N}_{\mathcal{H}},\left\|\cdot\right\|_{2\to2}\right)\leq C\frac{w(\mathcal{H})}{\sqrt{b}}$. In addition, we can get that $d_{F}\left(\mathcal{M}_{\mathcal{G}}\right)=d_{2}\left(\mathcal{G}\right)$, $d_{F}\left(\mathcal{N}_{\mathcal{H}}\right)=d_{2}\left(\mathcal{H}\right)$, $d_{2\to2}\left(\mathcal{M}_{\mathcal{G}}\right)=\frac{1}{\sqrt{b}}d_{2}\left(\mathcal{G}\right)$, $d_{2\to2}\left(\mathcal{N}_{\mathcal{H}}\right)=\frac{1}{\sqrt{b}}d_{2}\left(\mathcal{H}\right)$. Therefore, 
 \begin{align*}
     W&= \gamma_2(\mathcal{M}_{\mathcal{G}},\| \cdot \|_{2 \rightarrow 2}) \cdot \Big[\gamma_2(\mathcal{N}_{\mathcal{H}}, \| \cdot \|_{2 \rightarrow 2}) + d_F(\mathcal{N}_{\mathcal{H}}) ~\Big] + \gamma_2(\mathcal{N}_{\mathcal{H}},\| \cdot \|_{2 \rightarrow 2}) \cdot \Big[\gamma_2(\mathcal{M}_{\mathcal{G}}, \| \cdot \|_{2 \rightarrow 2}) + d_F(\mathcal{M}_\mathcal{G}) ~\Big] \\
     &\lesssim\frac{w(\mathcal{G})}{\sqrt{b}}\cdot\left(\frac{w(\mathcal{H})}{\sqrt{b}}+d_{2}\left(\mathcal{H}\right)\right)+\frac{w(\mathcal{H})}{\sqrt{b}}\left(\frac{w(\mathcal{G})}{\sqrt{b}}+d_{2}\left(\mathcal{G}\right)\right)\\
     &\lesssim\frac{w(\mathcal{G})w(\mathcal{H})}{b}+\frac{w(\mathcal{G})d_{2}\left(\mathcal{H}\right)+w(\mathcal{H})d_{2}(\mathcal{G})}{\sqrt{b}}=A,\\
     V&= \gamma_2(\mathcal{M}_\mathcal{G},\| \cdot \|_{2 \rightarrow 2})d_{2 \rightarrow 2}(\mathcal{N}_\mathcal{H}) + \gamma_2(\mathcal{N}_\mathcal{H},\| \cdot \|_{2 \rightarrow 2})d_{2 \rightarrow 2}(\mathcal{M}_\mathcal{G})\\
     &\qquad\qquad\qquad\qquad + \min \Big\{ d_F(\mathcal{M}_\mathcal{G}) \cdot d_{2 \rightarrow 2}(\mathcal{N}_\mathcal{H}), d_F(\mathcal{N}_\mathcal{H}) \cdot  d_{2 \rightarrow 2}(\mathcal{M}_\mathcal{G}) \Big\}\\
     &=\frac{w(\mathcal{G})}{\sqrt{b}}\cdot\frac{d_{2}(\mathcal{H})}{\sqrt{b}}+\frac{w(\mathcal{H})}{\sqrt{b}}\cdot\frac{d_{2}(\mathcal{G})}{\sqrt{b}}+\min\left\{d_{2}(\mathcal{G})\cdot\frac{d_{2}(\mathcal{H})}{\sqrt{b}},d_{2}(\mathcal{H})\cdot\frac{d_{2}(\mathcal{G})}{\sqrt{b}}\right\}\\
     &=\frac{w(\mathcal{G})d_{2}(\mathcal{H})+w(\mathcal{H})d_{2}(\mathcal{G})}{b}+\frac{d_{2}(\mathcal{G})d_{2}(\mathcal{H})}{\sqrt{b}}=B,\\
     U & = d_{2 \to 2}(\mathcal{M}_\mathcal{G}) \; d_{2 \to 2}(\mathcal{N}_\mathcal{H})=\frac{d_{2}(\mathcal{G})}{\sqrt{b}}\cdot\frac{d_{2}(\mathcal{H})}{\sqrt{b}}=\frac{d_{2}(\mathcal{G})d_{2}(\mathcal{H})}{b}=C.
 \end{align*}

 According to Theorem~\ref{thm:cross_product}, we can see that
 \begin{align*}
     &\mathbb{P}\left(\sup_{g\in\mathcal{G},h\in\mathcal{H}}\left|g^{\top}R^{\top}Rh-g^{\top}h\right|\geq c_1 A + \epsilon\right) \\
     =&\mathbb{P}\left(\sup_{M_{g}\in\mathcal{M}_{\mathcal{G}},N_{h}\in\mathcal{N}_{\mathcal{H}}}\left|r^{\top}M_{g}^{\top}N_{h}r-M_{g}^{\top}N_{h}^{\top}\right|>c_{1}A+\epsilon\right)\\
     \leq&\mathbb{P}\left(\sup_{M_{g}\in\mathcal{M}_{\mathcal{G}},N_{h}\in\mathcal{N}_{\mathcal{H}}}\left|r^{\top}M_{g}^{\top}N_{h}r-M_{g}^{\top}N_{h}^{\top}\right|>c_{1}W+\epsilon\right)\\
     \leq&2\exp\left(-c_{2}\min\left\{\frac{\epsilon^{2}}{V^{2}},\frac{\epsilon}{U}\right\}\right)\\
     \leq&2\exp\left(-c_{2}\min\left\{\frac{\epsilon^{2}}{B^{2}},\frac{\epsilon}{C}\right\}\right)
 \end{align*}
 then we finish the proof.
 \end{proof}
 \federated*
 \begin{proof}
 Denote $\mathcal{G}$ and $\mathcal{H}$ the sets of all possible loss gradients and the eigenvectors of $\mathbf{H}$ respectively. According to the algorithm, we can write the update in the sync step as:
 \begin{align*}
     \theta_{t+1}-\theta_{t}&=-\eta_{\text{global}}\texttt{desk}\left(\texttt{sk}\left(\bar{\Delta}_{t-1}\right)\right)\\
     &=-\eta_{\text{global}}\texttt{desk}\left(\frac{1}{C}\sum_{c=1}^{C}{\Delta}_{c,t}\right)\\
     &=-\eta_{\text{global}}\texttt{desk}\left(\frac{1}{C}\sum_{c=1}^{C}\texttt{sk}\left(\eta_{\text{local}}\sum_{k=1}^{K}\nabla\mathcal{L}_{c}(\theta_{c,t,k})\right)\right)\\
     &=-\eta_{\text{global}}R_{t}^{\top}\left(\frac{\eta_{\text{local}}}{C}\sum_{c=1}^{C}R_{t}\sum_{k=1}^{K}\nabla\mathcal{L}_{c}(\theta_{c,t,k})\right)\\
     &=-\eta\frac{1}{C}\sum_{c=1}^{C}R_{t}^{\top}R_{t}\sum_{k=1}^{K}\nabla\mathcal{L}_{c}(\theta_{c,t,k})
 \end{align*}
 in which $\eta = \eta_{\text{global}}\eta_{\text{local}}$. By Taylor expansion, we have
 \begin{align}
     \mathcal{L}(\theta_{t+1})=\mathcal{L}(\theta_{t})+\nabla\mathcal{L}(\theta_{t})^{\top}(\theta_{t+1}-\theta_{t})+\frac{1}{2}(\theta_{t+1}-\theta_{t})^{\top}\hat{H}_{\mathcal{L},t}(\theta_{t+1}-\theta_{t})\label{eq:distributed_original}
 \end{align}
 \textbf{Bounding $T_{1}$:}
 
 For each term in $T_{1}$, we have
 \begin{align}
     &\nabla\mathcal{L}(\theta_{t})^{\top}(\theta_{t+1}-\theta_{t})\notag\\
     =&-\nabla\mathcal{L}(\theta_{t})^{\top}\left(\frac{\eta}{C}\sum_{c=1}^{C}R_{t}^{\top}R_{t}\sum_{k=1}^{K}\nabla\mathcal{L}_{c}(\theta_{c,t,k})\right)\notag\\
     =&-\frac{\eta}{C}\nabla\mathcal{L}\left(\theta_{t}\right)^{\top}\sum_{c=1}^{C}R_{t}^{\top}R_{t}\sum_{k=1}^{K}\nabla\mathcal{L}_{c}(\theta_{c,t,k})\notag\\
     =&-\eta K\left\langle\nabla\mathcal{L}\left(\theta_{t}\right),\frac{1}{C}\sum_{c=1}^{C}\nabla\mathcal{L}_{c}\left(\theta_{t}\right)\right\rangle-\eta\left\langle\nabla\mathcal{L}\left(\theta_{t}\right),\frac{1}{C}\sum_{c=1}^{C}\sum_{k=1}^{K}\left(\nabla\mathcal{L}_{c}\left(\theta_{c,t,k}\right)-\nabla\mathcal{L}_{c}\left(\theta_{t}\right)\right)\right\rangle\notag\\
     &
     -\eta\left\langle\nabla\mathcal{L}\left(\theta_{t}\right),\frac{1}{C}\sum_{c=1}^{C}\sum_{k=1}^{K}\left(R_{t}^{\top}R_{t}\nabla\mathcal{L}_{c}\left(\theta_{c,t,k}\right)-\nabla\mathcal{L}_{c}\left(\theta_{c,t,k}\right)\right)\right\rangle\label{eq:T_{1}}
 \end{align}
 For the first term, according to the definition,
 \begin{align*}
     \left\langle\nabla\mathcal{L}(\theta_{t}),\frac{1}{C}\sum_{c=1}^{C}\nabla\mathcal{L}_{c}(\theta_{t})\right\rangle=\left\|\nabla\mathcal{L}(\theta_{t})\right\|_{2}^{2}
 \end{align*}
 so
 \begin{align}
     \sum_{t=0}^{T-1}\left\langle\nabla\mathcal{L}(\theta_{t}),\frac{1}{C}\sum_{c=1}^{C}\nabla\mathcal{L}_{c}(\theta_{t})\right\rangle=\sum_{t=0}^{T-1}\left\|\nabla\mathcal{L}(\theta_{t})\right\|_{2}^{2}\label{ineq:T_1_1}
 \end{align}
 For the second term, we have
 \begin{align}
     &\left\langle\nabla\mathcal{L}(\theta_{t}),\frac{1}{C}\sum_{c=1}^{C}\sum_{k=1}^{K}\left(\nabla\mathcal{L}_{c}(\theta_{c,t,k})-\nabla\mathcal{L}_{c}(\theta_{t})\right)\right\rangle\notag\\
     =&\frac{1}{C}\sum_{c=1}^{C}\sum_{k=1}^{K}\left\langle\nabla \mathcal{L}(\theta_t),\hat{H}_{\mathcal{L}}^{c,t,k}\left(\theta_{c,t,k}-\theta_{t}\right)\right\rangle\notag\\
     =&\frac{\eta_{\text{local}}}{C}\sum_{c=1}^{C}\sum_{k=1}^{K}\left\langle\nabla\mathcal{L}(\theta_t),\hat{H}_{\mathcal{L}}^{c,t,k}\sum_{\kappa=1}^{k}g_{c,t,\kappa}\right\rangle\notag\\
     \geq&-\frac{\eta_{\text{local}}}{N}\sum_{c=1}^{C}\sum_{k=1}^{K}\left\|\nabla \mathcal{L}(\theta_t)\right\|_{2}\cdot \Lambda_{\max}\sum_{\kappa=1}^{k}\left\|g_{c,t,\kappa}\right\|_{2}\notag\\
     =&-\frac{\eta_{\text{local}}}{N}\cdot N\cdot G^{2}\cdot\Lambda_{\max}\sum_{k=1}^{K}k\notag\\
     \geq&-\frac{\eta_{\text{local}}\Lambda_{\max}K^{2}G^{2}}{2};\label{ineq:T_1_2}
 \end{align}
 For the third term, according to Theorem~\ref{thm:sketching_guarantee}, we have that with probability at least $1-\frac{2\delta}{T}$,
 \begin{align}
     &\left|\left\langle\nabla\mathcal{L}\left(\theta_{t}\right),\frac{1}{C}\sum_{c=1}^{C}\sum_{k=1}^{K}\left(R_{t}^{\top}R_{t}\nabla\mathcal{L}_{c}(\theta_{c,t,k})-\nabla\mathcal{L}_{c}(\theta_{c,t,k})\right)\right\rangle\right|\notag\\
     =&\frac{1}{C}\left|\sum_{c=1}^{C}\sum_{k=1}^{K}\left(\nabla\mathcal{L}(\theta_{t})^{\top}R_{t}^{\top}R_{t}\nabla\mathcal{L}_{c}(\theta_{c,t,k})-\nabla\mathcal{L}(\theta_{t})^{\top}\nabla\mathcal{L}_{c}(\theta_{c,t,k})\right)\right|\notag\\
     \leq&K\sup_{g\in\mathcal{G},h\in\mathcal{G}}\left|g^{\top}R_{t}^{\top}R_{t}h-g^{\top}h\right|\notag\\
     \leq&KZ\left(\mathcal{G},\mathcal{G},\frac{\delta}{T}\right)\notag\\
     =&K\left(\frac{w^{2}(\mathcal{G})}{b}+\frac{Gw(\mathcal{G})}{\sqrt{b}}+\frac{Gw(\mathcal{G})\sqrt{\log(T/\delta)}}{b}+\frac{G^{2}\sqrt{\log(T/\delta)}}{\sqrt{b}}+\frac{G^{2}\log(T/\delta)}{b}\right)\notag\\
     =&\frac{Kw^{2}(\mathcal{G})}{b}+\frac{KG}{\sqrt{b}}\left(w(\mathcal{G})+G\sqrt{\log(T/\delta)}\right)\left(1+\frac{\sqrt{\log(T/\delta)}}{\sqrt{b}}\right)\label{ineq:T_1_3}
 \end{align}
 Substituting \eqref{ineq:T_1_1}-\eqref{ineq:T_1_3} into \eqref{eq:T_{1}}, we have that with probability at least $1-\frac{2\delta}{T}$,
 \begin{align}
     &\nabla\mathcal{L}(\theta_{t})^{\top}(\theta_{t+1}-\theta_{t})\notag\\=&-\eta K\left\langle\nabla\mathcal{L}\left(\theta_{t}\right),\frac{1}{C}\sum_{c=1}^{C}\nabla\mathcal{L}_{c}\left(\theta_{t}\right)\right\rangle-\eta\left\langle\nabla\mathcal{L}\left(\theta_{t}\right),\frac{1}{C}\sum_{c=1}^{C}\sum_{k=1}^{K}\left(\nabla\mathcal{L}_{c}\left(\theta_{c,t,k}\right)-\nabla\mathcal{L}_{c}\left(\theta_{t}\right)\right)\right\rangle\notag\\
     &-\eta\left\langle\nabla\mathcal{L}\left(\theta_{t}\right),\frac{1}{C}\sum_{c=1}^{C}\sum_{k=1}^{K}\left(R_{t}^{\top}R_{t}\nabla\mathcal{L}_{c}(\theta_{c,t,k})-\nabla\mathcal{L}_{c}(\theta_{c,t,k})\right)\right\rangle\notag\\
     \leq&-\eta K\left\|\nabla\mathcal{L}(\theta_{t})\right\|_{2}^{2}+\frac{\eta\eta_{\text{local}}\Lambda_{\max}K^{2}G^{2}}{2}+\eta KZ\left(\mathcal{G},\mathcal{G},\frac{\delta}{T}\right)\label{ineq:T_1_5}
 \end{align}
 in which
 \begin{align}
     Z\left(\mathcal{G},\mathcal{G},\frac{\delta}{T}\right)=\frac{w^{2}(\mathcal{G})}{b}+\frac{G}{\sqrt{b}}\left(w(\mathcal{G})+G\sqrt{\log(T/\delta)}\right)\left(1+\frac{\sqrt{\log(T/\delta)}}{\sqrt{b}}\right)\label{eq:f_g_g}
 \end{align}
 \textbf{Bounding $T_{2}$:}
 
 According to Assumption~\ref{assump:hessian-spectrum-app}, we can get that
 \begin{align}
     \left|\left(\theta_{t+1}-\theta_{t}\right)^{\top}\hat{H}_{\mathcal{L},t}(\theta_{t+1}-\theta_{t})\right|&\leq\left(\theta_{t+1}-\theta_{t}\right)^{\top}\mathbf{H}(\theta_{t+1}-\theta_{t})\notag\\&=\eta_{\text{global}}^{2}\left(\texttt{desk}\left(\bar{\tilde{\Delta}}_{t}\right)\right)^{\top}\left(\sum_{i=1}^{p}\Lambda_{i}v_{i}v_{i}^{\top}\right)\left(\texttt{desk}\left(\bar{\Delta}_{t}\right)\right)\notag\\
     &=\eta_{\text{global}}^{2}\sum_{i=1}^{p}\Lambda_{i}\left|\left(\texttt{desk}\left(\bar{\Delta}_{t}\right)\right)^{\top}v_{i}\right|^{2}\label{eq:T_2_1}
 \end{align}
 For each $i\in[d]$, we have
 \begin{align*}
     \left|\left(\texttt{desk}(\bar{\Delta}_{t})\right)^{\top}v_{i}\right|&=\left|\left\langle\texttt{desk}(\bar{\tilde{\Delta}}_{t}),v_{i}\right\rangle\right|\notag\\
     &=\frac{\eta_{\text{local}}}{C}\left|\left\langle \texttt{desk}\left(\sum_{c=1}^{C}\texttt{sk}\left(\Delta_{c,t}\right)\right),v_{i}\right\rangle\right|\notag\\
     &=\frac{\eta_{\text{local}}}{C}\left|\left\langle \sum_{c=1}^{C}R_{t}^{\top}R_{t}\Delta_{c,t},v_{i}\right\rangle\right|\notag\\
     &=\frac{\eta_{\text{local}}}{C}\left|\left\langle \sum_{c=1}^{C}\sum_{k=1}^{K}R_{t}^{\top}R_{t}\nabla\mathcal{L}_{c}(\theta_{c,t,k}),v_{i}\right\rangle\right|
 \end{align*}
 According to Theorem~\ref{thm:sketching_guarantee}, we have that with probability at least $1-\frac{2\delta}{dT}$,
 \begin{align}
     &\left|\left(\texttt{desk}(\bar{\Delta}_{t})\right)^{\top}v_{i}\right|\notag\\
     =&\left|\frac{\eta_{\text{local}}}{C}\left\langle \sum_{c=1}^{C}\sum_{k=1}^{K}R_{t}^{\top}R_{t}\nabla\mathcal{L}_{c}(\theta_{c,t,k}),v_{i}\right\rangle\right|\notag\\
     =&\left|\frac{\eta_{\text{local}}}{C} \sum_{c=1}^{C}\sum_{k=1}^{K}\left(v_{i}^{\top}R_{t}^{\top}R_{t}\nabla\mathcal{L}_{c}(\theta_{c,t,k})-v_{i}^{\top}\nabla\mathcal{L}_{c}(\theta_{c,t,k})\right)+\frac{\eta_{\text{local}}}{C} \sum_{c=1}^{C}\sum_{k=1}^{K}v_{i}^{\top}\nabla\mathcal{L}_{c}(\theta_{c,t,k})\right|\notag\\
     \leq&\left|\frac{\eta_{\text{local}}}{C} \sum_{c=1}^{C}\sum_{k=1}^{K}\left(v_{i}^{\top}R_{t}^{\top}R_{t}\nabla\mathcal{L}_{c}(\theta_{c,t,k})-v_{i}^{\top}\nabla\mathcal{L}_{c}(\theta_{c,t,k})\right)\right|+\left|\frac{\eta_{\text{local}}}{C} \sum_{c=1}^{C}\sum_{k=1}^{K}v_{i}^{\top}\nabla\mathcal{L}_{c}(\theta_{c,t,k})\right|\notag\\
     \leq&\left|\frac{\eta_{\text{local}}}{C} \sum_{c=1}^{C}\sum_{k=1}^{K}\left(v_{i}^{\top}R_{t}^{\top}R_{t}\nabla\mathcal{L}_{c}(\theta_{c,t,k})-v_{i}^{\top}\nabla\mathcal{L}_{c}(\theta_{c,t,k})\right)\right|+\left|\frac{\eta_{\text{local}}}{C} \sum_{c=1}^{C}\sum_{k=1}^{K}v_{i}^{\top}\nabla\mathcal{L}_{c}(\theta_{c,t,k})\right|\notag\\
    \leq&\eta_{\text{local}}K\sup_{g\in\mathcal{G},h\in\mathcal{H}}\left|g^{\top}R_{t}^{\top}R_{t}h-g^{\top}h\right|+\left|\frac{\eta_{\text{local}}}{C} \sum_{c=1}^{C}\sum_{k=1}^{K}v_{i}^{\top}\nabla\mathcal{L}_{c}(\theta_{c,t,k})\right|\notag\\
     \leq&\eta_{\text{local}}KZ\left(\mathcal{G},\mathcal{H},\frac{\delta}{T}\right)+\left|\frac{\eta_{\text{local}}}{C} \sum_{c=1}^{C}\sum_{k=1}^{K}v_{i}^{\top}\nabla\mathcal{L}_{c}(\theta_{c,t,k})\right|\notag\\
     =&\frac{\eta_{\text{local}}Kw(\mathcal{G})w(\mathcal{H})}{b}+\frac{\eta_{\text{local}}Kw(\mathcal{G})+\eta_{\text{local}}KGw(\mathcal{H})}{\sqrt{b}}+\frac{\eta_{\text{local}}K\sqrt{\log(T/\delta)}\left(w(\mathcal{G})+Gw(\mathcal{H})\right)}{b}\notag\\
     &+\frac{\eta_{\text{local}}KG\sqrt{\log(T/\delta)}}{\sqrt{b}}+\frac{\eta_{\text{local}}KG\log(T/\delta)}{b}+\left|\frac{\eta_{\text{local}}}{C} \sum_{c=1}^{C}\sum_{k=1}^{K}v_{i}^{\top}\nabla\mathcal{L}_{c}(\theta_{c,t,k})\right|\label{ineq:T_2_1}
 \end{align}
 Substituting \eqref{ineq:T_2_1} into \eqref{eq:T_2_1}, we have that with probability at least $1-\frac{2\delta}{T}$,
 \begin{align}
     \left(\theta_{t+1}-\theta_{t}\right)^{\top}&\hat{H}_{\mathcal{L},t}(\theta_{t+1}-\theta_{t})\leq\eta_{\text{global}}^{2}\sum_{i=1}^{p}\Lambda_{i}\left|\left(\texttt{desk}\left(\bar{\Delta}_{t}\right)\right)^{\top}v_{i}\right|^{2}\notag\\
     &=\eta_{\text{global}}^{2}\sum_{i=1}^{p}\Lambda_{i}\left(\eta_{\text{local}}KZ\left(\mathcal{G},\mathcal{H},\frac{\delta}{T}\right)+\left|\frac{\eta_{\text{local}}}{C} \sum_{c=1}^{C}\sum_{k=1}^{K}v_{i}^{\top}\nabla\mathcal{L}_{c}(\theta_{c,t,k})\right|\right)^{2}\notag\\
     &\leq2\eta^{2}K^{2}Z^{2}\left(\mathcal{G},\mathcal{H},\frac{\delta}{T}\right)\sum_{i=1}^{p}\left|\Lambda_{i}\right|+2\eta^{2}\sum_{i=1}^{p}\Lambda_{i}\left(\frac{1}{C} \sum_{c=1}^{C}\sum_{k=1}^{K}v_{i}^{\top}\nabla\mathcal{L}_{c}(\theta_{c,t,k})\right)^{2}\notag\\
     &\leq2\eta^{2}K^{2}\kappa\Lambda_{\max}Z^{2}\left(\mathcal{G},\mathcal{H},\frac{\delta}{T}\right)+2\Lambda_{\max}\eta^{2}K^{2}G^{2}\label{ineq:T_2_2}
 \end{align}
 in which 
 \begin{align}
     Z\left(\mathcal{G},\mathcal{H},\frac{\delta}{T}\right)&=\frac{w(\mathcal{G})w(\mathcal{H})}{b}+\frac{w(\mathcal{G})+Gw(\mathcal{H})}{\sqrt{b}}+\frac{\sqrt{\log(T/\delta)}\left(w(\mathcal{G})+Gw(\mathcal{H})\right)}{b}\notag\\&+\frac{G\sqrt{\log(T/\delta)}}{\sqrt{b}}+\frac{G\log(T/\delta)}{b}\label{eq:f_g_h}
 \end{align}
 Substituting \eqref{ineq:T_1_5} and \eqref{ineq:T_2_2} into \eqref{eq:distributed_original}, and according to Assumption~\ref{assump:pl}, we have that with probability at least $1-\frac{4\delta}{T}$,
 \begin{align*}
     &\mathcal{L}(\theta_{t+1})-\mathcal{L}(\theta_{t})\\
     =&\nabla\mathcal{L}(\theta_{t})^{\top}(\theta_{t+1}-\theta_{t})+\frac{1}{2}(\theta_{t+1}-\theta_{t})^{\top}\hat{H}_{\mathcal{L},t}(\theta_{t+1}-\theta_{t})\\
     \leq&-\eta K\left\|\nabla\mathcal{L}(\theta_{t})\right\|_{2}^{2}+\frac{\eta\eta_{\text{local}}\Lambda_{\max}K^{2}G^{2}}{2}+\eta KZ\left(\mathcal{G},\mathcal{G},\frac{\delta}{T}\right)\\
     &+2\eta^{2}K^{2}\kappa\Lambda_{\max}\left(Z^{2}\left(\mathcal{G},\mathcal{H},\frac{\delta}{T}\right)+G^{2}\right)\\
     \leq&-\eta K\cdot2\mu\left(\nabla\mathcal{L}(\theta_{t})-\mathcal{L}^{*}\right)+\frac{\eta\eta_{\text{local}}\Lambda_{\max}K^{2}G^{2}}{2}+\eta KZ\left(\mathcal{G},\mathcal{G},\frac{\delta}{T}\right)\\
     &+2\eta^{2}K^{2}\kappa\Lambda_{\max}\left(Z^{2}\left(\mathcal{G},\mathcal{H},\frac{\delta}{T}\right)+G^{2}\right)
 \end{align*}
 According to \cite[Theorem 4]{banerjee2024loss}, $w\left(\mathcal{G}\right)\leq O(1)$ when $L=\Omega(\log m)$. Since $\mathcal{H}$ is the finite set of $d$ eigenvectors of $\mathbf{H}$, $w(\mathcal{H})\leq\log p$~\cite{vers18}. Substituting these adn $b=\frac{\text{polylog}\left(\frac{TNp^{2}}{\delta}\right)}{\varepsilon^{2}}$ into \eqref{eq:f_g_g} and \eqref{eq:f_g_h}, we have
 \begin{align*}
     Z\left(\mathcal{G},\mathcal{G},\frac{\delta}{T}\right)&=\frac{w^{2}(\mathcal{G})}{b}+\frac{G}{\sqrt{b}}\left(w(\mathcal{G})+G\sqrt{\log(T/\delta)}\right)\left(1+\frac{\sqrt{\log(T/\delta)}}{\sqrt{b}}\right)\\
     &\lesssim\frac{1}{b}+\frac{G}{\sqrt{b}}\left(1+G\sqrt{\log(T/\delta)}\right)\left(1+\frac{\sqrt{\log(T/\delta)}}{\sqrt{b}}\right)\\
     &\lesssim G^{2}+\varepsilon^{2};\\
     Z\left(\mathcal{G},\mathcal{H},\frac{\delta}{T}\right)&=\frac{w(\mathcal{G})w(\mathcal{H})}{b}+\frac{w(\mathcal{G})+Gw(\mathcal{H})}{\sqrt{b}}+\frac{\sqrt{\log(T/\delta)}\left(w(\mathcal{G})+Gw(\mathcal{H})\right)}{b}\notag\\&+\frac{G\sqrt{\log(T/\delta)}}{\sqrt{b}}+\frac{G\log(T/\delta)}{b}\\
     &\lesssim\frac{\log d}{b}+\frac{1+G\log d}{\sqrt{b}}+\frac{\sqrt{\log(T/\delta)}\left(1+G\log d\right)}{b}+\frac{G\sqrt{\log(T/\delta)}}{\sqrt{b}}+\frac{G\log(T/\delta)}{b}\\
     &\lesssim \varepsilon\left(1+G\right).
 \end{align*}
 In addition, according to~\cite{banerjee2023restricted}, $\Lambda_{\max}\leq\frac{c_{H}c_{l}}{\sqrt{m}}+c_{s}\varrho^{2}$. Combining these with $\eta=\eta_{\text{local}}$, we can get that
 \begin{align*}
     &\mathcal{L}(\theta_{t+1})-\mathcal{L}^{*}\\
     \leq&\left(1-2\mu\eta K\right)\cdot\left(\nabla\mathcal{L}(\theta_{t})-\mathcal{L}^{*}\right)+\frac{\eta\eta_{\text{local}}\Lambda_{\max}K^{2}G^{2}}{2}+\eta KZ\left(\mathcal{G},\mathcal{G},\frac{\delta}{T}\right)+2\eta^{2}K^{2}\kappa\Lambda_{\max}Z^{2}\left(\mathcal{G},\mathcal{H},\frac{\delta}{T}\right)+2\eta^{2}\\
     \leq&\left(1-2\mu\eta K\right)\cdot\left(\nabla\mathcal{L}(\theta_{t})-\mathcal{L}^{*}\right)+\frac{\eta^{2}K^{2}G^{2}}{2}\left(\frac{c_{H}c_{l}}{\sqrt{m}}+c_{s}\varrho^{2}\right)+c_{1}\eta K(G^{2}+\varepsilon^{2})\\&+2\eta^{2}K^{2}\kappa\cdot\left(\frac{c_{H}c_{l}}{\sqrt{m}}+c_{s}\varrho^{2}\right)\cdot c_{2}\varepsilon^{2}\left(G^{2}+1\right)+2\eta^{2}K^{2}G^{2}\cdot\left(\frac{c_{H}c_{l}}{\sqrt{m}}+c_{s}\varrho^{2}\right)\\
     =&\left(1-2\mu\eta K\right)\cdot\left(\nabla\mathcal{L}(\theta_{t})-\mathcal{L}^{*}\right)+\frac{5\eta^{2}K^{2}G^{2}}{2}\left(\frac{c_{H}c_{l}}{\sqrt{m}}+c_{s}\varrho^{2}\right)+c_{1}\eta K(G^{2}+\varepsilon^{2})\\
     &+2\eta^{2}K^{2}\kappa\cdot\left(\frac{c_{H}c_{l}}{\sqrt{m}}+c_{s}\varrho^{2}\right)\cdot c_{2}\varepsilon^{2}\left(G^{2}+1\right)
 \end{align*}
 Iterating from 0 to $T-1$, we can get that
 \begin{align*}
     &\mathcal{L}(\theta_{T})-\mathcal{L}^{*}\\
     \leq&\left(1-2\mu\eta K\right)^{T}\left(\mathcal{L}(\theta_{0})-\mathcal{L}^{*}\right)+\frac{1}{2\mu\eta K}\Big(\frac{5\eta^{2}K^{2}G^{2}}{2}\left(\frac{c_{H}c_{l}}{\sqrt{m}}+c_{s}\varrho^{2}\right)+c_{1}\eta K(G^{2}+\varepsilon^{2})\\&+2\eta^{2}K^{2}\kappa\cdot\left(\frac{c_{H}c_{l}}{\sqrt{m}}+c_{s}\varrho^{2}\right)\cdot c_{2}\varepsilon^{2}\left(G^{2}+1\right)\Big)\\
     \leq&\left(\mathcal{L}(\theta_{0})-\mathcal{L}^{*}\right)e^{-2\mu\eta KT}+\frac{5\eta KG^{2}}{4\mu}\left(\frac{c_{H}c_{l}}{\sqrt{m}}+c_{s}\varrho^{2}\right)+\frac{c_{1}\left(G^{2}+\varepsilon^{2}\right)}{2\mu}+\frac{\eta K\kappa\varepsilon^{2}\left(G^{2}+1\right)}{\mu}\left(\frac{c_{H}c_{l}}{\sqrt{m}}+c_{s}\varrho^{2}\right)
 \end{align*}
 then we finish the proof.
 \end{proof}

\section{Linear Regression with Sketching}
\label{sec:regression_app}
We observe $n$ i.i.d. samples $(\x_i, y_i)$ from the linear model $y_i = \x_i^\top \beta^* + \varepsilon_i$, where $\x_i \in \mathbb{R}^d$ are sub-Gaussian covariates with covariance $\Sigma$ and $\varepsilon_i$ are independent sub-Gaussian noise terms. Further $\beta \in \cB \subseteq \R^d$. 
We solve a least squares problem to compute $\hat{\beta}^s$ using the sketched inputs and corresponding responses, i.e., using $(\mathbf{S}\x_i,y_i)$. Subsequently we de-sketch the estimate using $\mathbf{S}^\top \hat{\beta}^s$ and use this to make predictions. Note that solving the least squares problem in a lower dimensional sketched space is computationally faster. Our goal is to bound the error $(\mathbf{S}^\top \hat{\beta}^s - \beta^*)^\top \x$, for $\x \in \cX$. 

We assume that the sketching matrix $\mathbf{S}$ has all it's entries drawn i.i.d.\ from $N(0,1/b)$. Now consider $\x \in \cX$ and assume $\|\x\| = 1$. We have

\begin{align*}
    (\mathbf{S}^\top \hat{\beta}^s - \beta^*)^\top \x &= \langle \mathbf{S}^\top \hat{\beta}^s, \x\rangle - \langle \mathbf{S}\beta^*, \mathbf{S}\x \rangle + \langle \mathbf{S}\beta^*, \mathbf{S}\x \rangle - \langle \beta^*, \x\rangle \\
    &= \langle \hat{\beta}^s - \mathbf{S}\beta^*, \mathbf{S}\x \rangle + {\beta^*}^{\top} (\mathbf{S}^\top\mathbf{S} - I) \x 
\end{align*}

The ordinary least–squares (OLS) estimator is
\[
\hat{\beta} \;=\; ({X^s}^{\top}X^s)^{-1}{X^s}^{\top},
\]
where the $i$-th row of $X^s$ is $\mathbf{S}\x_i$. Further suppose $\varepsilon$ be an $n$ dimensional vector whose $i$-th row is $\varepsilon_i$. Then
\[
\hat{\beta}^s - \mathbf{S}\beta^{*} \;=\; ({X^s}^{\top}{X^s})^{-1}{X^s}^{\top}\varepsilon.
\]

Following a standard analysis, we have with probability $1-\delta$
\begin{align*}
    \|\mathbf{S}^\top \hat{\beta}^s - \beta^*\|_{2}
\;\le\;
\frac{\sigma}{\sqrt{\lambda_{\min}({X^s}^{\top}{X^s})}}
\Bigl(\sqrt{b}+\sqrt{2\log(1/\delta)}\Bigr).
\end{align*}
where $\sigma^2$ is the variance of the noise process $\varepsilon_i$. Therefore 
\begin{align*}
    \sup_{\x \in \cX}(\mathbf{S}^\top \hat{\beta}^s - \beta^*)^\top \x 
    &= \sup_{\x \in \cX} \langle \hat{\beta}^s - \mathbf{S}\beta^*, \mathbf{S}\x \rangle + {\beta^*}^{\top} (\mathbf{S}^\top\mathbf{S} - I) \x \\
    &\leq  \|\mathbf{S}^\top \hat{\beta}^s - \beta^*\|_2 \underbrace{\sup_{\x \in \cX} \|\mathbf{S}\x\|_2}_{I} + \underbrace{\sup_{\x \in \cX} {\beta^*}^{\top} (\mathbf{S}^\top\mathbf{S} - I) \x}_{II}
\end{align*}
Term $I$ can be handled using the one set result from \cite{krahmer2014suprema}. Using a construction as in Section~\ref{subsec:JL} and thereafter invoking Theorem~3.1 from \cite{krahmer2014suprema} we have

\begin{align*}
        &P\Bigg\{\sup_{u \in \cU}|\|\mathbf{S}\x\|_2^2 - \E \|\mathbf{S}\x\|_2^2| \gtrsim \left[ \frac{1}{{b}} \omega^2(\cX) + \frac{1}{\sqrt{b}}\omega(\cX) \right] + \epsilon \Bigg\}\\
        & \lesssim \exp \left(  - \min \left\{ \frac{\epsilon^2}{\big({  \frac{1}{b} \omega(\cX) + \frac{1}{\sqrt{b}} }\big)^2}, \frac{\epsilon }{{\frac{1}{b}}} \right\} \right)
\end{align*}
Next using Theorem~\ref{thm:cross_product} we have
\begin{align*}
        P\Bigg\{\sup_{\beta^* \in \cB, \x \in \cX}| {\beta^*}^{\top} (\mathbf{S}^\top\mathbf{S} - I) \x| &\geq {\color{Red} \frac{1}{{b}} \omega(\cB)\omega(\cX) + \frac{1}{\sqrt{b}}(\omega(\cB) + \omega(\cX))} + \epsilon \Bigg\}\\
        & \lesssim \exp \left(  - \min \left\{ \frac{\epsilon^2}{\big({\color{Blue} \frac{1}{b} (\omega(\cB) + \omega(\cX)) +  \frac{1}{\sqrt{b}}}\big)^2}, \frac{\epsilon}{{\color{Green}\frac{1}{b}}} \right\} \right)
    \end{align*}

Choose $\epsilon^2 = ({ \frac{1}{b} (\omega(\cB) + \omega(\cX)) +  \frac{1}{\sqrt{b}}}\big)^2 \Bar{\epsilon}^2$, and take a union bound over both the events to get with probability $1 - \exp(- \Bar{\epsilon}^2)$
\begin{align*}
    \sup_{u \in \cU}|\|\mathbf{S}\x\|_2^2 - \E \|\mathbf{S}\x\|_2^2| &\lesssim \left[ \frac{1}{{b}} \omega^2(\cX) + \frac{1}{\sqrt{b}}\omega(\cX) \right] + ({ \frac{1}{b} (\omega(\cB) + \omega(\cX)) +  \frac{1}{\sqrt{b}}})\Bar{\epsilon}\\
    \sup_{\beta^* \in \cB, \x \in \cX}| {\beta^*}^{\top} (\mathbf{S}^\top\mathbf{S} - I) \x| &\lesssim \frac{1}{{b}} \omega(\cB)\omega(\cX) + \frac{1}{\sqrt{b}}(\omega(\cB) + \omega(\cX)) + ({ \frac{1}{b} (\omega(\cB) + \omega(\cX)) +  \frac{1}{\sqrt{b}}})\Bar{\epsilon}
\end{align*}

Taking a union bound over all the high probability events we have that with probability $1 - \delta$ for all $\x \in \cX$

\begin{align*}
    (\mathbf{S}^\top \hat{\beta}^s - \beta^*)^\top \x &\lesssim \frac{\sigma}{\sqrt{\lambda_{\min}({X^s}^{\top}{X^s})}}
\Bigl(\sqrt{b}+\sqrt{2\log(1/\delta)}\Bigr).\left[ \frac{1}{{b}} \omega^2(\cX) + \frac{1}{\sqrt{b}}\omega(\cX) \right] \\ & + \frac{1}{{b}} \omega(\cB)\omega(\cX) + \frac{1}{\sqrt{b}}(\omega(\cB) + \omega(\cX)) + ({ \frac{1}{b} (\omega(\cB) + \omega(\cX))  +  \frac{1}{\sqrt{b}}})\sqrt{\log(\delta^{-1})}
\end{align*}

The above bound depends on the Gaussian widths $\omega(\cX)$ and $\omega(\cB)$, instead of depending on the ambient dimension $d$. In cases where $\omega(\cX)$ and $\omega(\cB)$ are small, one can choose the sketching dimension $b$ to balance the terms and obtain a tighter bound.

\section{Extension to Sum of Random Quadratic Forms}
\label{sec:app_proof_sumTtheorem}

\subsection{Expected Analysis}
\label{sec:sum_version_expected}
In this section we provide an expected deviation bound on the random quadratic form. For simplicity we consider the single set version. 

\begin{theorem}
    {Let $\mathcal{A} \subset \mathbb{R}^{m \times n}$ be a set of matrices and let $\epsilon$ be a Rademacher vector of length $n$. Then}
        \[
        \mathbb{E} \sup_{A \in \mathcal{A}} \left\| A \bxi_t \right\|_2^2 - \mathbb{E} \left\| A \bxi_t \right\|_2^2 \leq C_1 \left( \gamma_2(\mathcal{A}, \| \cdot \|_{2 \to 2})^2 + d_F(\mathcal{A}) \gamma_2(\mathcal{A}, \| \cdot \|_{2 \to 2}) + d_F(\mathcal{A}) d_{2 \to 2}(\mathcal{A}) \right).
        \]
\end{theorem}

\bigskip

\noindent\textbf{Proof.} By the decoupling inequality of Theorem 2.4 \citep{krahmer2014suprema}, we have
\begin{align}
\mathbb{E}\; \mathcal{C}_{\mathcal{A}}(\bxi^{1:T}) 
&= \mathbb{E} \sup_{A \in \mathcal{A}} \sum_{t=1}^{T} \left\| A \bxi_t \right\|_2^2 - \mathbb{E}  \left\| A \bxi_t \right\|_2^2 
= \mathbb{E} \sup_{A \in \mathcal{A}} \left| \sum_{t=1}^{T}\sum_{j \neq k} \bxi_{t,j} \bxi_{t,k} (A^\top A)_{j,k} \right| \notag \\
&\leq 4 \mathbb{E} \sup_{A \in \mathcal{A}} \left| \sum_{t=1}^{T}\sum_{j,k} \bxi_{t,j}' \bxi_{t,k} (A^\top A)_{j,k} \right| \notag\\
&\lesssim \gamma_2(\mathcal{A}, \| \cdot \|_{2 \to 2}) \mathbb{E} N_{\mathcal{A}}(\bxi^{1:T}) + \sup_{A \in \mathcal{A}} \mathbb{E} |\sum_{t=1}^{T} \langle A \bxi_t, A \bxi_t' \rangle |.
\label{eq:expected_eq1}
\end{align}
where $N_{\mathcal{A}}(\bxi^{1:T}) := \sup_{A \in \mathcal{A}} \| \sum_{t=1}^{T} A \bxi_t \|_2,$ and the inequality follows by Lemma~\ref{lemm:offd-lp-sum} by setting $M=N=A$.
Since
\begin{align*}
\mathbb{E} | \sum_{t=1}^{T}\langle A \bxi_t, A \bxi'_t \rangle | &\overset{(a)}{\leq} \mathbb{E} | \sum_{t=1}^{T} \varepsilon_t\langle A \bxi_t, A \bxi'_t \rangle | \overset{(b)}{\leq} \E_{\bxi_t,\bxi'_t} \sqrt{\sum_{t=1}^{T} |\langle A \bxi_t, A \bxi_t' \rangle |^2 }
\\& \overset{(c)}{\leq}  \sqrt{\sum_{t=1}^{T} \E_{\bxi_t,\bxi'_t}|\langle A \bxi_t, A \bxi_t' \rangle |^2 } = \sqrt{T}\| A^\top A \|_F \leq \sqrt{T}\| A \|_{2 \to 2} \| A \|_F,
\end{align*}
we have
\begin{equation}
\sup_{A \in \mathcal{A}} \mathbb{E} | \sum_{t=1}^{T} \langle A \bxi_t, A \bxi_t' \rangle | 
\leq \sqrt{T} d_{2 \to 2}(\mathcal{A}) d_F(\mathcal{A}) \leq \sqrt{T} d_F(\mathcal{A})^2.
\end{equation}
Here $(a)$ follows by symmetrization \citep{leta91}, $(b)$ follows by Kinchinte \citep{vers18} and $(c)$ follows by Jensen's inequality.
We conclude that
\begin{align*}
(\mathbb{E} N_{\mathcal{A}}(\bxi^{1:T}))^2 &\leq \mathbb{E} N_{\mathcal{A}}(\bxi^{1:T})^2 
\leq \mathbb{E}\; \mathcal{C}_{\mathcal{A}}(\bxi^{1:T}) + \sqrt{T} d_F(\mathcal{A}) \\
&\lesssim \gamma_2(\mathcal{A}, \| \cdot \|_{2 \to 2}) \mathbb{E}\; N_{\mathcal{A}}(\bxi^{1:T}) + \sqrt{T} d_F(\mathcal{A})^2,
\end{align*}
so that
\[
\mathbb{E} N_{\mathcal{A}}(\bxi^{1:T}) \lesssim \gamma_2(\mathcal{A}, \| \cdot \|_{2 \to 2}) + \sqrt{T} d_F(\mathcal{A}).
\]

Plugging this into \eqref{eq:expected_eq1} yields the claim.
\subsection{High Probability Analysis}
\label{sec:sum_version_hp}

Our objective is to develop large deviation bound for the following random variable.
\begin{equation}
    C_{\cM,\cN}(\bxi^{1:T}) ~ \;\; \triangleq ~\sup_{M \in \cM, N \in \cN} \bigg| \sum_{t=1}^{T} (\bxi_t^\top M^\top N \bxi_t) - \E(\bxi_t^\top M^\top N \bxi_t) \bigg|~ 
    \label{eq:caxi_T} 
\end{equation}

To develop large deviation bounds on $C_{\cM,\cN}(\bxi^{1:T})$, as in the proof of Theorem~\ref{thm:cross_product} we decompose the quadratic form into terms depending on the off-diagonal and the diagonal elements of $M^\top N$ respectively as follows.
    \begin{align*}
        B_{\cM, \cN}(\bxi^{1:T}) ~ & \triangleq ~\sup_{M\in\cM, N\in\cN} \left| \sum_{t=1}^{T} \sum_{\substack{j,k=1\\j \neq k}}^n \bxi_{t,j} \bxi_{t,k} \langle M_j, N_k \rangle \right| ~,
    \\
        D_{\cM, \cN}(\bxi^{1:T}) ~ & \triangleq ~\sup_{M\in\cM, N\in\cN} \left| \sum_{t=1}^{T} \sum_{j=1}^n (|\bxi_{t,j}|^2 - \E|\bxi_{t,j}|^2) \langle M_j, N_j \rangle \right| ~,
    \end{align*}
    where $M_i$ and $N_i$ are the $i$-th row of $M$ and $N$ respectively.

Our large deviation bound for $C_{\cM,\cN}(\bxi^{1:T})$ is based on bounding $\| C_{\cM, \cN}(\bxi^{1:T}) \|_{L_p}$ via $\| B_{\cM, \cN}(\bxi^{1:T}) \|_{L_p}$ and $\| D_{\cM, \cN}(\bxi) \|_{L_p}$, yielding
\begin{equation}
    \| C_{\cM, \cN}(\bxi^{1:T})\|_{L_p} \leq {\color{Red}W \cdot \sqrt{T}} + \sqrt{p} \cdot {\color{Blue}V \cdot \sqrt{T}} + p \cdot {\color{Green}U \cdot \sqrt{T}}, \quad \forall p \geq 1~.
\end{equation}
Using standard moment bounds and Markov’s inequality \citep{will91,vers12}, this gives for all $\epsilon > 0$:
\begin{align}
    \mathbb{P}\left(|C_{\cM, \cN}(\bxi^{1:T})| \geq {\color{Red}W \cdot \sqrt{T}} + \epsilon \right) &\leq \exp\left\{-\min\left(\frac{\epsilon^2}{4\;{\color{Blue}V \cdot T}^2}, \frac{\epsilon}{2\;{\color{Green}U \cdot \sqrt{T}}}\right)\right\}. \label{eq:sum_T_mark2}
\end{align}
Following the proof of Lemma~\ref{lemm:Decomposition} we can show that:
\begin{align}
    \| C_{\cM, \cN}(\bxi^{1:T}) \|_{L_p} & \leq \| B_{\cM, \cN}(\bxi^{1:T})\|_{L_p}  +  \| D_{\cM, \cN}(\bxi^{1:T}) \|_{L_p}~.
\end{align}
Next we bound $\| B_{\cM, \cN}(\bxi^{1:T})\|_{L_p}$ and $\| D_{\cM, \cN}(\bxi^{1:T}) \|_{L_p}$ using Theorem~\ref{theo:diagonalLpBound-sumT} and Theorem~\ref{theo:diagonalLpBound} respectively. Note that collecting terms corresponding to {\color{Red}W}, {\color{Blue}V} and {\color{Green}U} and combining them with the observation in \eqref{eq:sum_T_mark2} completes the proof of Theorem~\ref{theo:sumTtheorem}. Next we bound $\|B_{\cM, \cN}(\bxi^{1:T})\|_{L_p}$ and $\|D_{\cM, \cN}(\bxi^{1:T})\|_{L_p}$.

We first control the off-diagonal term in the following theorem.

\begin{theorem}
    \label{theo:diagonalLpBound-sumT}
    Let $\bxi_t$ for all $t \in [T]$ be stochastic processes satisfying Assumption~\ref{asmp:xi}. Then, for all $p \geq 1$, we have
    \begin{align*}
        &\| B_{\cM, \cN}(\bxi^{1:T})\|_{L_p} \lesssim \sqrt{T}\Bigg[{\color{Red}\gamma_2(\cM,\| \cdot \|_{2 \rightarrow 2})} \cdot \Big({\color{Red}\gamma_2(\cN, \| \cdot \|_{2 \rightarrow 2}) + d_F(\cN)} + {\color{Blue}\sqrt{p}~ d_{2 \rightarrow 2}(\cN)}~\Big)\nonumber \\
        & \quad\quad\quad\quad\quad + {\color{Red}\gamma_2(\cN,\| \cdot \|_{2 \rightarrow 2})} \cdot \Big({\color{Red}\gamma_2(\cM, \| \cdot \|_{2 \rightarrow 2}) + d_F(\cM)} + {\color{Blue}\sqrt{p}~ d_{2 \rightarrow 2}(\cM)}~\Big) \nonumber
        \\ & \quad\quad + {\color{Blue}~ \sqrt{p}~ \min \Big\{  d_F(\cM) \cdot d_{2 \rightarrow 2}(\cN),d_F(\cN) \cdot d_{2 \rightarrow 2}(\cM)} \Big\} \nonumber + {\color{Green}p \cdot d_{2 \rightarrow 2}(\cM) \cdot d_{2 \rightarrow 2}(\cN)}~.\Bigg]    
    \end{align*}
\end{theorem}

\subsection{Proof of Theorem~\ref{theo:diagonalLpBound-sumT}}

    We use Lemma~\ref{lemma:decouplingOffDiag} with $F(x) = |x|^p, p \geq 1$ as the convex function and set $B_{j,k} = \langle M_j, N_k \rangle$. Then
    \begin{align}
        \| B_{\cM, \cN}(\bxi^{1:T}) \|_{L_p} & = \E \sup_{M\in\cM, N \in \cN} F\left( \sum_{t=1}^{T}\sum_{\substack{j,k=1\\j \neq k}}^n \bxi_{t,j} \bxi_{t,k} \langle M_j, N_k \rangle  \right)\nonumber\\
        & \leq E \sup_{M\in\cM, N \in \cN} F\left( 4 \sum_{t=1}^{T}\sum_{ \substack{j,k=1\\j \neq k}}^n \bxi_{t,j} \bxi_{t,k} \langle M_j, N_k \rangle  \right)\nonumber\\
        &\lesssim \left\| \sup_{M \in \cM, N \in \cN} \left| \sum_{t=1}^{T}\langle M \bxi_t, N \bxi'_t \rangle \right| \right\|_{L_p}~.
        \label{eq:diag_first_upper_bound-sum}
    \end{align}
    Note that for fixed $M,N,$ the term $ \left|\sum_{t=1}^{T} \langle M \bxi_t, N \bxi'_t \rangle \right|$ conditioned on $\bxi'_t$ is sub-gaussian and therefore its $L_p$ norm can be bounded. However, the $\sup_{M \in \cM, N \in \cN}$ inside the $\| \cdot \|_{L_p}$ does not let us use this approach. The next theorem therefore upper bounds $\left\| \sup_{M \in \cM, N \in \cN} \left| \sum_{t=1}^{T}\langle M \bxi_t, N \bxi'_t \rangle \right| \right\|_{L_p}$ by $\sup_{M \in \cM, N \in \cN} \left\| \sum_{t=1}^{T} \langle M \bxi_t, N \bxi'_t \rangle \right\|_{L_p}$ plus some additional complexity terms. Unlike \cite{krahmer2014suprema} the inner product contains two different matrices $M$ and $N$, and therefore we consider two separate admissible sequences (cf. definition ~\ref{def:gamma_2}) $\{T_r(\cM)\}_{r=0}^{\infty}$ and $\{T_r(\cN)\}_{r=0}^{\infty}$ of $\cM$ and $\cN$ respectively. We then use a generic chaining argument by creating two separate increment sequences for $\cM$ and $\cN$ and is detailed in the following theorem.
    \begin{lemma}
    Let $\bxi_t$ for all $t \in [T]$ be stochastic processes satisfying Assumption~\ref{asmp:xi}, and $\bxi'_t$ be a decoupled tangent sequence to $\bxi_t$. Then, for every $p \geq 1$,
        \begin{align}
            \left\| \sup_{M \in \cM, N \in \cN} \sum_{t=1}^{T}\langle M \bxi_t, N \bxi'_t \rangle \right\|_{L_p}
            &\lesssim \;\gamma_2(\cM,\| \cdot \|_{2 \rightarrow 2}) \cdot \Big[\gamma_2(\cN, \| \cdot \|_{2 \rightarrow 2}) + d_F(\cN) + \sqrt{p} d_{2 \rightarrow 2}(\cN)~\Big]\nonumber \\
            & + \;\gamma_2(\cN,\| \cdot \|_{2 \rightarrow 2}) \cdot \Big[\gamma_2(\cM, \| \cdot \|_{2 \rightarrow 2}) + d_F(\cM) + \sqrt{p} d_{2 \rightarrow 2}(\cM)~\Big] \nonumber
            \\ & \qquad\qquad\qquad\qquad + \sup_{M \in \cM, N \in \cN} \| \langle M \bxi_t, N \bxi'_t \rangle \|_{L_p}~,
        \end{align}
    \label{lemm:offd-lp-sum}
    \end{lemma}

    \begin{proof}[\textbf{Proof of Lemma~\ref{lemm:offd-lp-sum}}]
             Without loss of generality, assume $\cM$ and $\cN$ are finite~\citep{tala14}. 
                Let $\{T_r(\cM)\}_{r=0}^{\infty}$ and $\{T_r(\cN)\}_{r=0}^{\infty}$ be admissible sequences for $\cM$ and $\cN$ for which the minimum in the definition
                of $\gamma_2(\cM, \| \cdot \|_{2 \rightarrow 2})$ and $\gamma_2(\cN, \| \cdot \|_{2 \rightarrow 2})$ are attained respectively. Let 
                \begin{align*}
                    \pi_r M &= d_{2 \rightarrow 2}(M,T_r(\cM)) = \underset{B \in T_r(\cM)}{\argmin} ~\| B - A\|_{2 \rightarrow 2} \qquad \text{and} \qquad \Delta_r M = \pi_r M - \pi_{r-1} M~.\\
                    \pi_r N &= d_{2 \rightarrow 2}(N,T_r(\cN)) \;=\; \underset{B \in T_r(\cN)}{\argmin} ~\| B - A\|_{2 \rightarrow 2} \qquad \text{and} \qquad \Delta_r N = \pi_r N - \pi_{r-1} N~.
                \end{align*}
                For any given $p \geq 1$, let $\ell$ be the largest integer for which $2^{\ell} \leq 2 p$.
                Then,
                \begin{align*}
                    &\sum_{t=1}^{T}\langle M \bxi_t, N \bxi'_t\rangle - \langle (\pi_{\ell} M) \bxi_t, (\pi_{\ell} N) \bxi'_t \rangle = \sum_{t=1}^{T}\sum_{r = \ell}^{\infty}\langle (\pi_{r+1} M) \bxi_t, (\pi_{r+1} N) \bxi'_t\rangle - \langle (\pi_{r} M) \bxi_t, (\pi_{r} N) \bxi'_t \rangle \\
                    & = \sum_{t=1}^{T}\sum_{r = \ell}^{\infty}\langle (\pi_{r}M +  \Delta_{r+1}M) \bxi_t, (\pi_{r+1}N) \bxi'_t\rangle - \langle (\pi_{r} M) \bxi_t, (\pi_{r+1}N - \Delta_{r+1} N) \bxi'_t \rangle \\
                    &= \sum_{t=1}^{T}\sum_{r = \ell}^{\infty}\langle (\pi_{r} M) \bxi_t, (\pi_{r+1} N) \bxi'_t\rangle + \langle (\Delta_{r+1} M) \bxi_t, (\pi_{r+1} N) \bxi'_t\rangle \\
                    & \qquad\qquad\qquad\qquad\qquad\qquad\qquad\qquad\qquad\qquad - \langle (\pi_{r} M) \bxi_t, (\pi_{r+1} N) \bxi'_t \rangle + \langle (\pi_{r} M) \bxi_t, (\Delta_{r+1} N) \bxi'_t \rangle   \\
                    &= \sum_{t=1}^{T}\sum_{r = \ell}^{\infty} \langle (\Delta_{r+1} M) \bxi_t, (\pi_{r+1} N) \bxi'_t\rangle + \langle (\pi_{r} M) \bxi_t, (\Delta_{r+1} N) \bxi'_t \rangle   \\
                \end{align*}
                Now by applying triangle inequality, we have
                \begin{align}
                \left| \sum_{t=1}^{T} \langle M \bxi_t, N \bxi'_t\rangle - \langle (\pi_{\ell} M) \bxi_t, (\pi_{\ell} N) \bxi'_t \rangle \right|
                & \leq \underbrace{\left| \sum_{t=1}^{T}\sum_{r=\ell}^{\infty} \langle (\Delta_{r+1} M) \bxi_t, (\pi_{r+1} N) \bxi'_t \rangle \right|}_{S_1}  
                \nonumber \\ 
                & \qquad\qquad\qquad\qquad + \underbrace{\left| \sum_{t=1}^{T}\sum_{r=\ell}^{\infty} \langle (\pi_{r} M) \bxi_t, (\Delta_{r+1} N) \bxi'_t \rangle \right|}_{S_2}~. 
                \label{eq:l3tri-sum_T}
                \end{align}
                We first consider $S_1$. Let us define
                \begin{equation*}
                    X_r(M,N) = \sum_{t=1}^{T}\langle (\Delta_{r+1} M) \bxi_t, (\pi_{r+1} N) \bxi'_t \rangle~. 
                \end{equation*}
                Conditioning $X_r(M,N)$ on $\bxi'_1,\ldots,\bxi'_T$, we note 
                \begin{align*}
                 X_r(M,N)~ \Big|~\bxi'_1,\ldots,\bxi'_T &=   \sum_{t=1}^{T}\langle (\Delta_{r+1} M) \bxi_t, (\pi_{r+1} N) \bxi'_t \rangle ~\Big|~\bxi'_1,\ldots,\bxi'_T \\
                 & = \sum_{t=1}^{T}\langle  \bxi_t, (\Delta_{r+1} M)^T (\pi_{r+1} N) \bxi'_t \rangle ~\Big|~\bxi'_1,\ldots,\bxi'_T
                \end{align*}
                is a sub-Gaussian random variable and therefore using Azuma-Hoeffding bound~\citep{bolm13,vers18} gives
                \begin{align*}
                    P\left( |X_r(M,N) | > u \bigg\| \sum_{t=1}^{T}(\Delta_{r+1} M)^T (\pi_{r+1} N) \bxi'_t \bigg\|_2 ~ \Big|~ \bxi'_1,\ldots,\bxi'_T~ \right) \leq 2 \exp(-u^2/2)~.
                \end{align*}
                Using $u = t 2^{r/2}$, we get
                \begin{align*}
                P\left( |X_r(M,N)
                | > t 2^{r/2} \bigg\| \sum_{t=1}^{T} (\Delta_{r+1} M)^T (\pi_{r+1} N) \bxi'_t \bigg\|_2 ~ \Big|~ \bxi'_1,\ldots,\bxi'_T~ \right) \leq 2 \exp(-t^2 2^r/2)~.
                \label{eq:azho}
                \end{align*}
                Since
                \begin{align*}
                \left| \sum_{t=1}^{T}(\Delta_{r+1} M)^T (\pi_{r+1} N) \bxi'_t \right| 
                \leq \| \Delta_{r+1} M \|_{2 \rightarrow 2} \sup_{N \in \cN} \Big\| \sum_{t=1}^{T} N \bxi'_t \Big\|_2~.
                \end{align*}
                we have 
                \begin{align*}
                P\left( |X_r(M,N) | > t 2^{r/2} \| \Delta_{r+1} M \|_{2 \rightarrow 2} \sup_{N \in \cN} \Big\| \sum_{t=1}^{T} N \bxi'_t \Big\|_2 ~ \Big|~ \bxi'_1,\ldots,\bxi'_T~ \right) \leq 2 \exp(-t^2 2^r/2)~.
                \end{align*}
                Now, since $|\{\pi_r M : M \in \cM \}| = |T_r(\cM)| \leq 2^{2^r}$ and $|\{\pi_r N : N \in \cM \}| = |T_r(\cN)| \leq 2^{2^r}$, by union bound, we get 
                 \begin{align*}
                P\bigg( \sup_{M \in \cM, N \in \cN}  &  ~\sum_{r=\ell}^{\infty} |X_r(M,N)| > t \left( \sup_{M \in \cM} \sum_{r=\ell}^{\infty} 2^{r/2} \| \Delta_{r+1} M \|_{2 \rightarrow 2} \right) \cdot \sup_{N \in \cN} \Big\| \sum_{t=1}^{T} N \bxi'_t \Big\|_2 ~ \Big|~ \bxi'_1,\ldots,\bxi'_T~ \bigg) \\  
                 & \leq 2 \sum_{r=\ell}^{\infty} |T_{r}(\cM)| \cdot |T_{r+1}(\cM)| \cdot |T_{r+1}(\cN)| \cdot \exp(-t^2 2^r/2) \\
                 & \leq 2 \sum_{r=\ell}^{\infty} 2^{2^{r+2}} \cdot \exp (-t^2 2^r/2) \\
                 & \leq 2 \exp(-2^{\ell} t^2)~,
                \end{align*}
                for all $t \geq t_0$, a constant. Next, note that
                \begin{align*}
                    \sup_{M \in \cM} \sum_{r=\ell}^{\infty} 2^{r/2} \| \Delta_{r+1} M \|_{2 \rightarrow 2} & = \gamma_2(\cM, \| \cdot \|_{2 \rightarrow 2} ).
                \end{align*}
                Therefore we have 
                \begin{align*}
                P\bigg( \sup_{M \in \cM, N \in \cN}  &  ~\sum_{r=\ell}^{\infty} |X_r(M,N)| > t \gamma_2(\cM,  \|\cdot \|_{2 \rightarrow 2}) \sup_{N \in \cN} \Big\| \sum_{t=1}^{T} N \bxi'_t \Big\|_2 ~ \Big|~ \bxi'_1,\ldots, \bxi'_T~ \bigg) 
                \leq 2 \exp(-p t^2)~,
                \end{align*}
                since $p \leq 2^{\ell}$ by construction which implies with $V(\bxi') = \gamma_2(\cM,  \|\cdot \|_{2 \rightarrow 2}) \sup_{N \in \cN} \Big\| \sum_{t=1}^{T} N \bxi'_t \Big\|_2$, for $t \geq t_0$ we have
                \begin{align*}
                    P\left( S_1 \geq t V(\bxi') ~\big|~\bxi' \right) \leq 2 \exp(- p t^2)~.
                \end{align*}
                Note that
                \begin{align*}
                    \| S_1 \|_{L_p}^p & = \E_{\bxi,\bxi'} S_1^p = E_{\bxi'} \int_0^{\infty} p t^{p-1} P(S_1 > t ~\big|~ \bxi') dt~,
                \end{align*}
                and
                \begin{align*}
                    \int_0^{\infty} p t^{p-1} P(S_1 > t ~\big|~ \bxi') dt & \leq c^p V(\bxi')^p + \int_{cV(\bxi')}^{\infty} p t^{p-1} P(S_1 > t ~\big|~ \bxi') dt \\
                    & \leq c^p V(\bxi')^p + V(\bxi')^p \int_{c}^{\infty} p \tau^{p-1} P( S_1 > \tau V(\bxi') | \bxi') d\tau \\
                    & \leq c_1^p V(\bxi')^p~,
                \end{align*}
                where $c \geq t_0,c_1$ are suitable constants that depend on $L$. As a result, $\| S_1 \|_{L_p} \leq c_1 V(\bxi') = c_1 \|V(\bxi)\|_{L_p}$, i.e., we have the following bound on $S_1$.
                \begin{align*}
                    \| S_1 \|_{L_p} \lesssim \gamma_2(\cM,  \|\cdot \|_{2 \rightarrow 2}) \Big\|\sup_{N \in \cN} \Big\| \sum_{t=1}^{T} N \bxi'_t \Big\|_2 \Big\|_{L_p}
                \end{align*}
                 Note that a similar analysis follows for $S_2$, and we can bound $\| S_2 \|_{L_p}$. As a result
                \begin{equation}
                    \| S_1 + S_2 \|_{L_p}  \lesssim \gamma_2(\cM,  \|\cdot \|_{2 \rightarrow 2}) \cdot \bigg\|\sup_{N \in \cN} \Big\| \sum_{t=1}^{T} N \bxi'_t \Big\|_2 \bigg\|_{L_p} \!\!\!+ \gamma_2(\cN,  \|\cdot \|_{2 \rightarrow 2}) \cdot \bigg\|\sup_{M \in \cM} \Big\| \sum_{t=1}^{T} M \bxi'_t \Big\|_2 \bigg\|_{L_p}
                    \label{eq:s1s2-sum_T}
                \end{equation}
                Further, since $| \{ \pi_{\ell} M : M \in \cM \} | \leq 2^{2^{\ell}} \leq \exp(2p)$, and $| \{ \pi_{\ell} N : N \in \cN \} | \leq 2^{2^{\ell}} \leq \exp(2p)$ we have 
                \begin{align*}
                    \E \sup_{M \in \cM, N \in \cN} & \Big| \sum_{t=1}^{T} \langle (\pi_{\ell} M) \bxi_t, (\pi_{\ell} N) \bxi'_t \rangle \Big|^p  \leq \sum_{M \in T_{\ell}(M), N \in T_{\ell}(N)} \E \Big| \sum_{t=1}^{T} \langle M\bxi_t, N \bxi'_t\rangle \Big|^p \\
                    &\leq 2^{2p} 2^{2p} \sup_{M \in \cM, N \in \cN} E \Big| \sum_{t=1}^{T} \langle M \bxi_t, N \bxi'_t \rangle \Big|^p = 2^{4p} \sup_{M \in \cM, N \in \cN}  \Big| \sum_{t=1}^{T} \langle M \bxi_t, N \bxi'_t \rangle \Big|^p~,
                \end{align*}
                so that
                \begin{equation}
                    \left\| \sup_{M \in \cM, N \in \cN} \Big| \sum_{t=1}^{T} \langle (\pi_{\ell} M) \bxi_t, (\pi_{\ell} N) \bxi'_t \rangle \Big|\right\|_{L_p} \leq 16~  \sup_{M \in \cM, N \in \cN} \Big\|\sum_{t=1}^{T} \langle M \bxi_t, N \bxi'_t \rangle \Big\|_{L_p}~.
                    \label{eq:s0-sum_T}
                \end{equation}
                Combining \eqref{eq:l3tri-sum_T}, \eqref{eq:s1s2-sum_T}, and \eqref{eq:s0-sum_T} and using triangle inequality we have
                \begin{align}
                    \left\|  \sup_{M \in \cM, N \in \cN} \sum_{t=1}^{T}\langle M \bxi, N \bxi' \rangle \right\|_{L_p} \!\!\!\! &\lesssim \gamma_2(\cM,  \|\cdot \|_{2 \rightarrow 2}) \cdot \bigg\|\sup_{N \in \cN} \Big\| \sum_{t=1}^{T} N \bxi'_t \Big\|_2 \bigg\|_{L_p} \nonumber\\
                    & + \gamma_2(\cN,  \|\cdot \|_{2 \rightarrow 2}) \cdot \bigg\|\sup_{M \in \cM} \Big\| \sum_{t=1}^{T} M \bxi'_t \Big\|_2 \bigg\|_{L_p}\nonumber\\
                    & \qquad\qquad + \sup_{M \in \cM, N \in \cN} \Big\|\sum_{t=1}^{T} \langle M \bxi_t, N \bxi'_t \rangle \Big\|_{L_p}~.
                    \label{eq:chaing_double_lp}
                \end{align}
                Note that unlike in the proof of Theorem~\ref{thm:cross_product}, we cannot use Theorem 3.5 from \citep{krmr14} to control $\Big\|\sup_{M \in \cM} \Big\| \sum_{t=1}^{T} M \bxi'_t \Big\|_2 \Big\|_{L_p}$. Below we derive new results to handle them.
                \begin{theorem}
                \label{thm:N_C-sumT}
                 Let $\bxi_t$ for all $t \in [T]$ be stochastic processes satisfying Assumption~\ref{asmp:xi}. Then
                    \begin{align*}
                        \Big\|\sup_{M \in \cM} \Big\| &\sum_{t=1}^{T} M \bxi_t \Big\|_2 \Big\|_{L_p} \leq \gamma_{2}(\cM, \|\cdot\|) + \sqrt{T} d_{F}(\cM) + \sqrt{pT} d_{2 \rightarrow 2}(\cM)
                    \end{align*}
                \end{theorem}
\begin{proof}
        We apply Theorem 2.3 from \cite{krahmer2014suprema} with the set $S = \{ M^\top x : x \in B_2^n,\, M \in \mathcal{M} \}$. Since $\bxi_t$ is $L$-subgaussian for all $t \in [T]$ we obtain
        \begin{align*}
            \Big\|\sup_{M \in \cM} \Big\| &\sum_{t=1}^{T} M \bxi_t \Big\|_2 \Big\|_{L_p}
            = \left( \mathbb{E} \sup_{M \in \mathcal{M},\, x \in B_2^n} | \sum_{t=1}^{T}\langle M \bxi_t, x \rangle |^p \right)^{1/p}
            = \left( \mathbb{E} \sup_{u \in S} |  \sum_{t=1}^{T} \langle \bxi_t, u \rangle |^p \right)^{1/p} \\
            &\lesssim_L  \E\sup_{M \in \cM} \Big\| \sum_{t=1}^{T} M \g_t \Big\|_2  + \sup_{u \in S} \left( \mathbb{E} | \sum_{t=1}^{T} \langle \bxi_t, u \rangle |^p \right)^{1/p} \\
            &\lesssim_L \E\sup_{M \in \cM} \Big\| \sum_{t=1}^{T} M \g_t \Big\|_2 + \sqrt{pT} \sup_{M \in \mathcal{M},\, x \in B_2^n} \| M^\top x \|_2 \\
            &\lesssim_L \E\sup_{M \in \cM} \Big\| \sum_{t=1}^{T} M \g_t \Big\|_2 + \sqrt{pT} d_{2 \to 2}(\mathcal{M})
        \end{align*}
        Next we handle $\E\sup_{M \in \cM} \Big\| \sum_{t=1}^{T} M \g_t \Big\|_2$. Using Theorem 2.5 in \cite{krahmer2014suprema} we have
        \begin{align*}
            \| C_{\cM}(\g^{1:T}) \|_{L_p}
            &= \left\| \sup_{M \in \mathcal{M}} \left|\sum_{t=1}^{T} \sum_{\substack{j,k \\ j \neq k}} g_{t,j} g_{t,k} \langle M^j, N^k \rangle + \sum_{t=1}^{T}\sum_j (g_{t,j}^2 - 1) \| M^j \|_2^2 \right| \right\|_{L_p} \\
            &\lesssim \left\| \sup_{M \in \mathcal{M}} \left|\sum_{t=1}^{T} \sum_{j,k} g_{t,j} g_{t,k}' \langle M^j, M^k \rangle \right| \right\|_{L_p}
            = \left\| \sup_{M \in \mathcal{M}} \sum_{t=1}^{T}\left \langle M \g_t, M \g'_t \right\rangle \right\|_{L_p} \\
            &\lesssim_L \gamma_2(\cM, \| \cdot \|_{2 \to 2}) \, \Big\|\sup_{M \in \cM} \Big\| \sum_{t=1}^{T} M \bxi_t \Big\|_2 \Big\|_{L_p} + \sup_{M \in \mathcal{M}} \left\| \sum_{t=1}^T \left\langle  M \g_t, M \g'_t \right\rangle \right\|_{L_p}.
        \end{align*}
        where in the last inequality we have used \eqref{eq:chaing_double_lp} with a single set $\cM$. 

        Fix $M \in \cM$ and set $S = \{ M^\top M x : x \in B_2^n \}$. Since the random vectors $\g_t, t \in [T]$ are subgaussian, the random variable $\sum_{t=1}^{T} \langle \g_t, M^\top M \g'_t \rangle$ is subgaussian conditionally on $\g'_t, t \in [T]$. Therefore,
        \begin{align*}
            \left\| \sum_{t=1}^T \left\langle  M \g_t, M \g'_t \right\rangle \right\|_{L_p}
            &= \left( \mathbb{E}_{\g'_t} \left( \left( \mathbb{E}_{\g_t} \big(\sum_{t=1}^{T}\langle \g_t, M^\top M \g'_t \rangle\big)^p \right)^{1/p} \right)^p \right)^{1/p} \\
            &\lesssim \left( \mathbb{E}_{\g'_t} \left( L \sqrt{p}^p \norm{ \sum_{t=1}^{T} M^\top M \g'_t}_2^p \right) \right)^{1/p} \\
            &= L \sqrt{p} \left( \mathbb{E}_{\g'_t} \norm{ \sum_{t=1}^{T} M^\top M \g'_t}_2^p \right)^{1/p} \\
            &= L \sqrt{p} \left( \mathbb{E}_{\g'_t} \sup_{y \in S} | \sum_{t=1}^{T}\langle y, \g'_t \rangle |^p \right)^{1/p}.
        \end{align*}

Now we use Theorem 2.3 from \citep{krahmer2014suprema}. The first term in the rhs can be bounded as
\begin{align*}
\mathbb{E}_{\g'_t} \sup_{y \in S} | \sum_{t=1}^{T}\langle y, \g'_t \rangle |^p
&= \mathbb{E} \norm{ \sum_{t=1}^T M^\top M \g'_t }_2 
\leq \left( \mathbb{E} \norm{ \sum_{t=1}^T M^\top M \g'_t }_2^2 \right)^{1/2} \\
&\leq \sqrt{T} \norm{ M^\top M }_F 
= \sqrt{T}\norm{M}_F \norm{M}_{2 \to 2}.
\end{align*}

For the second term,
\begin{align*}
\sup_{y \in S} \left( \mathbb{E} |\sum_{t=1}^T \langle y, \bxi'_t \rangle |^p \right)^{1/p}
& = \sup_{z \in B_2^n} \left( \mathbb{E} | \sum_{t=1}^T\langle M^\top M z, \bxi'_t \rangle |^p \right)^{1/p}\\
& \lesssim L \sup_{z \in B_2^n} \sqrt{pT} \norm{ M^\top M z }_2
= L \sqrt{pT} \norm{ M }_{2 \to 2}^2.
\end{align*}

Hence applying Theorem 2.3 from \cite{krahmer2014suprema} and taking the supremum over $M \in \cM$ we get
\begin{align*}
    \left\| \sum_{t=1}^T \left\langle  M \g_t, M \g'_t \right\rangle \right\|_{L_p} \lesssim \sqrt{T} \left( \sqrt{p} d_{F}(\cM)d_{2\rightarrow 2}(\cM) + p d_{2 \rightarrow 2}(\cM)\right)
\end{align*}
Therefore,
\begin{align*}
    \E(\sup_{M \in \cM} \Big\| \sum_{t=1}^{T} M \g_t \Big\|_2)^2 \leq C_{\cM}(\g^{1:T}) + d_{F}^2(\cM) \lesssim \gamma_{2}(\cM, \|\cdot\|)\E(\sup_{M \in \cM} \Big\| \sum_{t=1}^{T} M \g_t \Big\|_2) + \sqrt{T} \left( d^2_{F}(\cM)\right)
\end{align*}
which implies
\begin{align*}
    \E(\sup_{M \in \cM} \Big\| \sum_{t=1}^{T} M \g_t \Big\|_2) \leq \gamma_{2}(\cM, \|\cdot\|) + \sqrt{T}d_{F}(\cM)
\end{align*}
Combining all these we get
\begin{align*}
    \Big\|\sup_{M \in \cM} \Big\| &\sum_{t=1}^{T} M \bxi_t \Big\|_2 \Big\|_{L_p} \leq \gamma_{2}(\cM, \|\cdot\|) + \sqrt{T} d_{F}(\cM) + \sqrt{pT} d_{2 \rightarrow 2}(\cM)
\end{align*}
which completes the proof of Theorem~\ref{thm:N_C-sumT}.
 \end{proof}               
Therefore
    \begin{align}
        \| B_{\cM, \cN}(\bxi^{1:T}) \|_{L_p} &\lesssim \gamma_2(\cM,\| \cdot \|_{2 \rightarrow 2}) \cdot \Big[\gamma_2(\cN, \| \cdot \|_{2 \rightarrow 2}) + \sqrt{T}d_F(\cN) + \sqrt{pT} d_{2 \rightarrow 2}(\cN)~\Big]\nonumber \\
            & + \;\gamma_2(\cN,\| \cdot \|_{2 \rightarrow 2}) \cdot \Big[\gamma_2(\cM, \| \cdot \|_{2 \rightarrow 2}) + \sqrt{T}d_F(\cM) + \sqrt{pT} d_{2 \rightarrow 2}(\cM)~\Big] \nonumber
            \\ & \qquad\qquad\qquad\qquad + \sup_{M \in \cM, N \in \cN} \| \langle M \bxi, N \bxi' \rangle \|_{L_p}
            \label{eq:diagBound2-sumT}
    \end{align}
    Next we consider $\displaystyle\sup_{M \in \cM, N \in \cN}\Big\| \sum_{t=1}^{T}\langle M \bxi, N \bxi' \rangle \Big\|_{L_p}$ and bound it using the following Lemma.
    \begin{lemma}
        \label{lemm:offd-lp-2nd-sumT}
        Let $\bxi$ be a stochastic process satisfying Assumption~\ref{asmp:xi}, and let $\bxi'$ be a decoupled tangent sequence. Then, for every $p \geq 1$,
        \begin{align}
            \sup_{M \in \cM, N \in \cN}\Big\| \sum_{t=1}^{T} \langle M \bxi, N \bxi' \rangle \Big\|_{L_p} &\lesssim \min \Big\{ \sqrt{p} \cdot d_F(\cM) \cdot d_{2 \rightarrow 2}(\cN), \sqrt{p} \cdot d_F(\cN) \cdot d_{2 \rightarrow 2}(\cM) \Big\} \nonumber\\
            & \qquad\qquad + p \cdot d_{2 \rightarrow 2}(\cM) \cdot d_{2 \rightarrow 2}(\cN)~.
            \label{eq:offd-lp-2nd-sumT}
        \end{align}
    \label{lemm:offd-lp-2nd-sum}
    \end{lemma} 
    \begin{proof}[\textbf{Proof of Lemma~\ref{lemm:offd-lp-2nd-sum}}]
    Fix $M \in \cM, N \in \cN$ and let $S = \{M^\top N x : x \in B^n_2, M \in \cM, N\in \cN\}$, where $B^n_2 = \{x \in \R^n: \|x\|_2 \leq 1 \}$  \rdcomment{Define $B^n_2$}. Conditioned on $\bxi_t$, the random variable $\sum_{t=1}^{T} \langle \bxi_t, M^\top N \bxi'_t\rangle$ is sub-gaussian. Therefore for some global constant $\tilde{C}>0$, we have \rdcomment{cite Vershynin?}
    \begin{align*}
        \Big\|\sum_{t=1}^{T} \langle M \bxi, N \bxi' \rangle \Big\|_{L_p} 
        &\lesssim L\sqrt{p} \Bigg( \E_{\bxi'} \sup_{y \in S} |\sum_{t=1}^{T}\langle y, \bxi'_t \rangle|^p \Bigg)^{1/p}
    \end{align*}
    Now using Theorem 2.3 from \cite{krmr14} we have for every $p\geq 1$
    \begin{align*}
        \Bigg( \E_{\bxi'} \sup_{y \in S} |\sum_{t=1}^{T}\langle y, \bxi'_t \rangle|^p \Bigg)^{1/p} \lesssim \Big(\E_{\mathbf{g}_t} \sup_{y \in S} | \sum_{t=1}^{T}\langle \mathbf{g}_t,y \rangle| + \sup_{y \in S} (\E_{\bxi'_t} |\sum_{t=1}^{T}\langle \bxi'_t,y \rangle|^p)^{1/p}\Big)
    \end{align*}
    where $\mathbf{g}_t, t \in [T]$ are standard Gaussian vector. The first term in the rhs can be bounded as follows:
    \begin{align*}
        \E_{\mathbf{g}_t} \sup_{y \in S} |\sum_{t=1}^{T}\langle \mathbf{g}_t,y \rangle| &= \E_{\mathbf{g}} \| \sum_{t=1}^{T} M^\top N \g_t\|_2 \leq (\E_{\mathbf{g}} \| \sum_{t=1}^{T} M^\top N \g_t\|_2^2)^{1/2} \leq \sqrt{T}\|M^\top N\|_{F} \\
        &\leq \sqrt{T} \min \Big\{ \|M\|_{2\rightarrow 2} \|N\|_F, \|N\|_{2\rightarrow 2} \|M\|_F \Big\}
    \end{align*}
    Next the second term in the rhs can be bounded as follows:
    \begin{align*}
        \sup_{y \in S} (\E_{\bxi'} |\sum_{t=1}^{T} \langle \bxi'_t,y \rangle|^p)^{1/p} \leq L \sup_{z \in B^p_2} \sqrt{pT} \|M^\top N z\|_2 \leq L \sqrt{pT} \|M\|_{2 \rightarrow 2} \|N\|_{2 \rightarrow 2}.
    \end{align*}
Combining all the above bounds we get:
\begin{align}
     & \Big\| \langle M \bxi, N \bxi' \rangle \Big\|_{L_p} \lesssim \sqrt{pT} \min \Big\{ \|M\|_{2\rightarrow 2} \|N\|_F, \|N\|_{2\rightarrow 2} \|M\|_F \Big\} +  {p\sqrt{T}} \|M\|_{2 \rightarrow 2} \|N\|_{2 \rightarrow 2}
\end{align}
Now recall that for the set $\cM$, we have $d_F(\cM) = \sup_{M \in \cM} \| M \|_F$, and $d_{2 \rightarrow 2}(\cM) = \sup_{A \in \cM} \| A \|_{2 \rightarrow 2}$, which implies
\begin{align*}
    \sup_{M \in \cM, N \in \cN} \Big\| \sum_{t=1}^{T}\langle M \bxi_t, N \bxi'_t \rangle \Big\|_{L_p} & \lesssim \Bigg( \sqrt{pT}\min \Big\{ d_{2\rightarrow 2}(\cM) d_F(\cN), d_{2\rightarrow 2}(\cN) d_F(\cM) \Big\} \\
    &\qquad\qquad + p\sqrt{T} d_{2\rightarrow 2}(\cM) d_{2\rightarrow 2}(\cN)\Bigg).
\end{align*}
\end{proof}
Combining Lemma~\ref{lemm:offd-lp-2nd-sumT} with \eqref{eq:diagBound2-sumT} completes the proof of Theorem~\ref{theo:diagonalLpBound-sumT}.
\end{proof}
\subsection{The Diagonal Term \texorpdfstring{$D_{\cM, \cN}({\boldsymbol{\xi^{1:T}}})$}{textxiSumT}}
\label{subsection:diagonal-sum}

For the diagonal term, we have the following main result:

\begin{restatable}{theorem}{theoremDiagonal} 
\label{theo:offdiagonalLpBound-sumT}
Let $\bxi$ be a stochastic process satisfying Assumption~\ref{asmp:xi}. Then, for all $p \geq 1$, we have
    \begin{align*}
    &\left\| D_{\cM,\cN}(\bxi^{1:T}) \right\|_{L_p} \lesssim \sqrt{T}\Bigg[{\color{Red}\gamma_2(\cM,\| \cdot \|_{2 \rightarrow 2})} \cdot \Big({\color{Red}\gamma_2(\cN, \| \cdot \|_{2 \rightarrow 2}) + d_F(\cN)} + {\color{Blue}\sqrt{p}~ d_{2 \rightarrow 2}(\cN)}~\Big)\nonumber \\
        & \quad\quad\quad\quad\quad + {\color{Red}\gamma_2(\cN,\| \cdot \|_{2 \rightarrow 2})} \cdot \Big({\color{Red}\gamma_2(\cM, \| \cdot \|_{2 \rightarrow 2}) + d_F(\cM)} + {\color{Blue}\sqrt{p}~ d_{2 \rightarrow 2}(\cM)}~\Big) \nonumber
        \\ & \quad\quad + {\color{Blue}~ \sqrt{p}~ \min \Big\{  d_F(\cM) \cdot d_{2 \rightarrow 2}(\cN),d_F(\cN) \cdot d_{2 \rightarrow 2}(\cM)} \Big\} \nonumber + {\color{Green}p \cdot d_{2 \rightarrow 2}(\cM) \cdot d_{2 \rightarrow 2}(\cN)}~.\Bigg]    
    \end{align*}
\end{restatable}
\subsubsection{Proof of Theorem~\ref{theo:diagonalLpBound}}

By definition of $D_{\cM, \cN}(\bxi)$ and {from Lemma 9 in \cite{banerjee2019random}}
, we have
\begin{align*}
    \|D_{\cM, \cN}(\bxi^{1:T})\|_{L_p} & = \left\| \sup_{M\in\cM, N\in\cN} \Big|\sum_{t=1}^{T} \sum_{j=1}^n (|\bxi_{t,j}|^2 - \E|\bxi_{t,j}|^2) \langle M_j, N_j \rangle \Big| \right\|_{L_p}\!\! 
     \\
     & \leq 2 \left\| \sup_{M\in\cM, N\in\cN} \Big|\sum_{t=1}^{T} \sum_{j=1}^n \eps_{t,j} |\bxi_{t,j}|^2  \langle M_j, N_j \rangle \Big| \right\|_{L_p},
\end{align*}
where $\{ \eps_{t,j}\}$ is a set of independent Rademacher variables independent of $\bxi$. Let $\{ g_j \}$ be a sequence of independent Gaussian random variables. Since $\bxi_{t,j}$ is a $L$-sub-Gaussian random variable~\citep{vers18}, there is an absolute constant $c$ such that for all $t > 0$
\begin{align*}
    \P \left( |\xi_j|^2 \geq t L^2 \right) & \leq c \P(g_j^2 \geq t)~.
\end{align*}
Then, from contraction of stochastic processes (\cite[Lemma~{4.6}]{leta91}), we have
\begin{align}
    \| & D_{\cM, \cN}(\bxi^{1:T})) \|_{L_p}  \leq 2 \left\| \sup_{M\in\cM, N\in\cN} \Big| \sum_{t=1}^{T}\sum_{j=1}^n \eps_{t,j} |\bxi_{t,j}|^2  \langle M_j, N_j \rangle \Big| \right\|_{L_p} \notag\\
    & \leq 2c{L^{2}} \left\| \sup_{M\in\cM, N\in\cN} \Big|\sum_{t=1}^{T} \sum_{j=1}^n \eps_{t,j} |g_{t,j}|^2  \langle M_j, N_j \rangle \Big| \right\|_{L_p}\notag \\
    & \overset{(a)}{\leq} 2c{L^{2}}\left\| \sup_{M\in\cM, N\in\cN} \Big|\sum_{t=1}^{T} \sum_{j=1}^n \eps_{t,j} (|g_{t,j}|^2 - 1)  \langle M_j, N_j \rangle \Big| \right\|_{L_p}  + 2c{L^{2}}\left\| \sup_{M\in\cM, N\in\cN} \Big|\sum_{t=1}^{T} \sum_{j=1}^n \eps_{t,j}   \langle M_j, N_j \rangle \Big| \right\|_{L_p}\notag \\
    & \overset{(b)}{\leq} 4c{L^{2}}\left\| \sup_{M\in\cM, N\in\cN} \Big|\sum_{t=1}^{T} \sum_{j=1}^n (|g_{t,j}|^2 - 1)  \langle M_j, N_j \rangle \Big| \right\|_{L_p}  + 2c{L^{2}}\left\| \sup_{M\in\cM, N\in\cN} \Big|\sum_{t=1}^{T} \sum_{j=1}^n \eps_{t,j}   \langle M_j, N_j \rangle \Big| \right\|_{L_p} \notag \\ 
    & \leq {4cL^{2}} \left\| D_{\cM, \cN}(\g^{1:T}) \right\|_{L_p} + 2c{L^{2}}\left\| \sup_{M\in\cM, N\in\cN} \Big| \sum_{t=1}^{T}\sum_{j=1}^n \eps_{t,j}   \langle M_j, N_j \rangle \Big| \right\|_{L_p}~,
\label{eq:diagonal_1-sumT}
\end{align}
where (a) follows from Jensen's inequality and since $E|g_j|^2 = 1$, and (b) follows by de-symmetrization following {\cite[Lemma 11]{banerjee2019random}} 
and since the convex function here is 1-Lipschitz.

By triangle inequality, we have 
\begin{equation}
\begin{split}
    \left\| D_{\cM,\cN}(\g^{1:T}) \right\|_{L_p} & \leq \left\| C_{\cM,\cN}(\g^{1:T}) \right\|_{L_p} + \left\| B_{\cM,\cN}(\g^{1:T}) \right\|_{L_p}
 \end{split}
 \label{eq:diag11-sumT}
\end{equation}
In order to handle $\left\|C_{\mathcal{M},\mathcal{N}}(\g^{1:T})\right\|_{L_{p}}$, we use our new Theorem~\ref{thm:decouple}.
We have
\begin{align*}
\left\| C_{\cM,\cN}(\mathbf{g}^{1:T}) \right\|_{L_p} 
&= 
\sup_{M \in \cM, N \in \cN} \left|\sum_{t=1}^{T} \sum_{\substack{j,k=1\\j \neq k}}^n \mathbf{g}_{t,j} \mathbf{g}_{t,k} \langle M_j, N_k \rangle + \sum_{t=1}^{T}\sum_{j=1}^n (|\mathbf{g}_{t,j}|^2 - \E|\mathbf{g}_{t,j}|^2) \langle M_j, N_j \rangle \right|
\\
&\overset{(a)}{\leq} C
\left\| 
\sup_{M \in \cM, N \in \cN} 
\left| \sum_{t=1}^{T}\sum_{\substack{j,k=1}}^n \mathbf{g}_{t,j} \mathbf{g'}_{t,k} \langle M_j, N_k \rangle\right| \right\|_{L_p}
= 
\left\| 
\sup_{M \in \cM, N \in \cN}  
\left| \sum_{t=1}^{T}
\langle M\mathbf{g}_t, N\mathbf{g'}_t \rangle 
\right| 
\right\|_{L_p}
\\
&\overset{(b)}{\lesssim}  \;\gamma_2(\cM,\| \cdot \|_{2 \rightarrow 2}) \cdot \Big[\gamma_2(\cN, \| \cdot \|_{2 \rightarrow 2}) + d_F(\cN) + \sqrt{p} d_{2 \rightarrow 2}(\cN)~\Big]\nonumber \\
    & \quad + \;\gamma_2(\cN,\| \cdot \|_{2 \rightarrow 2}) \cdot \Big[\gamma_2(\cM, \| \cdot \|_{2 \rightarrow 2}) + d_F(\cM) + \sqrt{p} d_{2 \rightarrow 2}(\cM)~\Big] \nonumber
    \\ & \quad +  \min \Big\{ \sqrt{p} \cdot d_F(\cM) \cdot d_{2 \rightarrow 2}(\cN), \sqrt{p} \cdot d_F(\cN) \cdot d_{2 \rightarrow 2}(\cM) \Big\} \nonumber\\
        & \qquad\qquad + p \cdot d_{2 \rightarrow 2}(\cM) \cdot d_{2 \rightarrow 2}(\cN)~.
\end{align*}
{where (a) uses Theorem \ref{thm:decouple}, and (b) holds because of Lemma~\ref{lemm:offd-lp-sum} and Lemma~\ref{lemm:offd-lp-2nd-sum}}. Term $\left\| B_{\cM,\cN}(\g^{1:T}) \right\|_{L_p}$ can be bounded using Theorem~\ref{theo:offdiagonalLpBound-sumT} thus giving
\begin{align}
    \left\| D_{\cM,\cN}(\g^{1:T}) \right\|_{L_p} \lesssim \sqrt{T}\Bigg[&\gamma_2(\cM,\| \cdot \|_{2 \rightarrow 2}) \cdot \Big[\gamma_2(\cN, \| \cdot \|_{2 \rightarrow 2}) + d_F(\cN) + \sqrt{p} d_{2 \rightarrow 2}(\cN)~\Big]\nonumber \\
    & \quad + \sqrt{T} \gamma_2(\cN,\| \cdot \|_{2 \rightarrow 2}) \cdot \Big[\gamma_2(\cM, \| \cdot \|_{2 \rightarrow 2}) + d_F(\cM) + \sqrt{p} d_{2 \rightarrow 2}(\cM)~\Big] \nonumber
    \\ & \quad + \min \Big\{ \sqrt{p} \cdot d_F(\cM) \cdot d_{2 \rightarrow 2}(\cN), \sqrt{p} \cdot d_F(\cN) \cdot d_{2 \rightarrow 2}(\cM) \Big\} \nonumber\\
        & \qquad\qquad + p \cdot d_{2 \rightarrow 2}(\cM) \cdot d_{2 \rightarrow 2}(\cN)~.\Bigg] \label{eq:diagonal_2-sumT}   
\end{align}
Next we bound the second term $\displaystyle \left\| \sup_{M\in\cM, N\in\cN} \Big| \sum_{t=1}^{T}\sum_{j=1}^n \eps_{t,j}   \langle M_j, N_j \rangle \Big| \right\|_{L_p}$ using our new Theorem~\ref{theo:double_tree}.

{According to \cite[Lemma 7]{banerjee2019random}, 
\begin{align}
    &\left\| \sup_{M \in \mathcal{M}, N \in \mathcal{N}} 
\left| \sum_{t=1}^{T} \sum_{j=1}^n \varepsilon_{t,j} \langle M_j, N_j \rangle \right| \right\|_{L_p}\notag\\
=&\left(\mathbb{E}_{\varepsilon}\left[\sup_{M \in \mathcal{M}, N \in \mathcal{N}} 
\left|\sum_{t=1}^{T} \sum_{j=1}^n \varepsilon_{t,j} \langle M_j, N_j \rangle \right|\right]^{p}\right)^{1/p}\notag\\
=&\left(\mathbb{E}_{\varepsilon}\left[\sup_{M \in \mathcal{M}, N \in \mathcal{N}} 
\left| \sum_{t=1}^{T}\sum_{j=1}^n \varepsilon_{t,j} \langle M_j, N_j \rangle \right|^{p}\right]\right)^{1/p}\notag\\
\lesssim&\mathbb{E}_{\g}\left[\sup_{M \in \mathcal{M}, N \in \mathcal{N}} 
\left|\sum_{t=1}^{T} \sum_{j=1}^n \g_{t,j} \langle M_j, N_j \rangle \right|\right]+\sup_{M\in\mathcal{M},N\in\mathcal{N}}\left(\mathbb{E}_{\varepsilon}\left[\left|\sum_{t=1}^{T} \sum_{j=1}^n \varepsilon_{t,j} \langle M_j, N_j \rangle \right|^{p}\right]\right)^{1/p}\label{eq:rest_1-sumT}
\end{align}
in which $\g_{t,j}\sim\mathcal{N}(0,1)$ are independent. For the first term, define $\displaystyle X_{(M,N)} = \left| \sum_{t=1}^{T}\sum_{j=1}^n \g_{t,j} \langle M_j, N_j \rangle \right|$. Then
\begin{align*}
\left| X_{(M^1, N)} - X_{(M^2, N)} \right|
&= \left| \left|\sum_{t=1}^{T}\sum_{j=1}^n \g_{t,j} \left\langle M_j^1, N_j \right\rangle\right| - \left|\sum_{t=1}^{T}\sum_{j=1}^n \g_{t,j} \left\langle M_j^2, N_j \right\rangle \right| \right| \\
&= \left|\sum_{t=1}^{T} \sum_{j=1}^n \g_{t,j} \left\langle M_j^1 - M_j^2, N_j \right\rangle \right| \\
&\leq \sum_{t=1}^{T}\sum_{j=1}^n \g_{t,j} \left\langle M_j^1 - M_j^2, N_j \right\rangle
\end{align*}
Since $\g_j$ are standard normal random variables, therefore with probability $1 - 2e^{-u^2/2}$,
\begin{align*}
\left| X_{(M^1, N)} - X_{(M^2, N)} \right|
&\leq u \sqrt{T}\left(\sum_{j=1}^{n}\langle M_j^1 - M_j^2, N_j \rangle^{2}\right)^{\frac{1}{2}}\\
&\leq u \sqrt{T} \left(\sum_{j=1}^{n}\left\|M_j^1 - M_j^2\right\|_{2}^{2}\left\|N_j \right\|_{2}^{2}\right)^{\frac{1}{2}}\\
&\leq u \sqrt{T} d_{F}\left(\mathcal{N}\right)\left\|M^{1}-M^{2}\right\|_{2\to2}
\end{align*}
Similarly with probability $1 - 2e^{-u^2/2}$,
\begin{align*}
\left| X_{(M, N^1)} - X_{(M, N^2)} \right|\leq u\sqrt{T}d_{F}\left(\mathcal{M}\right)\left\|N^{1}-N^{2}\right\|_{2\to2}
\end{align*}
Using Theorem~\ref{theo:double_tree}, with probability $1 - 4e^{-u^2/2}$, 
\begin{align*}
\sup_{M \in \mathcal{M}, N \in \mathcal{N}} 
\left| \sum_{t=1}^{T} \sum_{j=1}^n \g_{t,j} \langle M_j, N_j \rangle \right|
\lesssim u\sqrt{T} \left( d_{F} (\cN)\gamma_2(\mathcal{M}, \|\cdot\|_{2\rightarrow 2}) + d_{F} (\cM)\gamma_2(\mathcal{N}, \|\cdot\|_{2\rightarrow 2}) \right).
\end{align*}
so we have
\begin{align}
    \mathbb{E}_{\g}\left[\sup_{M \in \mathcal{M}, N \in \mathcal{N}} 
\left| \sum_{t=1}^{T} \sum_{j=1}^n \g_j \langle M_j, N_j \rangle \right|\right]\lesssim \sqrt{T}(d_{F} (\cN)\gamma_2(\mathcal{M}, \|\cdot\|_{2\rightarrow 2}) + d_{F} (\cM)\gamma_2(\mathcal{N}, \|\cdot\|_{2\rightarrow 2})) .\label{eq:rest_2-sumT}
\end{align}
For the second term, for each $M\in\mathcal{M},N\in\mathcal{N}$, since $\varepsilon_{j}$ are Rademacher random variables, therefore with probability at least $1-2e^{-u^{2}}$,
\begin{align*}
    \left|\sum_{t=1}^{T} \sum_{j=1}^n \varepsilon_{t,j} \langle M_j, N_j \rangle \right|&\leq u\sqrt{T}\left(\sum_{j=1}^{n}\left\langle M_{j},N_{j}\right\rangle^{2}\right)^{\frac{1}{2}}\leq u\sqrt{T}\left(\sum_{j=1}^{n}\left\|M_{j}\right\|_{2}^{2}\left\|N_{j}\right\|_{2}^{2}\right)^{\frac{1}{2}}\\&\leq \sqrt{T}\min\left\{d_{F}\left(\mathcal{M}\right)d_{2\to2}\left(\mathcal{N}\right),d_{F}\left(\mathcal{N}\right)d_{2\to2}\left(\mathcal{M}\right)\right\}.
\end{align*}
According to \cite[Proposition 2.5.2]{vers18},
\begin{align*}
    \left(\mathbb{E}_{\varepsilon}\left[\left|\sum_{t=1}^{T} \sum_{j=1}^n \varepsilon_{t,j} \langle M_j, N_j \rangle \right|^{p}\right]\right)^{1/p}\lesssim \sqrt{pT}\min\left\{d_{F}\left(\mathcal{M}\right)d_{2\to2}\left(\mathcal{N}\right),d_{F}\left(\mathcal{N}\right)d_{2\to2}\left(\mathcal{M}\right)\right\},
\end{align*}
Using \eqref{eq:rest_2-sumT} and \eqref{eq:rest_1-sumT}, we have
\begin{align}
&\left\| \sup_{M \in \mathcal{M}, N \in \mathcal{N}} 
\left|\sum_{t=1}^{T} \sum_{j=1}^n \varepsilon_{t,j} \langle M_j, N_j \rangle \right| \right\|_{L_p}\notag\\
\lesssim&\mathbb{E}_{\g_t}\left[\sup_{M \in \mathcal{M}, N \in \mathcal{N}} 
\left|\sum_{t=1}^{T} \sum_{j=1}^n \g_{t,j} \langle M_j, N_j \rangle \right|\right] +  
\sup_{M\in\mathcal{M},N\in\mathcal{N}}\left(\mathbb{E}_{\varepsilon_t}\left[\left| \sum_{t=1}^{T} \sum_{j=1}^n \varepsilon_{t,j} \langle M_j, N_j \rangle \right|^{p}\right]\right)^{1/p}\notag\\
\lesssim& \sqrt{T}\Big[d_{F} (\cN)\gamma_2(\mathcal{M}, \|\cdot\|_{2\rightarrow 2}) + d_{F} (\cM)\gamma_2(\mathcal{N}, \|\cdot\|_{2\rightarrow 2})\Big] \notag\\
&+ \sqrt{pT}\min\left\{d_{F}\left(\mathcal{M}\right)d_{2\to2}\left(\mathcal{N}\right),d_{F}\left(\mathcal{N}\right)d_{2\to2}\left(\mathcal{M}\right)\right\}\label{eq:diagonal_3-sumT}
\end{align}
Substituting \eqref{eq:diagonal_2-sumT} and \eqref{eq:diagonal_3-sumT} into \eqref{eq:diagonal_1-sumT}, we get 
\begin{align*}
    \left\| D_{\cM,\cN}(\bxi^{1:T}) \right\|_{L_p} \lesssim \sqrt{T}\Bigg[&\gamma_2(\cM,\| \cdot \|_{2 \rightarrow 2}) \cdot \Big[\gamma_2(\cN, \| \cdot \|_{2 \rightarrow 2}) + d_F(\cN) + \sqrt{p} d_{2 \rightarrow 2}(\cN)~\Big]\nonumber \\
    & \quad + \gamma_2(\cN,\| \cdot \|_{2 \rightarrow 2}) \cdot \Big[\gamma_2(\cM, \| \cdot \|_{2 \rightarrow 2}) + d_F(\cM) + \sqrt{p} d_{2 \rightarrow 2}(\cM)~\Big] \nonumber
    \\ & \quad + \sqrt{p} \min \Big\{  d_F(\cM) \cdot d_{2 \rightarrow 2}(\cN),d_F(\cN) \cdot d_{2 \rightarrow 2}(\cM) \Big\} \nonumber\\
        & \qquad\qquad + p \cdot d_{2 \rightarrow 2}(\cM) \cdot d_{2 \rightarrow 2}(\cN)~.\Bigg]    
\end{align*}
That completes the proof of Theorem~\ref{theo:diagonalLpBound-sumT}.}

\section{Sketched Contextual Bandits}
\label{sec:app_bandits}
In a bandit problem, a learner needs to make sequential decisions over $T$ time steps.
At any round $t \in [T]$, the learner observes the context/action set $\cA_{t} \subseteq \bbR^d$. 
The learner chooses an arm $a_t$ and then the associated reward of the arm $r(a_t) \in [0,1]$ is observed. We make the following assumption on the reward.

\begin{asmp}[\textbf{Linear Reward}]
The reward $r(a_t)$ is given by
\(
r(a_t) = \langle a_t, \theta^* \rangle + \eta_t,
\) 
where $\theta^* \in \Theta^*$ is an unknown parameter vector and $\eta_t$ is a conditionally sub-Gaussian noise, i.e., $\forall \lambda \in \bbR, \E[e^{\lambda \eta_t} | a_1,\ldots a_t, \eta_1,\ldots,\eta_{t-1}] \leq \exp(\frac{\lambda^2}{2})$.

Further we assume that $\|a\|_2 = 1$ for all $a \in \cA, t \in [T]$.
\label{asmp:realizability_bandit-app}
\end{asmp}

\begin{definition}[\textbf{Regret}] The learner's goal is to minimize the regret for the selected actions $a_t, t\in[T]$, defined as
\begin{align*}
\reg(T) & = \E\Big[\sum_{t=1}^T \left( r(a^*) \!-\! r(a_t) \right)\Big] \nonumber \!=\! \sum_{t=1}^T \langle \theta^*, a^* \rangle \!-\! \langle \theta^*, a_t \rangle \;,
\end{align*}
where $a^* = {\argmax_{a_t \in \cA}} \; \langle \theta^*, a_t \rangle $, maximizes the expected reward in round $t$.
\end{definition}

\subsection{Sketched Linear UCB}
\begin{algorithm}[htbp!]
\caption{$\skLinUCB$\; (Sketched Linear UCB)}
\begin{algorithmic}[1]
\FOR {$t = 1, 2, ..., T$} 
\STATE Solve the least squares regression problem given by
\begin{align}
\label{eq:theta-hat-app}
    \hat{\theta}^{s}_{t} = \argmin_{\theta \in \R^b, \|\theta\|_{2} \leq 1} \sum_{i=1}^{t-1} \left( \langle \theta , \sketch_t a_i \rangle - r_{i}\right)^2 + \lambda \|\theta\|_2
\end{align}
\STATE Construct the Confidence Set $\cC^s_t$ according to \eqref{eq:confidence-app}.
\STATE Compute the the optimistic estimates of the parameter and the action:
\begin{align}
    (\tilde{\theta}^s_t, a^s_t) = \argmax_{\theta \in \cC_{t-1}^s, a \in \sketch_t \cA} \langle \theta, a \rangle
    \label{eq:theta_action_opt-app}
\end{align}
\STATE De-sketch the action $a_t = \sketch_t^\top a^{{s}}_t$
\STATE Play the action $a_t$ and observe the reward $r_t$.
\ENDFOR
\end{algorithmic}
\label{algo:sketch-LinUCB-app}
\end{algorithm}
We develop a sketched version of the popular algorithm LinUCB (Linear Upper Confidence Bound \citep{abbasi2011improved}) and is summarized in Algorithm~\ref{algo:sketch-LinUCB-app}. At every round $t$ the learner sketches the inputs using $\sketch_t \subseteq \R^{b \times d}$ whose entries are drawn i.i.d. from $N(0,1/b)$. It then solves a regularized least-squares problem (cf \eqref{eq:theta-hat-app} in Algorithm~\ref{algo:sketch-LinUCB-app}) in the $b$-dimensional sketched space to obtain an estimate \(\hat{\theta}_t^s\) of the unknown parameter. Subsequently, with $\bar{V}^{s}_{t} = \sum_{i=1}^{t} \sketch_t a_{i} (\sketch_t a_{i})^\top + \lambda I$, the learner constructs a confidence set \(\cC^s_t\) around the estimate:
\begin{align}
\label{eq:confidence-app}
\cC_t^s =  \Bigg\{ 
        \theta \in \bbr^b : \|\hat{\theta}^{s}_t - \sketch_t \theta^*\|_{\bar{V}^s_{t}} 
        \le
        \sqrt{%
            \log\!\Bigl(
                \frac{\det(\bar{V}^s_{t})^{1/2}\,\det(\lambda I)^{-1/2}}{\delta}
            \Bigr)}
        + \lambda^{1/2} \|\sketch_t \theta_*\|_2
      \Bigg\}.
\end{align}
The algorithm then uses this confidence set to compute an \emph{optimistic} parameter and action pair \(\bigl(\tilde{\theta}^s_t, a^s_t\bigr)\) (Line~4) where $\tilde{\theta}^s_t \in \cC_t^s$ and the action is in the \emph{sketched} action space $a \in \sketch_t \cA$. 
Once this sketched action is identified, it is ``\emph{de-sketched}'' (Line~5) to recover the corresponding action \(a_t\) in the original space. Finally, the chosen action \(a_t\) is played, and the observed reward \(r_t\) is used in subsequent rounds to refine future estimates. Our primary result in this section is the following decomposition for the regret.


\begin{tcolorbox}[
  width=\textwidth,
  colback=gray!5,        
  colframe=black,         
  arc=5pt,               
  boxrule=1pt,            
  left=5pt, right=5pt,  
  top=5pt,  bottom=2pt  
]
    \begin{restatable}[\textbf{Regret Decomposition for $\mathtt{Sk\mhyphen LinUCB}$}]{theorem}{ucb-app}
        Suppose Assumption~\ref{asmp:realizability_bandit-app} holds, the actions are selected according to Algorithm~\ref{algo:sketch-LinUCB-app}, and the de-sketched actions selected by Algorithm~\ref{algo:sketch-LinUCB-app} are in $\cA$. Then after some burn-in period $t_{\text{burn}}$, for some constant $C>0$, with probability $1 - \delta$
        \begin{align*}
            \reg(T)\! &\leq \!C\Bigg[\sqrt{b\log \left(T/\delta \right)} + \frac{1}{\sqrt{b}} \Big(\omega(\Theta_*) \!+\! \sqrt{\log(\frac{2}{\delta})}\Big)\Bigg] \\
        & \sqrt{Tb \log(1 + T\max_t\omega^2(\cA)/b^2)\log\log(2/\delta)}\\
    & \qquad \qquad + {\sum_{t=1}^{T} {\theta^*}^\top (I - \sketch_t^\top \sketch_t)a^*}
        \end{align*}
        where $\omega(\cM) = \E [\sup_{\theta \in \cM} \langle g, \theta \rangle]$, $g \sim N(0,I_d)$ is the Gaussian width of the set $\cM$.
        \label{thm:sk-LinUCB-app}
    \end{restatable}
\end{tcolorbox}
Note that term $I$ captures the regret in the sketched $b$ dimensional space while term $II$ captures the restricted isometry term due to random sketching. 

Before we move to the proof of Theorem~\ref{thm:sk-LinUCB-app}, we discuss what it entails. First note that we can bound term $II$ following the analysis in Section~\ref{sec:regression_app}, Section~\ref{subsec:RIP} and invoking Theorem~\ref{theo:sumTtheorem}, we get
\begin{align*}
        P\Bigg\{\sup_{\theta^* \in \Theta, a \in \cA}|\sum_{t=1}^T {\theta^*}^{\top} (\sketch_t^\top\sketch_t - I) a| &\geq \sqrt{T} ({\color{Red}\frac{1}{{b}} \omega(\Theta_*)\omega(\cA) + \frac{1}{\sqrt{b}}(\omega(\Theta_*) + \omega(\cA))}) + \epsilon \Bigg\}\\
        & \lesssim \exp \left(  - \min \left\{ \frac{\epsilon^2}{T\big({\color{Blue} \frac{1}{b} (\omega(\Theta_*) + \omega(\cA)) +  \frac{1}{\sqrt{b}}}\big)^2}, \frac{\epsilon}{\sqrt{T} {\color{Green}\frac{1}{b}}} \right\} \right)
    \end{align*}

Choose $\epsilon^2 = T({ \frac{1}{b} (\omega(\Theta_*) + \omega(\cA)) +  \frac{1}{\sqrt{b}}}\big)^2 \Bar{\epsilon}^2$, and take a union bound over both the events to get with probability $1 - \delta$
\begin{align*}
    \sup_{\theta^* \in \Theta, a \in \cA}|\sum_{t=1}^T {\theta^*}^{\top} (\sketch_t^\top\sketch_t - I) a| &\lesssim \sqrt{T}(\frac{1}{{b}} \omega(\Theta_*)\omega(\cA) + \frac{1}{\sqrt{b}}(\omega(\Theta_*) + \omega(\cA)) \\
    & \qquad\qquad + ({ \frac{1}{b} (\omega(\Theta_*) + \omega(\cA)) +  \frac{1}{\sqrt{b}}})\sqrt{\log (\delta^{-1})})
\end{align*}
Combining with Theorem~\ref{thm:sk-LinUCB-app} we have, with high probability
\begin{align*}
            \reg(T)\! &= \underbrace{\tilde{O} \Bigg(\sqrt{bT}\bigg[\sqrt{b} + \frac{1}{\sqrt{b}} \omega(\Theta_*)\bigg]\Bigg)}_{I} + \underbrace{\sqrt{T} \frac{1}{{b}} \omega(\Theta_*)\omega(\cA) + \frac{1}{\sqrt{b}}(\omega(\Theta_*) + \omega(\cA))}_{II}
        \end{align*}

Note that the above bound depends on the Gaussian widths of the parameter and action set instead of the ambient dimension $d$. In cases when the Gaussian widths are small (eg., sparse) one can choose the sketching dimension $b$ to obtain a tighter regret bound.

\emph{Proof of Theorem~\ref{thm:sk-LinUCB-app}} Recall that the regret is given by
\vspace{-1ex}
    \begin{gather*}
    \reg(T) = \sum_{t=1}^{T} \langle \theta^* , a^* \rangle - \langle \theta^* , a_t \rangle\\
    = \sum_{t=1}^{T}\langle \sketch_t \theta^* , \sketch_t a^* \rangle + \langle \theta^* , a^* \rangle - \langle \theta^* , a_t \rangle - \langle \sketch_t \theta^* , \sketch_t a^* \rangle
\end{gather*}
To be able to use the optimistic estimates $(\tilde{\theta}^s_t, a^s_t)$ computed in line 5 of Algorithm~\ref{algo:sketch-LinUCB-app}, we first show that the sketched unknown parameter belongs to the confidence set.
\begin{lemma}[\textbf{Confidence Ellipsoid}]
    Suppose $\cC_t^s$ be the confidence set defined as in \eqref{eq:confidence-app}. Then for all $t \in [T]$, with probability $1-\delta$, we have $\sketch_t \theta^* \in \cC_t^s$.
    \label{lemma:confidence-app}
\end{lemma}
Further note that $\sketch_t a^*_t \in \sketch_t \cA$ and therefore from the definition of the optimistic estimates, $\langle \sketch_t \theta^* , \sketch_t a^* \rangle \leq \langle \tilde{\theta}^s_t , a^s_t \rangle$. Using this we have
\begin{align*}
    \reg(T) 
    &\leq \sum_{t=1}^{T}\langle\tilde{\theta}^s_t , a^s_t \rangle + \langle \theta^* , a^* \rangle - \langle \theta^* , a_t \rangle - \langle \sketch_t \theta^* , \sketch_t a^* \rangle\\
    &\overset{(a)}{=} \sum_{t=1}^{T}\langle\tilde{\theta}^s_t , a^s_t \rangle - \langle \sketch_t\theta^* , a^s_t \rangle + \langle \sketch_t\theta^* , a^s_t \rangle  + \langle \theta^* , a^* \rangle - \langle \theta^* , a_t \rangle - \langle \sketch_t \theta^* , \sketch_t a^* \rangle\\  
    & {=} \sum_{t=1}^{T} \langle\tilde{\theta}^s_t - \sketch_t\theta^*, a^s_t \rangle + \langle \sketch_t\theta^* , a^s_t \rangle - \langle \theta^* , a_t \rangle  + \langle \theta^* , a^* \rangle  - \langle \sketch_t \theta^* , \sketch_t a^* \rangle \\
    & \overset{(b)}{=} \sum_{t=1}^{T} \langle\tilde{\theta}^s_t - \sketch_t\theta^*, a^s_t \rangle + \langle \sketch_t\theta^* , a^s_t \rangle - \langle \theta^* , \sketch_t^\top a^s_t \rangle + \langle \theta^* , a^* \rangle  - \langle \sketch_t \theta^* , \sketch_t a^* \rangle
\end{align*} 
where $(a)$ follows by adding and subtracting $\langle \sketch_t\theta^* , a^s_t \rangle$ and $(b)$ follows by noting that $a_t = \sketch_t^\top a_t^s$. Therefore
\begin{align*}
    \reg(T) &\leq \sum_{t=1}^{T} \langle\tilde{\theta}^s_t - \sketch_t\theta^*, a^s_t \rangle + \langle \sketch_t\theta^* , a^s_t \rangle - \langle \sketch_t \theta^* , a^s_t \rangle + \langle \theta^* , a^* \rangle  - \langle \sketch_t \theta^* , \sketch_t a^* \rangle\\
    &= \sum_{t=1}^{T} \langle\tilde{\theta}^s_t - \sketch_t\theta^*, a^s_t \rangle + {\theta^*}^\top (I - \sketch_t^\top \sketch_t)a^*\\
    &\leq \underbrace{\sum_{t=1}^{T} \big\|\tilde{\theta}^s_t \!-\! \sketch_t\theta^* \big\|_{\bar{V}^s_{t-1}} \|a^s_{t}\|_{(\bar{V}^s_{t-1})^{-1}}}_{(A)}  \!+\! \underbrace{\sum_{t=1}^{T} {\theta^*}^\top (I \!-\! \sketch_t^\top \sketch_t)a^*}_{(B)}
\end{align*}
where the last inequality follows by Cauchy Schwartz. Note that the inner products in term $(A)$ are $b$-dimensional. The price we paid for reducing the original $d$-dimensional inner product into this $b$-dimensional inner product is term $(B)$. Using the Confidence Ellipsoid in \eqref{eq:confidence-app} and elliptical potential lemma we bound term $(A)$ in the following lemma.
\begin{restatable}{lemma}{regretbdim}
    Suppose $(\tilde{\theta}^s_t, a^s_t)$ be as computed in \eqref{eq:theta_action_opt-app} and suppose $\lambda = 1$. Then for some constant $C>0$, with probability $1-\delta$
    \begin{align*}
        &\sum_{t=1}^{T} \big\|\tilde{\theta}^s_t - \sketch_t\theta^* \big\|_{\bar{V}^s_{t-1}} \|a^s_{t}\|_{(\bar{V}^s_{t-1})^{-1}} \\
        &\leq \cO\Bigg(\sqrt{b\log \left( \frac{1 + T}{\delta} \right)} + \frac{1}{\sqrt{b}} \Big(\omega(\Theta_*) + \sqrt{\log(\frac{2}{\delta})}\Big) \sqrt{Tb \log(1 + T\omega^2(\cA)/b^2)\log\log(2/\delta)}\Bigg)
    \end{align*}
    \vspace{-2ex}
    \label{lemma:b-dimensional-regret}
\end{restatable}
Finally using Lemma~\ref{lemma:b-dimensional-regret} we have for some constant $C>0$ with probability $1-\delta$
\begin{align*}
    \reg(T) &\leq \cO\Bigg(\sqrt{b\log \left( \frac{1 + T}{\delta} \right)} + \frac{1}{\sqrt{b}} \Big(\omega(\Theta_*) + \sqrt{\log(\frac{2}{\delta})}\Big) \\
        & \sqrt{Tb \log(1 + T\max_t\omega^2(\cA)/b^2)\log\log(2/\delta)}\Bigg)\\
    & \qquad \qquad + {\sum_{t=1}^{T} {\theta^*}^\top (I - \sketch_t^\top \sketch_t)a^*} \qquad\qquad\quad \qquad\quad\qquad\quad\qquad\quad \qed
\end{align*} 

\subsubsection{Proof of Auxiliary Lemmas}
\begin{proof}[\textbf{Proof of Lemma~\ref{lemma:confidence-app}}]
    Suppose $\displaystyle \bar{V}^{s}_{t} = \sum_{i=1}^{t-1} \sketch_t a_{i} {(\sketch_t a_{i})}^\top + \lambda I$. Then we can express $\hat{\theta}^{s}_t$ as follows:
    \begin{align*}
        \hat{\theta}^{s}_ t &= (\bar{V}^{s}_{t})^{-1} \Big( \sum_{i=1}^{t-1} (\sketch_t a_{i}) r_i \Big)\\
        &\overset{(a)}{=} (\bar{V}^{s}_{t})^{-1} \Big( \sum_{i=1}^{t-1} (\sketch_t a_{i}) \langle \theta^*, \sketch_t^\top (\sketch_t a_i) \rangle + \eta_i + \varepsilon_i \Big)
    \end{align*}
    where in $(a)$, $\varepsilon_i = \langle\theta^*,a_i - \sketch_t^\top \sketch_t a_i \rangle$ and we have used the fact that $r_i = \langle \theta^* , a_i \rangle + \eta_i$. Therefore
    \begin{align*}
        \hat{\theta}^{s}_t &= (\bar{V}^{s}_{t})^{-1} \Big( \sum_{i=1}^{t-1} (\sketch_t a_{i}) {(\sketch_t a_{i})}^\top \sketch_t \theta^* \Big) + (\bar{V}^{s}_{t})^{-1} \sum_{i=1}^{t-1} (\sketch_t a_{i}) \eta_i + (\bar{V}^{s}_{t})^{-1} \sum_{i=1}^{t-1} (\sketch_t a_{i}) \varepsilon_i \\
        &= (\bar{V}^{s}_{t})^{-1} \Big( \underbrace{\sum_{i=1}^{t-1} (\sketch_t a_{i}) {(\sketch_t a_{i})}^\top + \lambda I}_{\bar{V}^{s}_{t}}  \Big)\sketch_t \theta^*  - \lambda (\bar{V}^{s}_{t})^{-1} \sketch_t \theta^* + (\bar{V}^{s}_{t})^{-1} \sum_{i=1}^{t-1} (\sketch_t a_{i}) \eta_i + (\bar{V}^{s}_{t})^{-1} \sum_{i=1}^{t-1} (\sketch_t a_{i}) \eta_i\\
        &= \sketch_t \theta^* + (\bar{V}^{s}_{t} )^{-1} \sum_{i=1}^{t-1} (\sketch_t a_{i}) \eta_i - \lambda (\bar{V}^{s}_{t})^{-1} \sketch_t \theta^*+ (\bar{V}^{s}_{t})^{-1} \sum_{i=1}^{t-1} (\sketch_t a_{i}) \eta_i
    \end{align*}
Therefore 
\begin{align*}
    \|\hat{\theta}^{s}_t - \sketch_t \theta^*\| & \leq \underbrace{\Big\|\sum_{i=1}^{t-1} a^s_{i} \eta_i \Big\|_{(\bar{V}^{s}_{t})^{-1}}}_{I} + \underbrace{\Big\|\sum_{i=1}^{t-1} a^s_{i} \varepsilon_i \Big\|_{(\bar{V}^{s}_{t})^{-1}}}_{II} + \underbrace{\lambda^{1/2} \|\sketch_t \theta^*\|}_{III}\\
\end{align*}
Consider term $I$ first. Using the Self-Normalized Bound for Vector-Valued Martingales (Theorem 1, \cite{abbasi2011improved}) we have, with probability $1-\delta$
\begin{align*}
    \Big\|\sum_{i=1}^{t-1} a^s_{i} \eta_i \Big\|_{(\bar{V}^{s}_{t})^{-1}} \leq 2 \sqrt{\log \left(\frac{\det(\bar{V}^{s}_{t})^{1/2}\det(\lambda I_b)^{-1/2}}{\delta} \right)}
\end{align*}
Now using the fact that $\|a\| \leq 1$ and $\lambda = 1$ we have
\begin{align*}
    \Big\|\sum_{i=1}^{t-1} a^s_{i} \eta_i \Big\|_{(\bar{V}^{s}_{t})^{-1}} \leq 2 \sqrt{b\log \left( \frac{1 + t}{\delta} \right)}
\end{align*}

Next we handle term $II$. Note that $\varepsilon_i$ is not sub-Gaussian and therefore we cannot use the above. It rather follows a Bernstein type condition with the moment generating function given by: for every $\lambda$ with $|\lambda| < 1/b_{\text{Bern}}$,
\begin{align*}
\mathbb{E} \exp\{ \lambda (X - \mu) \} 
\le 
\exp\left( 
  \frac{\lambda^2 \sigma^2}{2} \cdot \frac{1}{1 - b_{\text{Bern}} |\lambda|} 
\right) .
\end{align*}
where $b_{\text{Bern}} = \Theta(\frac{1}{\sqrt{b}})$ and $\sigma^2 = \Theta(1/b)$. We use a recent result \citep{ziemann2025vector} that provides self-normalized bounds for such Bernstein type processes. We invoke Theorem~1 from \cite{ziemann2025vector} to conclude that

\[
\alpha \triangleq \left( \frac{ \sqrt{e(1+\nu)} \, \|S_\tau\|_{(V_\tau + \Gamma)^{-1} V (V_\tau + \Gamma)^{-1}} }{ \nu \sqrt{d+2} } - 1 \right) \vee 0.
\]

Then as long as ${V}_\tau + \Gamma \succeq e(1+\nu)^2 V \succeq (1+\nu)^2 e^{-1} (d+2) B_W^2 B_X^2$, we have that with probability at least $1 - \delta$:
\[
\|S_\tau\|^2_{(V_\tau + \Gamma)^{-1}} \leq 
\left( \frac{(1 + \alpha)^2}{1 + 2\alpha} \times \frac{1}{1 - \epsilon} \right)
\times \sigma^2 \times 
\left[ \log\left( \frac{ \det(V_\tau + \Gamma) }{ \det(V) } \right) + 2\log\left( \frac{1}{\delta} \right) \right].
\]
where ${V}_\tau = \bar{V}^s_\tau,V = \lambda I $ and $S_{\tau} = \Big\|\sum_{i=1}^{\tau-1} a^s_{i} \epsilon_i \Big\|_{(\bar{V}^{s}_{t})^{-1}}$.

We can set $\nu = \epsilon = 1/2$ and after a burn-in period $t_{\text{burn}}$ such that $\bar{V}^s_\tau + \Gamma \succeq (1+\nu)^2 e^{-1} (d+2) B_W^2 B_X^2$ and $ \sqrt{e(1+\nu)} \, \|S_\tau\|_{(V_\tau + \Gamma)^{-1} V (V_\tau + \Gamma)^{-1}} \leq \nu \sqrt{d+2}$ we have that with probability at least $1 - \delta$:
\[
\|S_\tau\|^2_{(V_\tau + \Gamma)^{-1}} \lesssim 
 \sigma^2 \times 
\left[ \log\left( \frac{ \det(V_\tau + \Gamma) }{ \det(V) } \right) + 2\log\left( \frac{1}{\delta} \right) \right]. \lesssim
     \sqrt{b\log \left( \frac{1 + t}{\delta} \right)}
\]

Finally, to control term $II$, since $\sketch_t \sim N(0,1/b)$ we have with probability $1 - 2 \exp(c\omega^2(\Theta_*) - p^2)$ for some $c>0$, (see \cite{bcfs14}, Theorem 5)
\begin{align*}
    \|\sketch_t \theta_*\| \leq 1 + \frac{\omega(\Theta_*) + p}{\sqrt{b}}
\end{align*}
Therefore with probability $(1-\delta)$, for some constant $C>0$ 
\begin{align*}
    \|\sketch_t \theta_*\| \leq 1 + \frac{C}{\sqrt{b}} (\omega(\Theta_*) + \sqrt{\log(2/\delta)})
\end{align*}

Combining all these it is immediate that $\sketch_t \theta^* \in \cC^s_t$. Further, plugging all the bounds back we get, with probability $(1-\delta)$, for some constant $C>0$
\begin{align*}
    \|\hat{\theta}^{s}_t - \sketch_t \theta^*\| & \leq 2 \sqrt{b\log \left( \frac{1 + t}{\delta} \right)} + 1 + \frac{C}{\sqrt{b}} (\omega(\Theta_*) + \sqrt{\log(2/\delta)})
\end{align*}
\end{proof}

\regretbdim*
\begin{proof}

    Note that $\tilde{\theta}^s_t \in \cC_t^s = \Bigg\{ \theta \in \bbr^b : \|\hat{\theta}^{s}_t - \sketch_t \theta^*\|_{\bar{V}^s_{t-1}}  \leq \cO\Bigg(\sqrt{b\log \left( \frac{1 + t}{\delta} \right)} + \frac{1}{\sqrt{b}} \Big(\omega(\Theta_*) + \sqrt{\log(\frac{2}{\delta})}\Big)\Bigg)\Bigg\}$
    which along with $t \leq T$ immediately implies that with probability $1-\delta$
    \begin{align*}
        \sum_{t=1}^{T}& \big\|\tilde{\theta}^s_t - \sketch_t\theta^* \big\|_{\bar{V}^s_{t-1}} \|a^s_{t}\|_{(\bar{V}^s_{t-1})^{-1}} \leq \cO\Bigg(\sqrt{b\log \left( \frac{1 + T}{\delta} \right)} + \frac{1}{\sqrt{b}} \Big(\omega(\Theta_*) + \sqrt{\log(\frac{2}{\delta})}\Big)\Bigg)\sum_{t=1}^{T} \|a^s_{t}\|_{(\bar{V}^s_{t-1})^{-1}}
    \end{align*}
    Using Cauchy Schwarz
\begin{align*}
    \sum_{t=1}^{T}& \big\|\tilde{\theta}^s_t - \sketch_t\theta^* \big\|_{\bar{V}^s_{t-1}} \|a^s_{t}\|_{(\bar{V}^s_{t-1})^{-1}} \leq \cO\Bigg(\sqrt{b\log \left( \frac{1 + T}{\delta} \right)} + \frac{1}{\sqrt{b}} \Big(\omega(\Theta_*) + \sqrt{\log(\frac{2}{\delta})}\Big)\Bigg)\\& \sqrt{T\sum_{t=1}^{T} \min\left(\|a^s_{t}\|^2_{(\bar{V}^s_{t-1})^{-1}},1\right)}
\end{align*}
    Using Lemma~11 from \cite{abbasi2011improved} we have for $\|a^s_{t}\| \leq L$,
    \begin{align*}
        \sum_{t=1}^{T} \min(\|a^s_{t}\|_{(\bar{V}^s_{t-1})^{-1}},1) \leq b\log\left(1+ \frac{TL^2}{b\lambda}\right)
    \end{align*}
Now since for any $a \in \cA$, $\|a\| \leq 1,$ using the same proof technique to bound $\|\sketch_t \theta_*\|$ in \cref{lemma:confidence-app} we have that with probability $1-\delta$, $\|a^s_t\| \leq 1 + C\frac{\omega(\cA) + \sqrt{\log(2/\delta)}}{\sqrt{b}}$, for some $C>0$. Taking a union bound for all $t \in [T]$ and with the event that $\tilde{\theta}^s_t \in \cC_t^s$ we have, with probability $1-\delta$
\begin{align*}
    \sum_{t=1}^{T}& \big\|\tilde{\theta}^s_t - \sketch_t\theta^* \big\|_{\bar{V}^s_{t-1}} \|a^s_{t}\|_{(\bar{V}^s_{t-1})^{-1}}\\
    &\leq \cO\Bigg(\sqrt{b\log \left( \frac{1 + T}{\delta} \right)} + \frac{1}{\sqrt{b}} \Big(\omega(\Theta_*) + \sqrt{\log(\frac{2}{\delta})}\Big) \sqrt{Tb \log(1 + T\omega^2(\cA)/b^2)\log\log(2/\delta)}\Bigg)
\end{align*}
\end{proof}

\subsection{Sketched Thompson Sampling}
Next we show that a similar decomposition can also be obtained for Thompson Sampling algorithm (\citep{agrawal2013thompson,TS12}).
We consider a finite action set as in \cite{agrawal2013thompson}, i.e., $\cA = \{\x_{t,1},\x_{t,2},\ldots, \x_{t,K}\}$, where $K$ is the number of arms. The learner chooses an action $i_t \in [K]$ and observes the reward $\langle \x_{t,i_t},\theta^* \rangle + \eta_t$, i.e., $a_t = \x_{t,i_t}$ in Assumption~\ref{asmp:realizability_bandit-app}.

We take a Bayesian approach by placing a Gaussian prior
\(
  \bm{\theta^s_t} \;\sim\; \mathcal{N}(\bm{\mu}_0^s, \bm{\Sigma}_0^s),
\)
where \(\bm{\mu}_0^s = \bm{0} \in \mathbb{R}^b\) and \(\bm{\Sigma}_0^s = \bm{I}_{b \times b} \in \mathbb{R}^{b \times b}\) are the prior mean and covariance matrix, respectively. Note that the mean vector and covariance matrix are in the $b$-dimensional space. After \(t\) observations, the posterior distribution of \({\theta^s_t}\) is given by
\(
  {\theta^s_t} \;\big|\; \{{\x}_{\tau,i_{\tau}}, r(\x_{\tau,i_{\tau}})\}_{\tau=1}^t 
  \;\sim\; \mathcal{N}(\bm{\mu}_{t-1}^s,\; v^2 (\bm{\Sigma^{s}}_{t-1})^{-1}),
\)
where the posterior mean \(\bm{\mu}_t^s\) and covariance \(\bm{\Sigma}_t^s\) are given by standard Bayesian linear regression update equations:
\begin{align}
    \bm{\Sigma^s}_t &= {\left(\bm{\Sigma^s}_0 +\sum_{\tau=1}^{t-1} \x^s_{\tau,a_{\tau}}\x_{\tau,a_{\tau}}^{s \top}\right)} \nonumber \\
    \bm{\mu}_t^s &= (\bm{\Sigma}_t^s)^{-1} \left( \sum_{\tau=1}^{t-1} \x^s_{\tau,i_{\tau}} r(\x_{\tau,i_{\tau}})\right)
    \label{eq:ts_update_eqs-app}
\end{align}
where $\x^s_{t,i} = \sketch_t \x_{t,i}$ is the sketched context vector.

\begin{algorithm}[t]
\caption{$\skLinTS$\; (Sketched Linear TS)}
\label{alg:sk-LinTS}
\begin{algorithmic}[1]
\STATE \textbf{Input:} variance parameter $v = \sqrt{9 b \log(t/\delta)}$
\FOR{$t=1,2,\ldots$}
    \STATE Sample $\theta^s_t$ from $\mathcal{N}(\bm{\mu}_{t}^s,\; v^2 (\bm{\Sigma^{s}}_{t})^{-1})$.
    \STATE Play arm $i_t \;:=\; \arg\max_{i \in [K]}\,\x_{t,i_{t}}^\top \theta^s_t$, and observe reward $r_t = r_{ \x_{t,i_{t}} }(t)$.
    \STATE Update $\bm{\Sigma^s}_t$ and $\bm{\mu}_t^s$ according to $\eqref{eq:ts_update_eqs-app}$.
\ENDFOR
\end{algorithmic}
\end{algorithm}
The regret decomposition for $\skLinTS$ is given by the following Theorem.
\begin{tcolorbox}[
  width=\textwidth,
  colback=gray!5,        
  colframe=black,         
  arc=5pt,               
  boxrule=1pt,            
  left=2pt, right=2pt,  
  top=5pt,  bottom=5pt  
]
    \begin{restatable}[\textbf{Regret Decomposition for $\mathtt{Sk\mhyphen LinTS}$}]{theorem}{ts-app}
        Suppose Assumption~\ref{asmp:realizability_bandit-app} holds and the actions are selected according to Algorithm~\ref{algo:sketch-LinUCB-app}. Then for some constant $C>0$, with probability $1 - \delta$
        \begin{align*}
            \reg(T) &\leq C (6\sqrt{b\log(KT)\log(T/\delta)} + \sqrt{b \log(T^3/\delta)})\\&\underbrace{\qquad\qquad (\sqrt{bT\log T} + Z \sqrt{T \log(1/\delta)}) \qquad}_{I}\\
            & \qquad + \underbrace{2 \sup_{\theta \in \Theta_*}  \sup_{a_t \in \cA, t \in [T]} \; \sum_{t=1}^{T} \theta^\top (S^\top S - I) a_t}_{II},\\
            &\text{where\;} Z = \tilde{\cO}\left(\frac{1}{b}\omega(\Theta_*)\right),
        \end{align*}
        and $\omega(\cM) = \E [\sup_{\theta \in \cM} \langle g, \theta \rangle]$, $g \sim N(0,I_d)$ is the Gaussian width of the set $\cM$.
        \label{thm:sk-LinTS}
    \end{restatable}
\end{tcolorbox}
As in the case of $\skLinUCB$\; regret decomposition, term $I$ captures the regret in the sketched $b$ dimensional space while term $II$ captures the restricted isometry term.

\begin{proof}
    The regret is given by
\begin{align*}
    \reg(T) &= \langle \theta^*, \x_{t,a^*} \rangle - \langle \theta^*, \x_{t,a_t} \rangle = \langle \theta^*, \x_{t,a^*} \rangle - \langle \theta^*, \x_{t,a_t} \rangle \\
    &+ \langle \sketch_t\theta^*, \x^s_{t,a^*} \rangle - \langle \sketch_t\theta^*, \x^s_{t,a_t} \rangle - \langle \sketch_t\theta^*, \x^s_{t,a^*} \rangle + \langle \sketch_t\theta^*, \x^s_{t,a_t} \rangle \\
    & = \langle \theta^*, \x_{t,a^*} \rangle - \langle \theta^*, \x_{t,a_t} \rangle + \langle \sketch_t\theta^*, \sketch_t\x_{t,a^*} \rangle - \langle \sketch_t\theta^*, \sketch_t\x_{t,a_t} \rangle - \langle \sketch_t\theta^*, \sketch_t\x_{t,a^*} \rangle + \langle \sketch_t\theta^*, \sketch_t\x_{t,a_t} \rangle\\
    & = \langle \sketch_t\theta^*, \x^s_{t,a^*} \rangle - \langle \sketch_t\theta^*, \x^s_{t,a_t} \rangle + 2 \sup_{\theta \in \Theta_*}  \sup_{a_t \in \cA, t \in [T]} \; \sum_{t=1}^{T} \theta^\top (S^\top S - I) a_t
\end{align*}
Suppose $\tilde{a}_t = \arg \min_{i \in [K]} \sqrt{{\x^s_{t,i}}^\top{\bm{\Sigma^s}_t}^{-1}\x^s_{t,i}}$. Then

\begin{align*}
    \langle \sketch_t\theta^*, \x_{t,a^*} \rangle - \langle \sketch_t\theta^*, \x_{t,a_t} \rangle & = \langle \sketch_t\theta^*, \x^s_{t,a^*} \rangle - \langle \sketch_t\theta^*, \x^s_{t,a_t} \rangle
    + \langle \sketch_t\theta^*, \x^s_{t,\tilde{a}_t} \rangle - \langle \sketch_t\theta^*, \x^s_{t,\tilde{a}_t} \rangle \\
    & \qquad + 2 \sup_{\theta \in \Theta_*}  \sup_{a_t \in \cA, t \in [T]} \; \sum_{t=1}^{T} \theta^\top (S^\top S - I) a_t 
\end{align*}
\begin{align}
    &= \underbrace{\langle \sketch_t\theta^*, \x^s_{t,a^*} \rangle - \langle \sketch_t\theta^*, \x^s_{t,\tilde{a}_t} \rangle + \langle \sketch_t\theta^*, \x^s_{t,\tilde{a}_t} \rangle - \langle \sketch_t\theta^*, \x^s_{t,a_t} \rangle}_{I} \nonumber\\
    & \qquad + 2 \sup_{\theta \in \Theta_*}  \sup_{a_t \in \cA, t \in [T]} \; \sum_{t=1}^{T} \theta^\top (S^\top S - I) a_t
    \label{eq:ts_regret_decomp}
\end{align}
We define the following events 
\begin{align}
    E_1 = \;\; &=\;\;\Bigl\{\,\forall i : \bigl|\;{\x^s_{t,i}}^\top {\hat{\theta}}_t - {\x^s_{t,i}}^\top \sketch_t \theta_*\;\bigr|\;\le\;(\sqrt{b\ln(t^3/\delta)} + 1 )\sqrt{\;{\x^s_{t,i}}^\top{\bm{\Sigma^s}_t}^{-1}\x^s_{t,i}}\;\Bigr\}
    \label{eq:E1}\\
    E_2 \;\; &=\;\;\Bigl\{\,\forall i : \bigl|\;\zeta_{t,i} - {\x^s_{t,i}}^\top {\hat{\theta}}_t\;\bigr|\;\le\;6\sqrt{b\log(Kt) \ln(2t/\delta)\;{\x^s_{t,i}}^\top{\bm{\Sigma^s}_t}^{-1}\x^s_{t,i}}\;\Bigr\}
    \label{eq:E2}
\end{align}
We bound term $I$ in the following lemma.
\begin{lemma}
    Suppose $E_1$ and $E_2$ be as defined in \cref{eq:E1} and \cref{eq:E2} and let $g_t = 6\sqrt{b\log(Kt)\log(t/\delta)} + \sqrt{b \log(t^3/\delta)} + 1$ and $p = \frac{1}{4e\sqrt{\pi}}$. Then for any filtration $\cF_{t-1}$ such that $E_1$ is true, we have
    \begin{align*}
        \E[\langle \sketch_t\theta^*, \x^s_{t,a^*} \rangle - \langle \sketch_t\theta^*, \x^s_{t,\tilde{a}_t} \rangle &+ \langle \sketch_t\theta^*, \x^s_{t,\tilde{a}_t} \rangle - \langle \sketch_t\theta^*, \x^s_{t,a_t} \rangle | \cF_{t-1}] \leq \frac{3}{p} g_t \left(\E[\sqrt{{\x^s_{t,i}}^\top{\bm{\Sigma^s}_t}^{-1}\x^s_{t,i}}|\cF_t] + \frac{1}{t^2}\right)
    \end{align*}
    \label{lemma:ts-lemma-1}
\end{lemma}
\begin{proof}
Note that
\begin{align*}
    \langle \sketch_t\theta^*, \x^s_{t,a^*} \rangle - \langle \sketch_t\theta^*, \x^s_{t,\tilde{a}_t} \rangle + \langle \sketch_t\theta^*, \x^s_{t,\tilde{a}_t} \rangle - \langle \sketch_t\theta^*, \x^s_{t,a_t} \rangle \\
    \leq 2 g_t \sqrt{{\x^s_{t,i}}^\top{\bm{\Sigma^s}_t}^{-1}\x^s_{t,i}} + g_t \sqrt{{\x^s_{t,\tilde{a}_t}}^\top{\bm{\Sigma^s}_t}^{-1}\x^s_{t,{a}_t}}
\end{align*}
follows when events $E_1$ and $E_2$ are assumed to be true. Therefore
\begin{align*}
    \E[\langle \sketch_t\theta^*, \x^s_{t,a^*} \rangle - \langle \sketch_t\theta^*, \x^s_{t,\tilde{a}_t} \rangle &+ \langle \sketch_t\theta^*, \x^s_{t,\tilde{a}_t} \rangle - \langle \sketch_t\theta^*, \x^s_{t,a_t} \rangle] \\
    &\leq \E[2 g_t \sqrt{{\x^s_{t,i}}^\top{\bm{\Sigma^s}_t}^{-1}\x^s_{t,i}} + g_t \sqrt{{\x^s_{t,\tilde{a}_t}}^\top{\bm{\Sigma^s}_t}^{-1}\x^s_{t,{a}_t}}|\cF_t] + P(E_2^{C})\\
    &\leq \frac{3}{p} g_t \E\left[\sqrt{{\x^s_{t,\tilde{a}_t}}^\top{\bm{\Sigma^s}_t}^{-1}\x^s_{t,{a}_t}}|\cF_t \right] + \frac{2g_t}{pt^2}
\end{align*}
where we have used Lemma 1 and 4 from \cite{agrawal2013thompson}.
\end{proof}
Now note that using Lemma~\ref{lemma:ts-lemma-1}, we have $Y_s = \sum_{t=1}^s(\langle \sketch_t\theta^*, \x^s_{t,a^*} \rangle - \langle \sketch_t\theta^*, \x^s_{t,\tilde{a}_t} \rangle + \langle \sketch_t\theta^*, \x^s_{t,\tilde{a}_t} \rangle - \langle \sketch_t\theta^*, \x^s_{t,a_t} \rangle)\1\{E_1\} - \frac{3}{p}g_t\left(\sqrt{{\x^s_{t,i}}^\top{\bm{\Sigma^s}_t}^{-1}\x^s_{t,i}} + \frac{1}{t^2}\right)$ is super-martingale with respect to the filtration $\cF_{t} = \sigma\{(r_i,a_i), 1\leq i \leq t-1, a_t\}$. The following lemma provides a bound on the absolute value of super-martingale process for every $t$ with high probability.

\begin{lemma}
    Suppose $X_t = (\langle \sketch_t\theta^*, \x^s_{t,a^*} \rangle - \langle \sketch_t\theta^*, \x^s_{t,\tilde{a}_t} \rangle + \langle \sketch_t\theta^*, \x^s_{t,\tilde{a}_t} \rangle - \langle \sketch_t\theta^*, \x^s_{t,a_t} \rangle)\1\{E_1\} - \frac{3}{p} g_t\left(\sqrt{{\x^s_{t,i}}^\top{\bm{\Sigma^s}_t}^{-1}\x^s_{t,i}} + \frac{1}{t^2}\right)$,  where $g_t = 6 \sqrt{b\log(tK)\ln(t/\delta)} + \sqrt{b\ln(t^3/\delta)} + 1$. Then with probability $(1-\delta)$ for some constant $C>0$
\begin{align*}
    |X_t| \leq C \Bigg(\frac{1}{b}\big(\omega(\Theta_*) + \sqrt{\log(2/\delta)}\big)(\sqrt{\log(KT)} + \sqrt{\log(2/\delta)}) + \frac{3}{p}g_t\frac{\sqrt{\log(KT)} + \sqrt{\log(2/\delta)}}{\sqrt{b}}
 \Bigg)
\end{align*}
\label{lemma:mar_bound}
\end{lemma}
\begin{proof}
Note that 
    \begin{align*}
        X_t &=(\langle \sketch_t\theta^*, \x^s_{t,a^*} \rangle - \langle \sketch_t\theta^*, \x^s_{t,\tilde{a}_t} \rangle + \langle \sketch_t\theta^*, \x^s_{t,\tilde{a}_t} \rangle - \langle \sketch_t\theta^*, \x^s_{t,a_t} \rangle)\1\{E_1\} - \frac{3}{p} g_t\left(\sqrt{{\x^s_{t,i}}^\top{\bm{\Sigma^s}_t}^{-1}\x^s_{t,i}} + \frac{1}{t^2}\right)\\
        &= (\langle \sketch_t\theta^*, \x^s_{t,a^*} \rangle - \langle \sketch_t\theta^*, \x^s_{t,a_t} \rangle)\1\{E_1\} - \frac{3}{p} g_t\left(\sqrt{{\x^s_{t,i}}^\top{\bm{\Sigma^s}_t}^{-1}\x^s_{t,i}} + \frac{1}{t^2}\right)
    \end{align*}
Now consider $\sqrt{{\x^s_{t,i}}^\top{\bm{\Sigma^s}_t}^{-1}\x^s_{t,i}}$. We know that $\bm{\Sigma^s}_t \succeq \lambda I_b$ implying ${\lambda_{\min}(\bm{\Sigma^s}_t}^{-1}) \leq \lambda$. Therefore
\begin{align*}
    \sqrt{{\x^s_{t,i}}^\top{\bm{\Sigma^s}_t}^{-1}\x^s_{t,i}} \leq \sqrt{\lambda}\|\x^s_{t,i}\|_2 \leq \|\x^s_{t,i}\|_2
\end{align*}
Since $\sketch_t \sim N(0,1/b)$ we have with probability $1 - 2 \exp(c\omega^2(\cA) - q^2)$ for some $c>0$, (see \cite{bcfs14}, Theorem 5)
\begin{align*}
    \|\sketch_t \x_{t,a}\| \leq 1 + \frac{\omega(\cA) + q}{\sqrt{b}}
\end{align*}
where $\cA = \{\x_{t,k}, t\in[T], k\in[K]\}$. Therefore with probability $(1-\delta)$, for all$t \in [T], k\in [K]$, for some constant $C>0$ 
\begin{align*}
    \|\x^s_{t,k}\| &\leq 1 + \frac{C}{\sqrt{b}} (\omega(\cA) + \sqrt{\log(2/\delta)})\\
    &\leq 1 + \frac{C}{\sqrt{b}} (\sqrt{\log(KT)} + \sqrt{\log(2/\delta)})
\end{align*}
Next, to control $|\langle \sketch_t\theta^*, \x^s_{t,a^*} \rangle - \langle \sketch_t\theta^*, \x^s_{t,a_t} \rangle|$, note that by Cauchy Schwarz 
\begin{align*}
    |\langle \sketch_t\theta^*, \x^s_{t,a^*} \rangle - \langle \sketch_t\theta^*, \x^s_{t,a_t} \rangle| \leq 2 \|\sketch_t\theta^*\| \|\x^s_{t,a_t}\|
\end{align*}
Using the same technique as above, we have with probability $(1-\delta)$
\begin{align*}
    \|\sketch_t\theta^*\| \leq 1 + \frac{C}{\sqrt{b}} (\omega(\Theta_*) + \sqrt{\log(2/\delta)})
\end{align*}
Taking a union bound over all the events we have, with probability $(1-\delta)$ for some constant $C>0$
\begin{align*}
    |X_t| \leq C \Bigg(\frac{1}{b}\big(\omega(\Theta_*) + \sqrt{\log(2/\delta)}\big)(\sqrt{\log(KT)} + \sqrt{\log(2/\delta)}) + \frac{3}{p}g_t\frac{\sqrt{\log(KT)} + \sqrt{\log(2/\delta)}}{\sqrt{b}}
 \Bigg)
\end{align*}
\end{proof}

Therefore taking a union bound over the event in \cref{lemma:mar_bound} and using using Azuma-Hoeffding inequality we have with probability $1-\delta$ we have
\begin{align*}
    \sum_{t=1}^{T} & \langle \sketch_t\theta^*, \x^s_{t,a^*} \rangle - \langle \sketch_t\theta^*, \x^s_{t,\tilde{a}_t} \rangle + \langle \sketch_t\theta^*, \x^s_{t,\tilde{a}_t} \rangle - \langle \sketch_t\theta^*, \x^s_{t,a_t} \rangle \\
    & \leq \frac{3}{p}g_T  \sum_{t=1}^T\left(\sqrt{{\x^s_{t,i}}^\top{\bm{\Sigma^s}_t}^{-1}\x^s_{t,i}} + \frac{1}{t^2}\right) + C\sqrt{2 \sum_{t} \frac{9}{p^2}g_t^2 Z^2\log(4/\delta)}\\
    &\leq \frac{3}{p}g_T 5\sqrt{b T \ln T}+ \frac{6}{p}g_T C\sqrt{\sum_{t} Z^2\log(1/\delta)}
\end{align*}
where $Z =\frac{1}{b}\big(\omega(\Theta_*) + \sqrt{\log(4/\delta)}\big)(\sqrt{\log(KT)} + \sqrt{\log(2/\delta)}) + \frac{3}{p}\frac{\sqrt{\log(KT)} + \sqrt{\log(2/\delta)}}{\sqrt{b}}$ and the last step follows from Lemma 11 of \cite{chu2011contextual}. Therefore with probability $1-\delta$ we have
\begin{align*}
    \sum_{t=1}^{T} & \langle \sketch_t\theta^*, \x^s_{t,a^*} \rangle - \langle \sketch_t\theta^*, \x^s_{t,\tilde{a}_t} \rangle + \langle \sketch_t\theta^*, \x^s_{t,\tilde{a}_t} \rangle - \langle \sketch_t\theta^*, \x^s_{t,a_t} \rangle \\
    & \leq C g_T (\sqrt{bT\log T} + Z \sqrt{T \log(1/\delta)})\\
    & \leq C (6\sqrt{b\log(KT)\log(T/\delta)} + \sqrt{b \log(T^3/\delta)} + 1)(\sqrt{bT\log T} + Z \sqrt{T \log(1/\delta)})
\end{align*}
Combining with \cref{eq:ts_regret_decomp} we get with probability $1-\delta$

\begin{align*}
    \reg(T) &\leq C (6\sqrt{b\log(KT)\log(T/\delta)} + \sqrt{b \log(T^3/\delta)} + 1)(\sqrt{bT\log T} + Z \sqrt{T \log(1/\delta)})\\
    & \qquad + 2 \sup_{\theta \in \Theta_*}  \sup_{a_t \in \cA, t \in [T]} \; \sum_{t=1}^{T} \theta^\top (S^\top S - I) a_t,
\end{align*}
completing the proof.
\end{proof}
\subsection{Experiments}

In this section, we evaluate the performance of \( \skLinUCB \) and \( \skLinTS \) in comparison to their unsketched counterparts, LinUCB and LinTS, on synthetic datasets. We consider a contextual bandit setting with a context dimension of \( d = 500 \) and \( K = 4 \) possible actions. The number of rounds is set to \( T = 10,000 \).

\paragraph{Dataset:} To illustrate the effects of sketching, we construct a dataset with low metric entropy  through truncation. The context vectors \( a_t \in \mathcal{A}_t \) and the unknown parameter \( \theta_{*} \) are sampled uniformly from the unit ball \( \cB(0, 1) \). We investigate three scenarios: (a) only the context vectors are sparse, (b) only the unknown parameter \( \theta_{*} \) is sparse, and (c) both the context vectors and the unknown parameter \( \theta_{*} \) are sparse. For sparsity, we apply truncation to both the context vectors and \( \theta_{*} \). Specifically, for a vector of dimension \( d = 500 \), we retain only the first \( s = 50 \) coordinates (0.9 sparsity). 

\begin{figure*}[!htbp]
    \centering
    \subfigure{\includegraphics[width=0.32\textwidth]{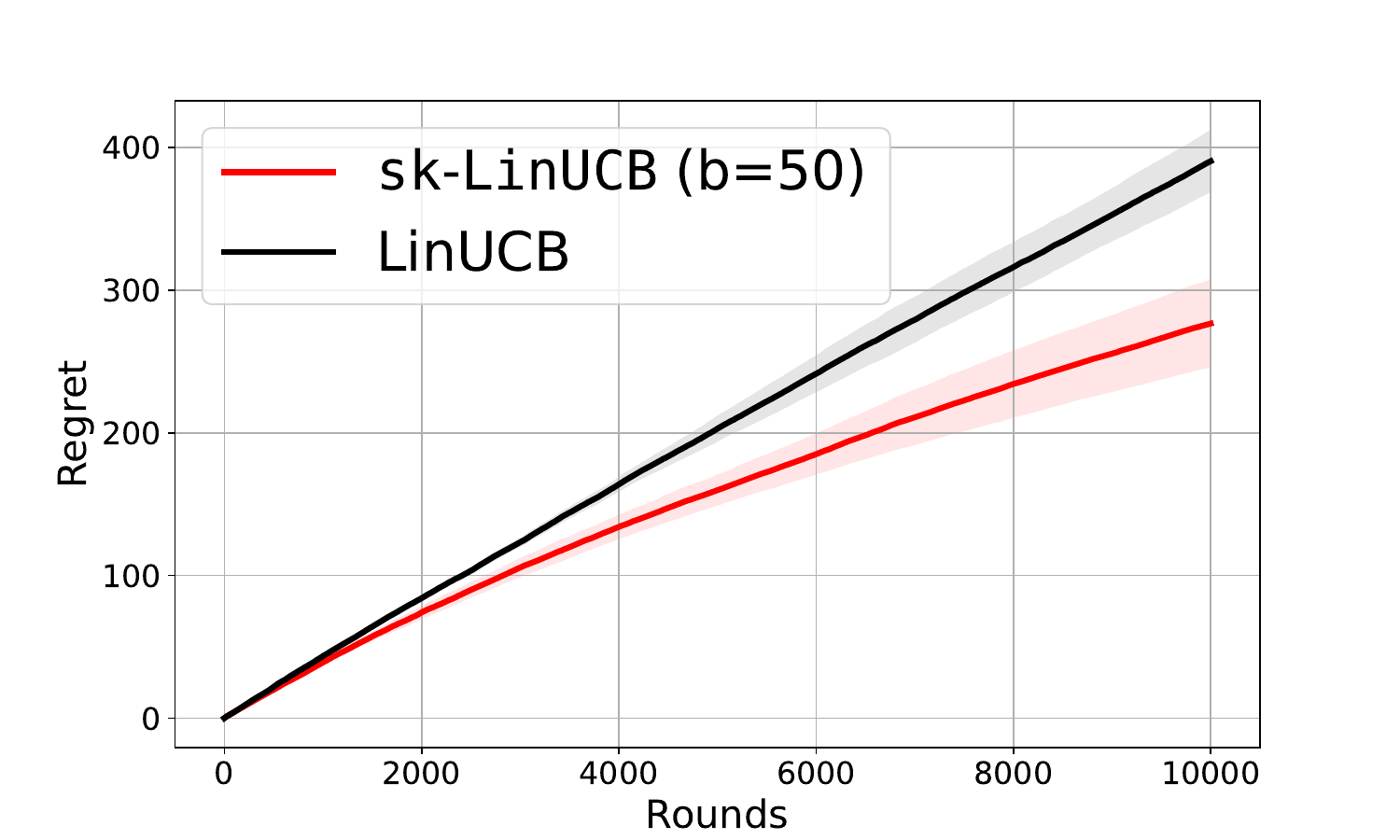}}
    \subfigure{\includegraphics[width=0.32\textwidth]{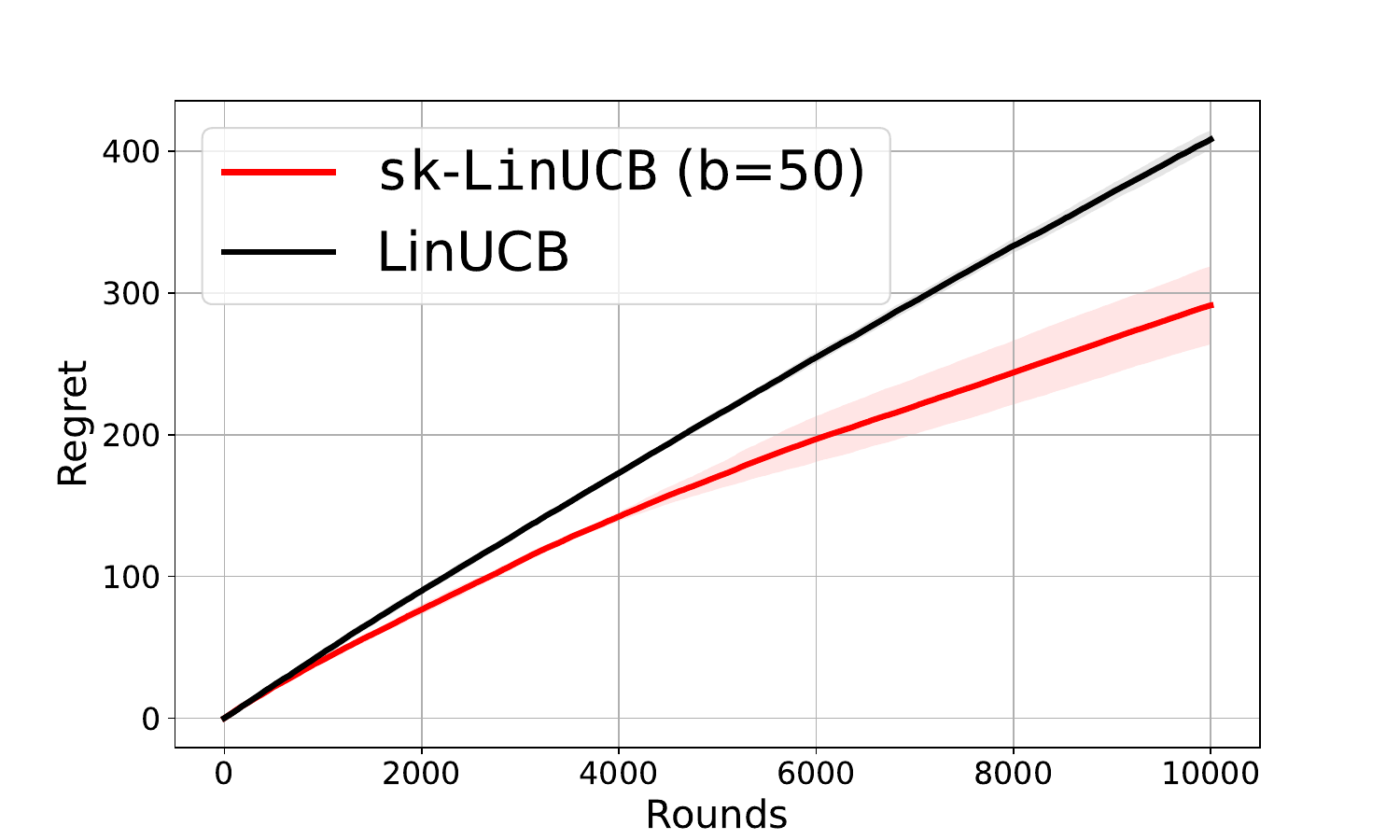}} 
    \subfigure{\includegraphics[width=0.32\textwidth]{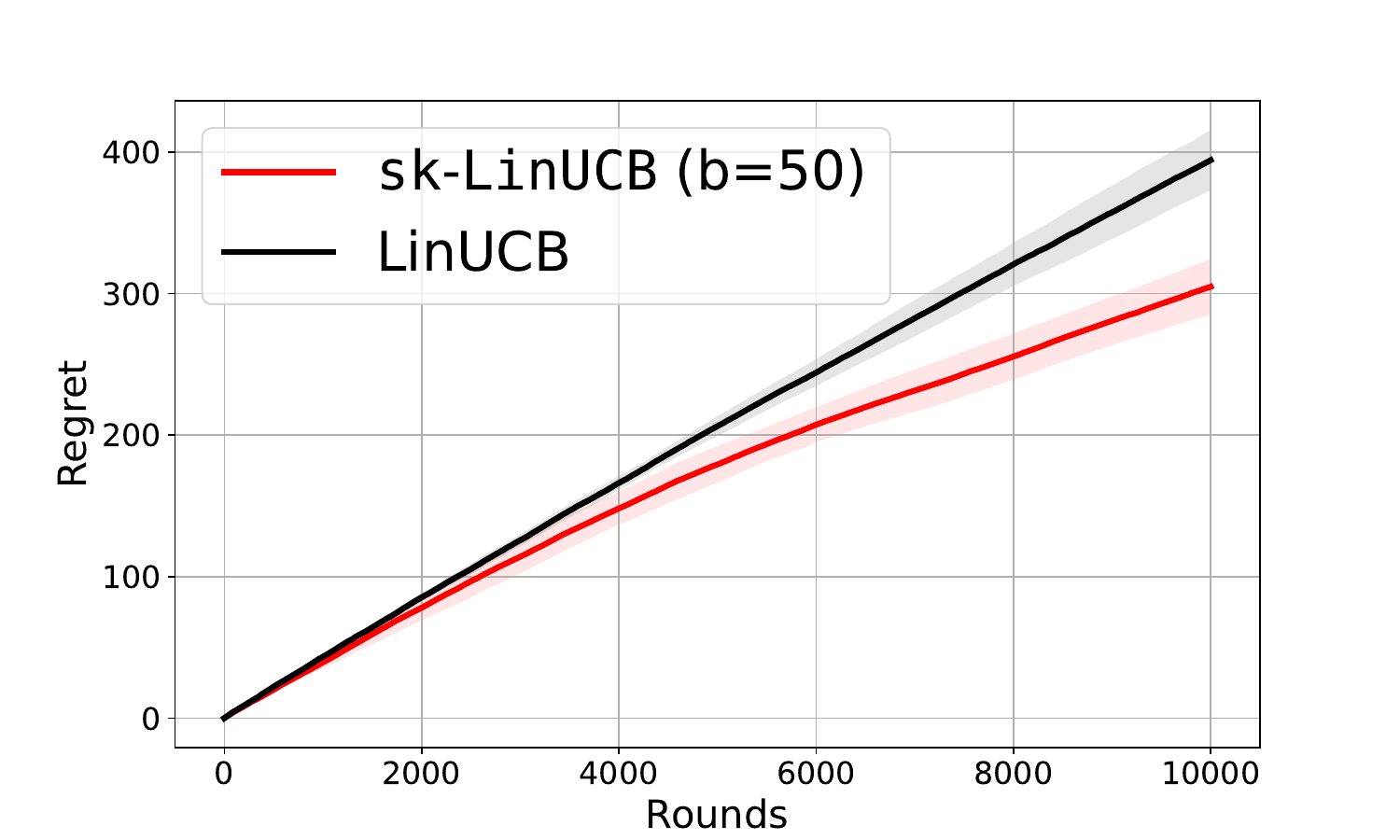}}
    \setcounter{subfigure}{0} 
    \subfigure[]{\includegraphics[width=0.32\textwidth]{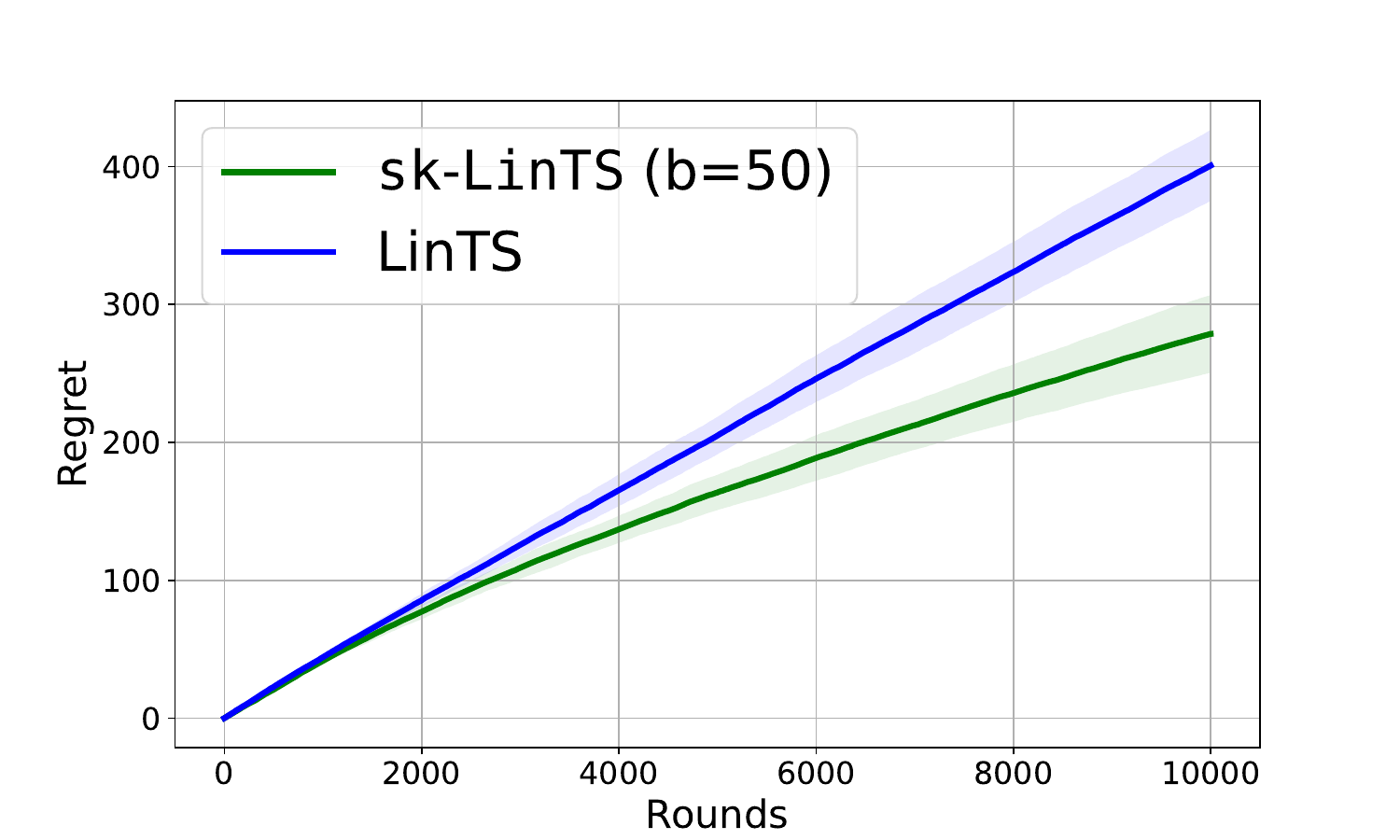}} 
    \subfigure[]{\includegraphics[width=0.32\textwidth]{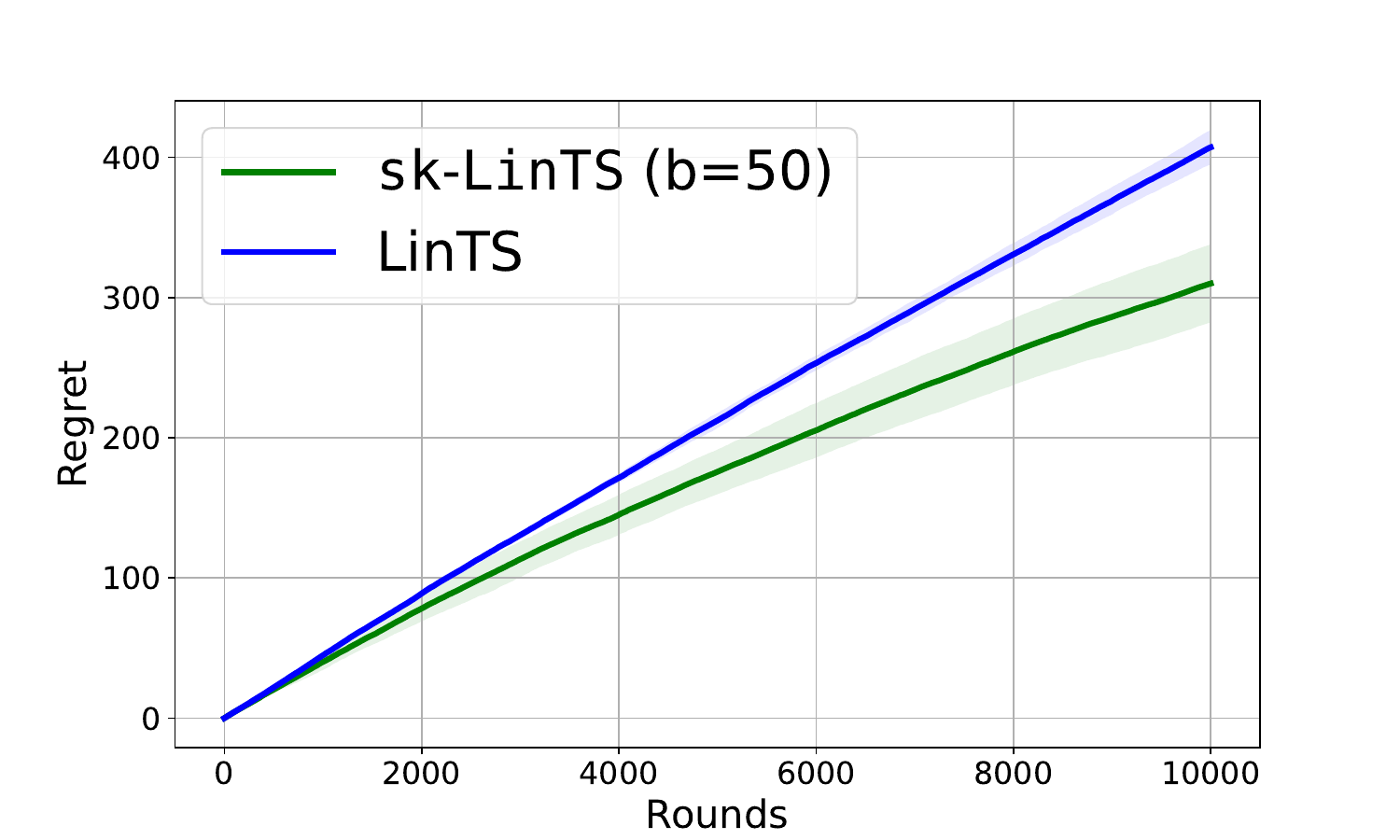}} 
    \subfigure[]{\includegraphics[width=0.32\textwidth]{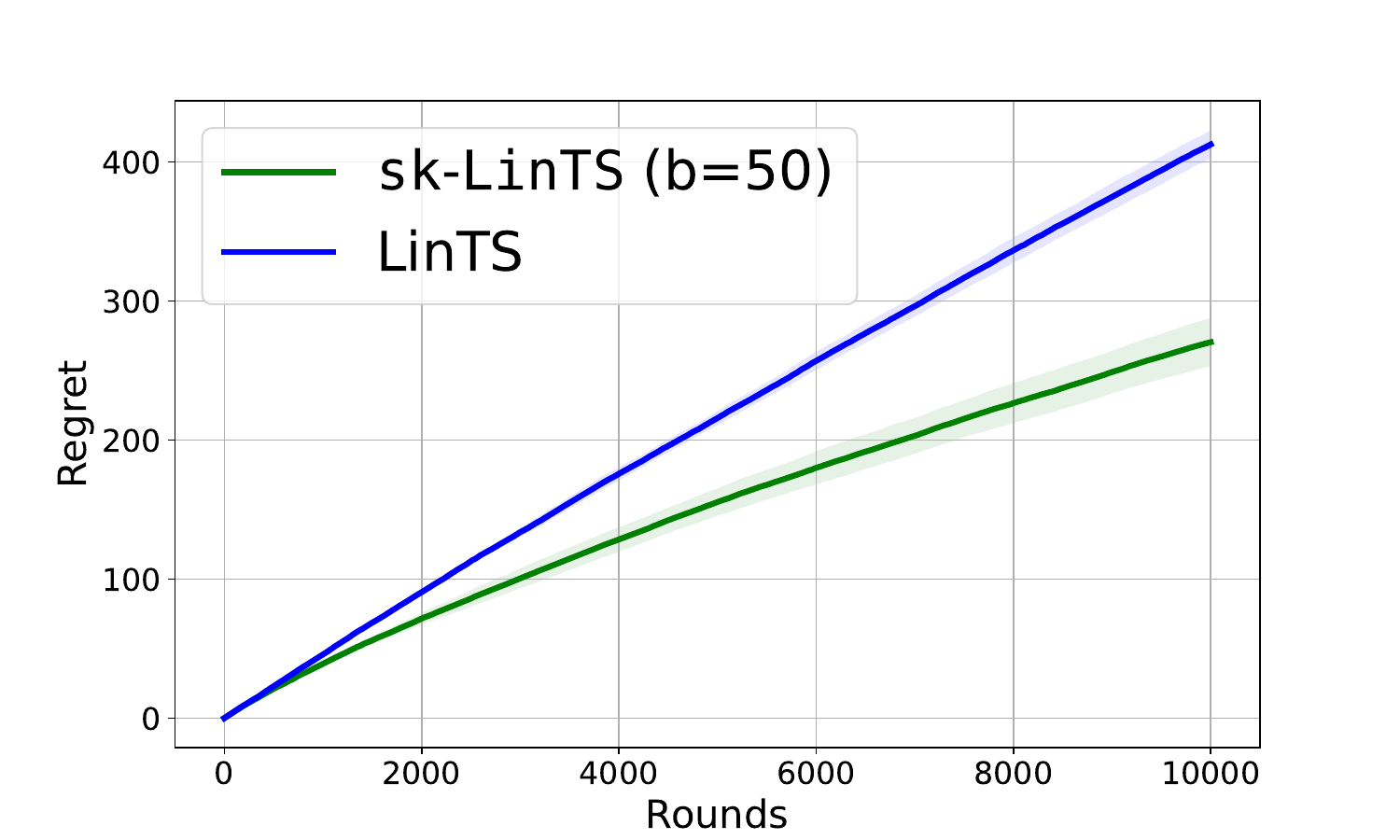}}
    
    \caption{Comparison of cumulative regret of \( \skLinUCB \) and \( \skLinTS \) with the baselines LinUCB and LinTS on a synthetic dataset, averaged over 5 runs. The results are shown for three sparsity cases: (a) context sparsity, (b) parameter sparsity, and (c) both context and parameter sparsity. The sketching dimension \( b = 50 \).}
    \label{fig: regret}
\end{figure*}

\begin{figure*}[!htbp]
    \centering
    \subfigure[$\skLinUCB$]{\includegraphics[width=0.4\textwidth]{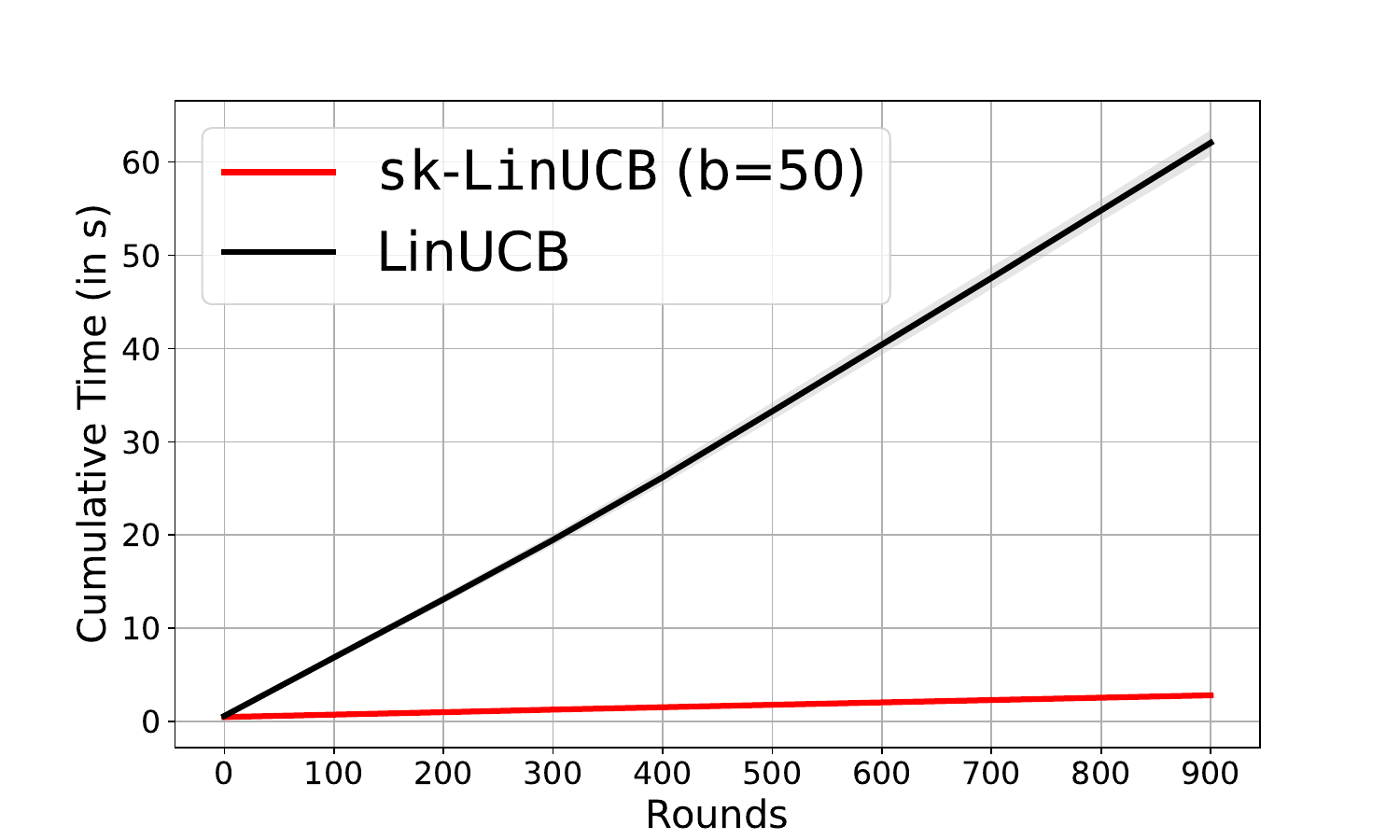}}
    \subfigure[$\skLinTS$]{\includegraphics[width=0.4\textwidth]{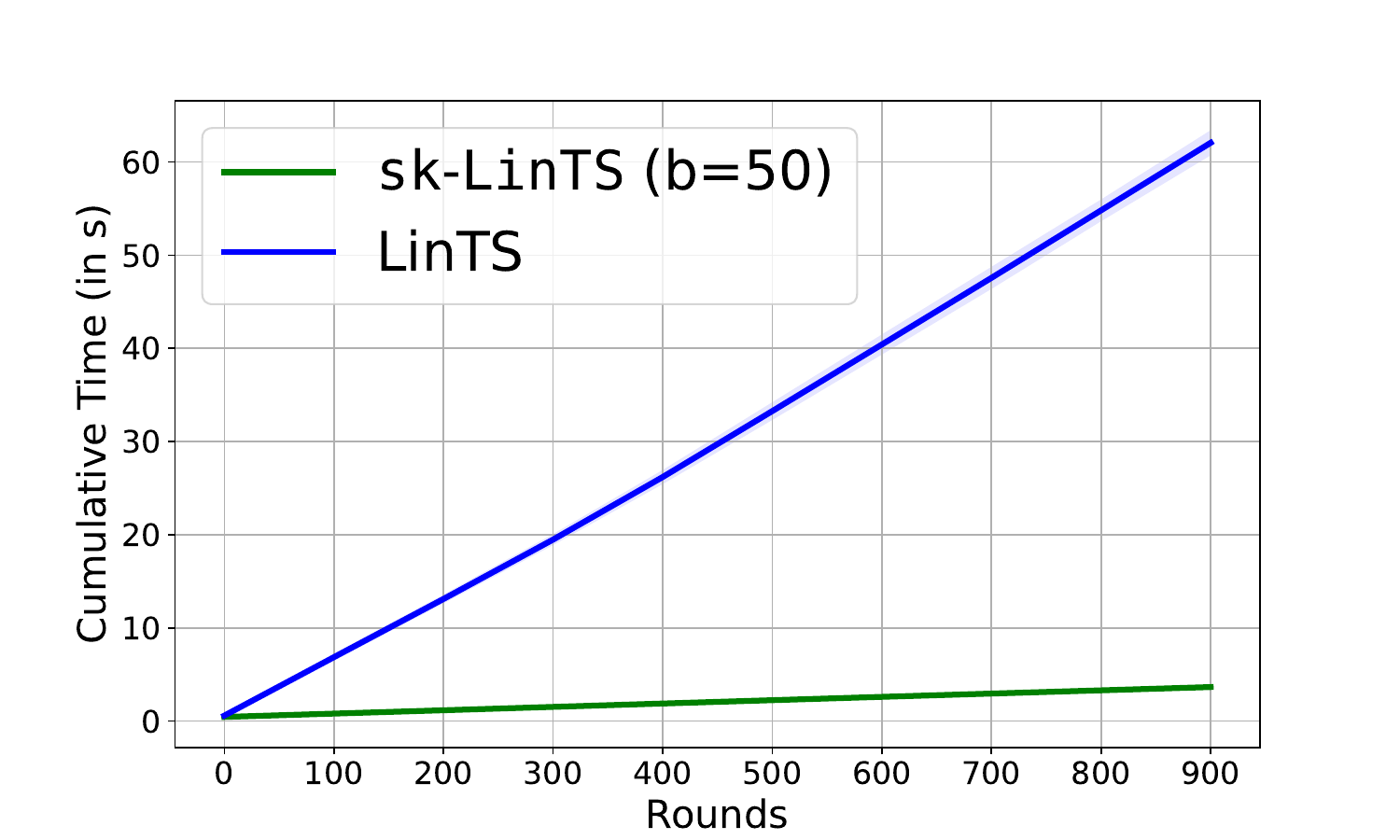}} 
    
    \caption{Comparison of run time (in seconds) for \( \skLinUCB \) and \( \skLinTS \) with the baselines LinUCB and LinTS on a synthetic dataset, averaged over 5 runs. The sketching dimension is denoted by \( b \).}
    \label{fig: comp_tim}
\end{figure*}

\paragraph{Sketching:} In all experiments, we use a random Gaussian matrix \( S \in \mathbb{R}^{b \times d} \), where each entry \( S_{ij} \sim \mathcal{N}(0, 1/b) \).  The sketching dimension is set to \( b = 50 \). The reward function is given by:
\(
    h(a_t) = a_t^T \theta_{*} + \eta_t,
\)
where \( \eta_t \sim \mathcal{N}(0, 1) \).
\paragraph{Results:}The results for both the baseline and sketched algorithms are shown in Figure~\ref{fig: regret}. In all three scenarios, the sketched algorithms achieve lower regret than their unsketched counterparts, even with a sketching dimension of \( b = 50 \). To compare computational cost, we also plot the cumulative runtime for each algorithm in Figure~\ref{fig: comp_tim}, highlighting significant computational savings for the sketched methods.


\end{document}